\documentclass[11pt]{article}

\usepackage[preprint]{acl}

\usepackage{times}
\usepackage{latexsym}
\usepackage{pifont}
\newcommand{\cmark}{\textcolor{green!60!black}{\ding{51}}} %
\newcommand{\xmark}{\textcolor{red}{\ding{55}}}           %
\usepackage{makecell}
\usepackage[utf8]{inputenc} %
\usepackage[T1]{fontenc}    %
\usepackage{hyperref}       %
\usepackage{url}            %
\usepackage{booktabs}       %
\usepackage{amsfonts}       %
\usepackage{nicefrac}       %
\usepackage{microtype}      %
\usepackage{xcolor}         %
\usepackage{times}
\usepackage{latexsym}
\usepackage{microtype}
\usepackage[table]{xcolor}
\usepackage{colortbl}
\usepackage{graphicx}
\usepackage{subcaption}
\usepackage{booktabs} %
\usepackage{hyperref}       %
\usepackage{url}            %
\usepackage{booktabs}       %
\usepackage{amsfonts}       %
\usepackage{nicefrac}       %
\usepackage{microtype}      %
\usepackage{wrapfig}
\usepackage{caption}
\usepackage{booktabs}
\usepackage{multirow}
\usepackage{caption}
\usepackage{amsfonts}       %
\usepackage{nicefrac}       %
\usepackage{microtype}      %
\usepackage{xcolor}         %
\usepackage{tablefootnote}
\usepackage{graphicx}
\usepackage{tikz}
\usepackage{pgfplots}
\pgfplotsset{compat=1.18}
\usepackage{enumitem}
\usepackage{tcolorbox}
\usepackage{enumitem}
\usepackage{xcolor}
\usepackage{xcolor}
\usepackage{colortbl}
\definecolor{lightpink}{RGB}{255, 228, 228}
\definecolor{darkpink}{RGB}{255, 200, 200}
\usepackage{algorithm}
\usepackage{algorithmic}
\usepackage{amsmath}
\usepackage{amssymb}
\usepackage{mathtools}
\usepackage{amsthm}
\usepackage{graphicx}
\usepackage{svg}
\usepackage{comment}
\usepackage[symbol]{footmisc}
\usepackage{booktabs, subcaption}
\usepackage[table,xcdraw]{xcolor}
\usepackage[capitalize,noabbrev]{cleveref}
\crefformat{section}{\S#2#1#3} %
\crefformat{subsection}{\S#2#1#3}
\crefformat{subsubsection}{\S#2#1#3}
\usepackage[T1]{fontenc}

\usepackage[utf8]{inputenc}

\usepackage{microtype}

\usepackage{inconsolata}
\usepackage{gensymb}
\usepackage{graphicx}
\usepackage{todonotes}

\title{Do You See Me : A Multidimensional Benchmark for Evaluating Visual Perception in Multimodal LLMs}

\author{Aditya Kanade \\
  Microsoft Research India \\
  \\\And
  Tanuja Ganu\thanks{Correspondence to: Tanuja Ganu, \texttt{tanuja.ganu@microsoft.com}} \\
  Microsoft Research India \\}

\begin{document}
\maketitle
\begin{figure*}[t]
  \centering
  \includegraphics[width=0.95\textwidth,height=0.45\textwidth]{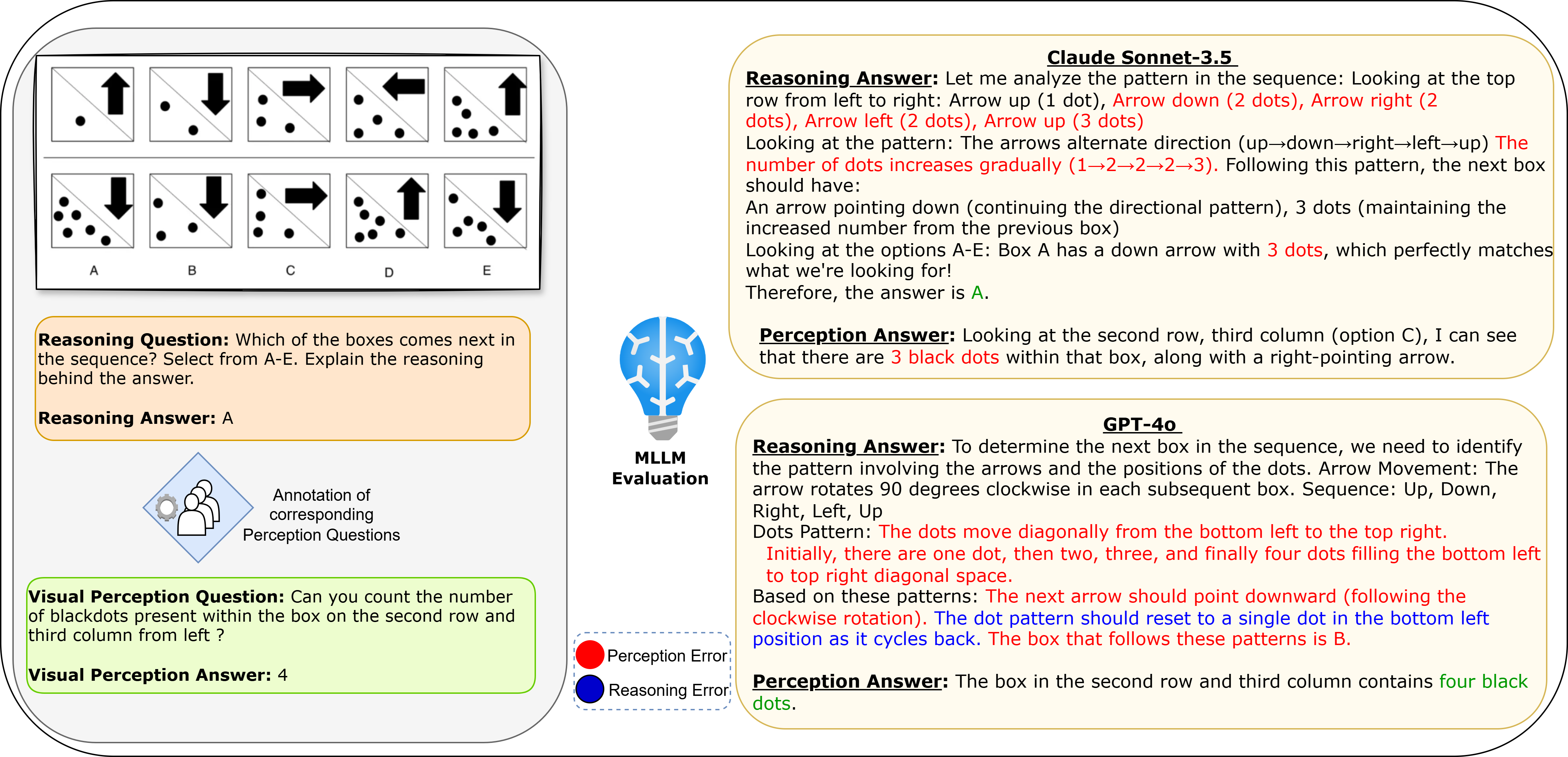}
  \caption{\textit{Visual Misinterpretations in Popular Multimodal LLMs}}
  \label{fig:main-fig}
  \vspace{-3mm}
\end{figure*}
Multimodal Large Language Models (MLLMs) show reasoning promise, yet their visual perception is a critical bottleneck. Paradoxically, MLLMs sometimes produce correct answers while misinterpreting crucial visual elements, masking these underlying perception failures. Our preliminary analysis on a joint perception-reasoning dataset revealed that 29\% of correct reasoning answers from a leading MLLM contained perception errors. To systematically study visual perception  abilities of MLLMs, we introduce \textbf{Do You See Me}- a scalable, programmatically generated benchmark with 1758 images and 2612 questions across seven core subtasks spanning 2D and 3D variants (twelve total tasks) providing parametric control over difficulty levels. The benchmark tasks are inspired by human psychology. Our evaluation of eleven leading MLLMs reveals a stark deficit: humans achieve 95.83\% accuracy, while top MLLMs average below 50\%. This performance gap widens drastically as task complexity increases. Further diagnostics show: (1) supervised finetuning offers only modest gains ($~11\%$), (2) models tend to exploit task “shortcuts” like MCQ formats over detailed visual analysis, and (3) Chain-of-Thought prompting can degrade complex visual tasks by verbalizing images into lossy text.
These findings expose the foundational perception limits in current MLLMs and highlight the need for robust visual perception improvements in MLLMs. The benchmark dataset, source code and evaluation scripts are available at\footnote{\url{https://github.com/microsoft/Do-You-See-Me}}.

\section{Introduction}
Multimodal Large Language Models (MLLMs) demonstrate moderate reasoning capabilities, yet a striking paradox remains: models can produce correct answers while fundamentally misinterpreting crucial visual elements. This phenomenon masks underlying perceptual failures. For instance, a leading model like Claude Sonnet-3.5~\cite{claude3.5sonnet} can arrive at the correct final answer for a logical pattern-completion task despite misinterpreting the visual components of the puzzle (see \cref{fig:main-fig}). Such cases, where correct reasoning outcomes mask foundational perception errors, raise a critical question: do MLLMs truly \emph{see} what they reason about?

This perceptual brittleness is a critical issue, as robust visual perception forms the foundation for trustworthy higher-order reasoning~\cite{chalfant1969central}. The problem has persisted largely because current evaluation paradigms are not equipped to diagnose it. While benchmarks such as MMVP~\cite{tong2024eyeswideshutexploring}, CV-Bench~\cite{tong2024cambrian1fullyopenvisioncentric}, and MVP-Bench~\cite{Li2024MVPBenchCL} provide valuable assessments of general multimodal capabilities, their methodologies are not primarily designed for the isolated and fine-grained analysis of perception. Specifically, their reliance on manual annotation limits scalability, they lack systematic control over task difficulty, and many sub-tasks within these benchmarks often treat perception as a secondary objective. This clear gap in current evaluation methodology motivates our investigation, which is guided by three key research questions: \textbf{RQ1: }\textit{To what extent do correct reasoning outcomes mask underlying perceptual failures?}; \textbf{RQ2: }\textit{How well do MLLMs perform on core visual perception skills across controlled difficulty levels compared to humans ?}; and \textbf{RQ3: }\textit{What structural limitations and failure modes underlie MLLMs’ visual perception deficits ?}

Our investigation begins by probing the premise of RQ1 with a preliminary study. We curated a joint perception-reasoning dataset of 150 logic-based IQ test problems drawn from MathVista~\cite{Lu2023MathVistaEM} and LogicVista~\cite{Xiao2024LogicVistaML}. Our analysis provided initial, striking evidence: for Claude-Sonnet-3.5, 29\% of its correctly answered reasoning questions still contained fundamental visual perception errors. This finding empirically affirms that correct reasoning outcomes can indeed mask perception failures, underscoring the need for a more rigorous and systematic evaluation of visual perception in MLLMs.

Towards this end, we introduce \textbf{Do You See Me}, a benchmark designed to systematically evaluate core visual perception skills in MLLMs. Grounded in established frameworks from human psychology~\cite{chalfant1969central}, the benchmark isolates fundamental abilities such as visual discrimination, figure-ground perception, and spatial relations. To overcome the limitations of prior work, \textbf{Do You See Me} is programmatically generated, ensuring scalability and enabling fine-grained analysis through parametric control over task difficulty. The benchmark comprises $1758$ images and $2612$ questions across seven core subtasks, presented in both 2D and 3D photorealistic scenes variants (twelve total tasks) to assess performance across varied visual contexts. Our evaluations on this benchmark reveal a striking deficit: while human participants achieve 95.83\%\footnote{Macro Average of 2D and 3D performance} accuracy, the best-performing MLLMs average below 50\%, with this performance chasm widening drastically as task complexity increases.

Beyond this performance gap, the benchmark's controlled design enables a deeper diagnostic analysis of MLLM failure modes. For instance, by reformulating an open-ended task into a multiple-choice question, we observed an accuracy jump from 23.19\% to 41.80\%, revealing a reliance on task cues over genuine visual analysis. Our analysis shows that Chain-of-Thought (CoT) prompting can hurt performance on tasks requiring holistic visual understanding, suggesting that translating rich visual input into step-by-step text introduces information loss. Finally, a large-scale supervised finetuning (SFT) experiment yielded only modest gains (approx. $11\%$), confirming that these perceptual limitations are foundational and not easily resolved by data scaling alone.

In summary, this work makes three primary contributions: (i) we introduce a scalable, programmatically generated benchmark, \textbf{Do You See Me}, designed to systematically evaluate core visual perception skills; (ii) we present a curated joint dataset to disentangle reasoning success from underlying perceptual accuracy, which provides the empirical motivation for our work; and (iii) we provide a comprehensive evaluation of eleven leading MLLMs that quantifies their perceptual deficits and provides a diagnostic analysis of their failure modes.

\section{Related Work}
While many benchmarks evaluate the high-level reasoning of Multimodal Large Language Models (MLLMs) \cite{Lu2023MathVistaEM, Zhang2024MathVerseDY, Lu2022LearnTE, Saikh2022ScienceQAAN, Li2024MultimodalAA, Yue2023MMMUAM}, a growing body of work reveals that their foundational visual perception is a critical bottleneck \cite{Zhang2024MathVerseDY, wu2024qbench}. For instance, recent studies demonstrate that even state-of-the-art models fail at trivial visual tasks like counting overlapping shapes or identifying a circled letter \cite{rahmanzadehgervi2025visionlanguagemodelsblind}. Such failures have been theoretically linked to the cognitive science concept of the binding problem, with MLLMs showing human-like performance degradation in tasks requiring feature binding, like conjunctive visual search \cite{campbell2025understandinglimitsvisionlanguage}. This highlights a clear gap between the models' touted complex reasoning abilities and their actual performance on basic perceptual tasks.

\begin{table}[ht]
\centering
\caption{Comparison of multimodal benchmarks across key dimensions.}
\label{tab:benchmark_comparison}
\resizebox{\linewidth}{!}{
\begin{tabular}{@{}lcccccc@{}}
\toprule
\textbf{Benchmark} & \textbf{\makecell{Automated\\Data\\Collection}} & \textbf{\makecell{Difficulty\\Levels}} & \textbf{Categories} & \textbf{Size} & \textbf{\makecell{Human\\Annotation}} \\
\midrule
MVP-Bench~\cite{Li2024MVPBenchCL} & \xmark & \xmark & \makecell{13 high-level\\5 low-level} & 605 & \cmark \\
MME~\cite{Fu2023MMEAC} & \xmark & \xmark & 14 & 1,147 & \xmark \\
CV-Bench~\cite{tong2024cambrian1fullyopenvisioncentric} & \xmark & \xmark & 4 & 2,638 & \xmark \\
MMVP~\cite{tong2024eyeswideshutexploring} & \xmark & \xmark & 9 & 300 & \cmark \\
\midrule
\textbf{Do You See Me (Ours)} & \cmark & \cmark & 7 & \textbf{2,612\textsuperscript{*}} & \cmark \\
\bottomrule
\end{tabular}}
\vspace{2pt}

\raggedright
\small\textsuperscript{*}Can be extended in an automated manner.
\end{table}

Existing benchmarks that focus on visual perception, such as MMVP \cite{tong2024eyeswideshutexploring}, CV-Bench \cite{tong2024cambrian1fullyopenvisioncentric}, MME \cite{Fu2023MMEAC}, and MVP-bench \cite{Li2024MVPBenchCL}, present several key methodological limitations. As summarized in Table~\ref{tab:benchmark_comparison}, they are often derived from common datasets like ImageNet \cite{Russakovsky2014ImageNetLS} and COCO \cite{Lin2014MicrosoftCC}, posing a risk of data contamination since these are likely part of MLLM training corpora. Furthermore, their reliance on manual annotation limits scalability, and they generally lack mechanisms for parametrically controlling task difficulty, which hinders a more fine-grained analysis of model capabilities.

Drawing inspiration from human psychology for a more principled evaluation, our work is grounded in established motor-free assessments such as the Test of Visual Perceptual Skills (TVPS) \cite{tvps} and the Motor-Free Visual Perception Test (MVPT) \cite{mvpt}. These tests are designed to systematically isolate and assess core perceptual dimensions \cite{chalfant1969central} while avoiding confounding factors like motor skills \cite{mvpt, tvps, hammill2016dtvp}. Accordingly, we introduce the Do You See Me benchmark, which leverages these established perceptual categories and a programmatic generation pipeline to offer a scalable and difficulty-controlled evaluation of MLLM visual skills, directly addressing the limitations of prior benchmarks.

\section{Preliminary Study - Joint Visual Perception and Reasoning Dataset}
\label{sec:prelimnary-study}

Most of the existing benchmarks assess MLLM's visual reasoning capabilities by solely relying on the final answer based accuracy.
However, this approach can obscure the exact source of errors.
In particular, three primary sources of error can arise: 1) Visual Perception- inaccuracies in identifying or interpreting elements in the provided image, 2) Reasoning- errors in the logical or conceptual steps used to arrive at the final answer, or 3) Arithmetic- mistakes in performing numerical or algebraic calculations. 
To accurately distinguish between different error sources, it is essential to analyze not only final answers but also the reasoning chains. We introduce a \emph{joint perception-reasoning} dataset specifically designed to separate visual perception errors from higher-level reasoning failures.

\vspace{2pt}
\noindent \textbf{Why IQ-Type Questions?} 
IQ-style diagrammatic questions primarily feature basic geometric shapes and patterns, minimizing reliance on domain-specific knowledge. This allows for a focused evaluation of visual perception and reasoning skills without introducing extraneous complexity.

\subsection{Data Collection}
\label{sec:joint-vpd-dc}

Our dataset is drawn from two established visual reasoning benchmarks: MathVista~\cite{Lu2023MathVistaEM} and LogicVista~\cite{Xiao2024LogicVistaML}. We selected logic-based tasks centered around geometric shapes and pattern recognition from:
\begin{itemize}[topsep=0pt, leftmargin=1.0em]
\setlength{\itemsep}{0pt}  %
\setlength{\parsep}{0pt}   %
\setlength{\topsep}{0pt}   %
    \item \emph{IQtest} subset of MathVista (focusing on spatial and pattern-based problems).
    \item \emph{Diagrams} subset of LogicVista (pattern completion tasks).
\end{itemize}
These subsets feature universally understood shapes in controlled layouts, allowing systematic evaluation of perception and reasoning. We curated $15$ problems from MathVista’s \emph{IQtest} and $135$ from LogicVista’s \emph{Diagrams}, yielding a total of $150$ examples in our final dataset.

\subsection{Data Annotation}
\label{sec:joint-vpd-an}
We extend each original problem $(I, R, A_R)$---where $I$ is the image, $R$ is the reasoning question, and $A_R$ is the corresponding ground-truth answer---by adding a \emph{visual perception} question $P$ with its ground-truth $A_P$. The extended sample is thus: $\bigl(I,\, (R, A_R),\, (P, A_P)\bigr)$. The perception questions are manually devised such that they are directly relevant to each reasoning question (e.g., ``How many triangles are in the figure?’’). More details on dataset creation and distribution are added in \cref{sec:appendix-prelim-dataset-stats}.

\section{Do You See Me}
\label{sec:do-you-see-me}

\begin{figure*}[h!]
    \centering
\includegraphics[width=\linewidth, height=0.5\linewidth]{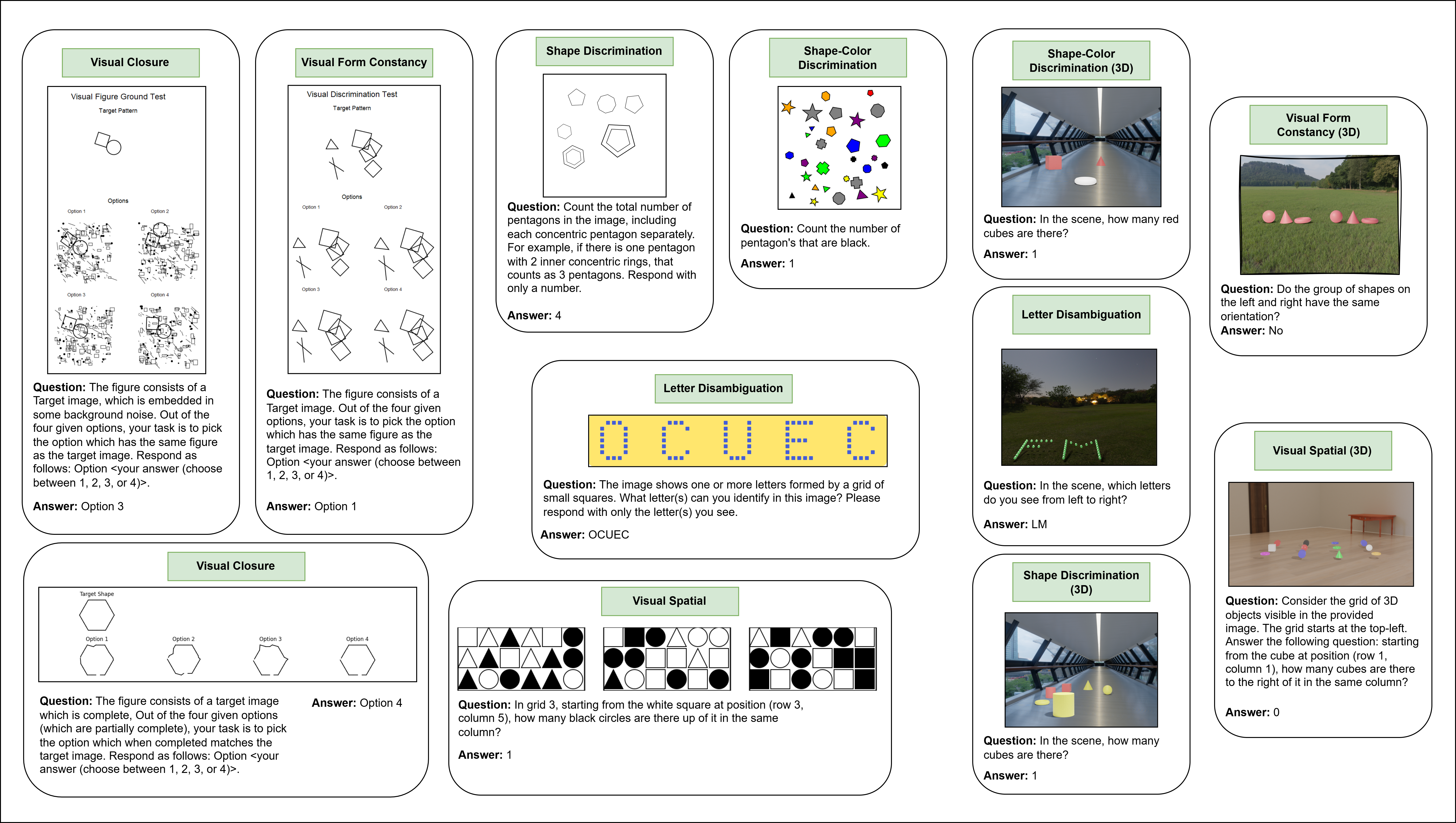}
    \caption{\textit{\textbf{Do You See Me} benchmark visual perception dimensions}}
    \label{fig:syn-dataset}
    \vspace{-3mm}
\end{figure*}

Human psychology systematically categorizes human visual perception as a combination of five core abilities \cite{chalfant1969central}: visual discrimination, the ability to recognize dominant features (e.g., position, shape, form, color); visual figure-ground, the ability to distinguish the main object from its background; visual memory, the ability to remember sequences of presented images; visual closure, the ability to complete partially obscured shapes; and visual spatial, the ability to perceive positions of objects relative to oneself and to other objects. Assessments such as the \textit{Test of Visual Perception Skills (TVPS)} \cite{tvps} and \textit{Motor-Free Visual Perception Test (MVPT)} \cite{mvpt} systematically evaluate these abilities through structured visual tasks and associated questions. While there are fundamental differences between MLLM and human visual processing, the perceptual categories defined in human psychology offer an established framework for structured evaluation of analogous capabilities in MLLMs. Building on these principles, we introduce the \textbf{Do You See Me} benchmark, a fully automated test suite designed to evaluate MLLM visual perception across dimensions analogous to human tests.

\subsection{General Benchmark Design} 
Consistent with many human visual perception tests, stimuli in the \textbf{Do You See Me} benchmark are primarily generated in a 2D setting using Scalable Vector Graphics (SVG), allowing for precise control over geometric properties. To enable a broader analysis of MLLM capabilities on more complex scenes, the benchmark also incorporates photorealistic 3D rendered settings for several subtasks. These 3D scenes are produced using Blender \cite{blender}, drawing on a prior work for scene setup \cite{candle}. For 3D tasks, standard shapes include five basic forms: sphere, cube, cone, cylinder, and torus. Correspondingly, 2D tasks utilize fundamental SVG-generated geometric figures such as capsules, stars, hexagons, circles, pentagons, rectangles, and triangles, \cref{tab:task_statistics} lists the distribution of 2D and 3D subtasks.

\begin{table}[t]
\centering
\small
\caption{Dataset statistics for 2D and 3D visual perception tasks in Do You See Me benchmark.}
\label{tab:task_statistics}
\resizebox{!}{0.4\linewidth}{
\begin{tabular}{@{}lp{2.8cm}ccc@{}}
\toprule
\textbf{Set.} & \textbf{Task} & \textbf{Imgs} & \textbf{Ques.} & \textbf{Type} \\
\midrule
\textbf{2D} & Shape Discrimination & 241 & 241 & Int \\
& Joint Shape-Color & 90 & 408 & Int \\
& Letter Discrimination & 135 & 135 & Text \\
& Form Constancy & 270 & 270 & MCQ \\
& Spatial Grids & 270 & 806 & Int \\
& Visual Figure-Ground & 90 & 90 & MCQ \\
& Visual Closure & 166 & 166 & MCQ \\
\cmidrule(lr){1-5}
\textbf{3D} & Shape Discrimination & 120 & 120 & Int \\
& Joint Shape-Color & 120 & 120 & Int \\
& Letter Discrimination & 96 & 96 & Text \\
& Form Constancy & 80 & 80 & MCQ \\
& Spatial Grids & 80 & 80 & Int \\
\midrule
\textbf{Total (2D)} & & \textbf{1,262} & \textbf{2,116} & \\
\textbf{Total (3D)} & & \textbf{496} & \textbf{496} & \\
\textbf{Overall} & & \textbf{1,758} & \textbf{2,612} & \\
\bottomrule
\end{tabular}}
\end{table}

The \textbf{Do You See Me} benchmark probes distinct facets of visual perception (in 2D and photorealistic 3D) in MLLMs through seven core subtasks. \textit{Shape Discrimination} involves identifying or counting specified shapes, often in cluttered or occluded scenes, and requires a numeric answer. \textit{Joint Shape-Color Discrimination} requires linking shapes with their designated colors, also necessitating a numeric response.
\textit{Letter Disambiguation} focuses on recognizing textual characters despite varied presentation conditions, with answers provided as text. The \textit{Form Constancy} task challenges MLLMs to identify the correct matching group from four options; the other three present slight transformations (e.g., size, rotation). \textit{Spatial Grids} assesses comprehension of object arrangements and relationships within grid structures. In \textit{Visual Figure-Ground}, the aim is to select a target pattern amidst a noisy background. \textit{Visual Closure} requires choosing the option that correctly completes a partially presented shape. A distinct 3D version of \textit{Visual Figure-Ground} is not included, as our photorealistic 3D rendered scenes inherently test this ability. Furthermore, a 3D equivalent for \textit{Visual Closure} is omitted due to ambiguities in judging 3D shape closure from a single camera perspective. \cref{tab:visual-perception-constraints} and Table \ref{tab:visual-perception-constraints-3D} defines parameters for fine-grained control over image generation. To establish difficulty-modulating parameter ranges for each subtask, a preliminary evaluation was conducted. These limits were determined by observing when the GPT-4o \cite{openai2024gpt4ocard} model started to exhibit a high degree of failure.

A detailed description of the benchmark's construction, including the specific control parameters for modulating difficulty in each subtask, can be found in \cref{sec:appendix-do-you-see-me}. We open-source the synthetic data generation code and the dataset at \footnote{https://anonymous.4open.science/r/DoYouSeeMe-F52E/README.md}.

\subsection{Synthetic Data Generation and Human Performance Benchmarking}
\label{sec:human-perf-benchmarking}
To benchmark human performance against current MLLMs, we conducted a study with fifteen human subjects using this dataset. For each subtask, participants were presented with two randomly selected visual perception questions for each combination of control parameters. They typed their answers and rated each question's difficulty (Easy, Moderate, or Hard). To minimize bias and ensure consistency, all participants underwent a calibration phase before each subtask, involving seven distinct examples spanning the subtask's difficulty range, and everyone received the same calibration and main study examples. Human accuracy for each subtask was determined by comparing their answers to the ground truth. Further details on the human benchmarking protocol are available in \cref{sec:appendix-human-benchmarking}.

\section{Experiments}

\subsection{Experimental Setup and Evaluation}
\label{sec:exp-eval}

We evaluate the models listed in \cref{tab:models-evaluated} on Do You See Me. To ensure fair comparison, all models receive identical visual content and uniform textual prompts. Performance of models is measured in terms of final answer accuracy. We closely follow evaluation protocols laid out by \cite{Lu2023MathVistaEM, Xiao2024LogicVistaML}. Please refer \cref{sec:appendix-experimental-details} for a detailed evaluation protocol.

\begin{table}[h!]
\centering
\caption{Overview of the MLLMs evaluated in our study.}
\label{tab:models-evaluated}
\resizebox{\columnwidth}{!}{%
\begin{tabular}{@{}ll@{}}
\toprule
\textbf{Closed-Source (Proprietary API)} & \textbf{Open-Source (Local Inference)} \\
\midrule
\makecell[l]{\textsc{Gemini-1.5 Flash} \\ {\footnotesize\cite{geminiteam2024gemini15unlockingmultimodal}}} & \makecell[l]{\textsc{Llama-3.2-11B-Vision} \\ {\footnotesize\cite{grattafiori2024llama3herdmodels}}} \\
\addlinespace
\makecell[l]{\textsc{GPT-4o} \\ {\footnotesize\cite{openai2024gpt4ocard}}} & \makecell[l]{\textsc{Gemma-3-12B-Instruct} \\ {\footnotesize\cite{gemmateam2025gemma3technicalreport}}} \\
\addlinespace
\makecell[l]{\textsc{Claude-Sonnet-3.5} \\ {\footnotesize\cite{claude3.5sonnet}}} & \makecell[l]{\textsc{Phi-4-Multimodal-Instruct-5.7B} \\ {\footnotesize\cite{abdin2024phi4technicalreport}}} \\
\addlinespace
\makecell[l]{\textsc{Gemini-2.0-Flash} \\ {\footnotesize\cite{GoogleDeepmindGemini2024}}} & \makecell[l]{\textsc{Qwen2.5-VL-7B-Instruct} \\ {\footnotesize\cite{wang2024qwen2vlenhancingvisionlanguagemodels}}} \\
\addlinespace
\makecell[l]{\textsc{Gemini-2.5-Flash} \\ {\footnotesize\cite{comanici2025gemini25pushingfrontier}}} & \makecell[l]{\textsc{InternVL2.5-8B} \\ {\footnotesize\cite{chen2025expandingperformanceboundariesopensource}}} \\
\addlinespace
\makecell[l]{\textsc{OpenAI-O3} \\ {\footnotesize\cite{openai2024openaio1card}}} & \\
\bottomrule
\end{tabular}%
}
\end{table}

\begin{table*}[t]
\centering
\caption{Performance of various models on 2D and 3D visual perception tasks. Values are reported as mean accuracy$_{\text{std. dev.}}$. Row colors distinguish between Human performance (yellow), closed-source models (blue), and open-source models (gray).}
\label{tab:visual_perception_all_combined}
\definecolor{opensourcecolor}{gray}{0.9} %
\definecolor{closedblue}{RGB}{235, 242, 250}  %
\definecolor{humanyellow}{RGB}{253, 245, 205}

\resizebox{\textwidth}{!}{%
\begin{tabular}{>{\raggedright\arraybackslash}m{3.5cm}|ccccccc|c||ccccc|c}
\toprule
& \multicolumn{8}{c||}{\textbf{2D Visual Perception Tasks}} & \multicolumn{6}{c}{\textbf{3D Visual Perception Tasks}} \\
\cmidrule(lr){2-9} \cmidrule(l){10-15}
\textbf{Model} & \begin{tabular}[c]{@{}c@{}}Figure\\ Ground\end{tabular} & \begin{tabular}[c]{@{}c@{}}Visual\\ Spatial\end{tabular} & \begin{tabular}[c]{@{}c@{}}Color\\ Disamb.\end{tabular} & \begin{tabular}[c]{@{}c@{}}Shape\\ Disamb.\end{tabular} & \begin{tabular}[c]{@{}c@{}}Letter\\ Disamb.\end{tabular} & \begin{tabular}[c]{@{}c@{}}Form\\ Const.\end{tabular} & \begin{tabular}[c]{@{}c@{}}Visual\\ Closure\end{tabular} & \begin{tabular}[c]{@{}c@{}}Avg.\\ Acc.\end{tabular} & \begin{tabular}[c]{@{}c@{}}Color\\ Disamb.\end{tabular} & \begin{tabular}[c]{@{}c@{}}Letter\\ Disamb.\end{tabular} & \begin{tabular}[c]{@{}c@{}}Shape\\ Disamb.\end{tabular} & \begin{tabular}[c]{@{}c@{}}Form\\ Const.\end{tabular} & \begin{tabular}[c]{@{}c@{}}Visual\\ Spatial\end{tabular} & \begin{tabular}[c]{@{}c@{}}Avg.\\ Acc.\end{tabular} \\
\midrule
\rowcolor{humanyellow}
Human & 98.52$_{3.8}$ & 93.33$_{3.1}$ & 100.00$_{0.0}$ & 89.72$_{8.17}$ & 84.94$_{6.12}$ & 98.15$_{1.35}$ & 94.72$_{1.84}$ & \textbf{94.19}$_{3.48}$ & 96.11$_{1.45}$ & 99.58$_{1.13}$ & 96.44$_{0.83}$ & 97$_{2.71}$ & 98.21$_{1.08}$ & \textbf{97.46}$_{1.44}$ \\
\midrule
\rowcolor{closedblue}
Claude Sonnet-3.5 & 41.48$_{2.8}$ & 35.28$_{0.83}$ & 76.8$_{0.14}$ & 33.75$_{1.1}$ & 15.56$_{0.74}$ & 90.74$_{0.37}$ & 57.07$_{2.49}$ & 50.09$_{0.95}$ & 66.67$_{1.18}$ & 20.83$_{2.95}$ & 73.33$_{0.0}$ & 18.12$_{0.88}$ & 34.38$_{0.88}$ & 42.66$_{1.17}$ \\
\rowcolor{closedblue}
Gemini-1.5-Flash & 37.04$_{0.64}$ & 13.19$_{0.38}$ & 46.98$_{0.37}$ & 15.14$_{1.27}$ & 27.16$_{0.86}$ & 80.37$_{0.0}$ & 49.6$_{0.91}$ & 38.49$_{0.39}$ & 78.89$_{0.48}$ & 5.9$_{0.6}$ & 46.67$_{2.2}$ & 54.17$_{1.44}$ & 31.25$_{3.31}$ & 43.37$_{1.60}$ \\
\rowcolor{closedblue}
GPT-4o & 29.26$_{4.49}$ & 27.42$_{1.88}$ & 74.51$_{0.65}$ & 23.19$_{0.64}$ & 32.35$_{0.86}$ & 74.94$_{1.19}$ & 56.94$_{2.48}$ & 45.51$_{1.28}$ & 96.67$_{1.44}$ & 18.4$_{4.21}$ & 80.83$_{0.83}$ & 42.92$_{5.64}$ & 37.08$_{8.04}$ & 55.18$_{4.03}$\\
\rowcolor{closedblue}
O3 (2025-04-16) & 25.19$_{3.39}$ & 35.16$_{0.43}$ & 81.11$_{0.25}$ & 39.7$_{1.94}$ & 53.71$_{1.11}$ & 66.34$_{1.67}$ & 49.51$_{4.73}$ & 50.10$_{1.50}$ & 86.39$_{0.96}$ & 22.57$_{3.35}$ & 80.83$_{1.67}$ & 29.17$_{1.44}$ & 41.67$_{3.15}$ & 52.12$_{2.11}$ \\
\rowcolor{closedblue}
Gemini 2.5 Flash & 8.15$_{1.7}$ & 54.28$_{1.18}$ & 88.4$_{0.37}$ & 53.89$_{1.92}$ & 30.62$_{1.13}$ & 51.85$_{1.34}$ & 51.59$_{1.91}$ & 48.39$_{0.51}$ & 71.39$_{1.27}$ & 4.17$_{2.08}$ & 77.78$_{2.1}$ & 36.67$_{4.02}$ & 9.17$_{4.73}$ & 39.83$_{2.84}$\\
\rowcolor{closedblue}
Gemini 2.0 Flash & 34.44$_{1.11}$ & 39.66$_{1.65}$ & 88.73$_{0.42}$ & 40.83$_{0.72}$ & 35.8$_{1.13}$ & 79.63$_{1.34}$ & 72.82$_{2.09}$ & 55.98$_{0.51}$ & 89.44$_{1.27}$ & 17.71$_{1.04}$ & 85.28$_{0.48}$ & 25.83$_{1.44}$ & 47.08$_{4.39}$ & 53.06$_{1.72}$ \\
\midrule
\rowcolor{opensourcecolor}
Qwen2.5-VL-7B-Instruct & 29.63$_{4.49}$ & 25.35$_{0.72}$ & 61.6$_{0.71}$ & 17.22$_{0.24}$ & 4.69$_{0.43}$ & 56.79$_{1.07}$ & 47.62$_{3.09}$ & 34.70$_{1.49}$ & 78.89$_{0.48}$ & 0.69$_{0.6}$ & 79.72$_{0.96}$ & 79.58$_{1.44}$ & 36.67$_{7.22}$ & 55.11$_{2.14}$ \\
\rowcolor{opensourcecolor}
Gemma-3-12B-instruct & 31.11$_{1.92}$ & 10.22$_{0.14}$ & 64.13$_{1.84}$ & 12.92$_{1.44}$ & 8.15$_{0.0}$ & 44.94$_{2.14}$ & 28.97$_{1.72}$ & 28.63$_{0.81}$ & 81.11$_{0.96}$ & 3.82$_{0.6}$ & 53.89$_{0.48}$ & 45.0$_{0.0}$ & 32.5$_{0.0}$ & 43.26$_{0.40}$\\
\rowcolor{opensourcecolor}
Phi-4-Multimodal-Instruct-5.7B & 23.7$_{3.21}$ & 18.44$_{1.37}$ & 24.67$_{1.84}$ & 7.78$_{1.2}$ & 6.42$_{0.43}$ & 23.58$_{1.5}$ & 18.85$_{0.34}$ & 17.63$_{0.89}$ & 65.0$_{2.2}$ & 0.69$_{0.6}$ & 47.5$_{1.44}$ & 17.5$_{2.17}$ & 30.42$_{9.04}$ & 32.22$_{3.09}$ \\
\rowcolor{opensourcecolor}
InternVL2.5-8B & 30.00$_{6.42}$ & 16.00$_{0.36}$ & 55.64$_{0.25}$ & 19.17$_{1.68}$ & 0.74$_{0.43}$ & 34.81$_{3.32}$ & 29.76$_{3.78}$ & 26.58$_{2.41}$ & 82.50$_{0.36}$ & 0.00$_{0}$ & 66.67$_{1.34}$ & 53.75$_{2.23}$ & 33.75$_{4.43}$ & 47.33$_{1.67}$ \\
\rowcolor{opensourcecolor}
Llama3.2-11B-Vision-Instruct & 26.67$_{2.34}$ & 6.08$_{1.23}$ & 25.74$_{0.97}$ & 3.33$_{1.43}$ & 11.85$_{0.72}$ & 22.22$_{1.31}$ & 22.02$_{2.21}$ & 16.84$_{1.46}$ & 78.33$_{0.72}$ & 2.08$_{0.23}$ & 69.17$_{1.74}$ & 90.00$_{1.89}$ & 10.00$_{2.21}$ & 49.91$_{1.35}$\\
\bottomrule
\end{tabular}%
}
\end{table*}

\vspace{-2mm}
\subsection{Experimental Results}
\subsubsection{Joint Perception-Reasoning Dataset}
\label{sec:joint-vpd-obs}
To probe the relationship between reasoning success and perceptual accuracy, we evaluate MLLMs on a \textbf{joint perception--reasoning dataset}. Each sample in this dataset contains an image $(I)$, a perception question $(P)$, and a reasoning question $(R)$. Each model is separately prompted to answer both questions with explicit instructions to provide a detailed chain‐of‐thought. We employ an LLM‐based grader with expert human oversight (detailed in Section~\ref{sec:exp-eval}) to score the correctness of the answers.

\begin{figure}[h!]
    \centering
    \begin{subfigure}{0.48\textwidth}
        \centering
        \includegraphics[width=\linewidth]{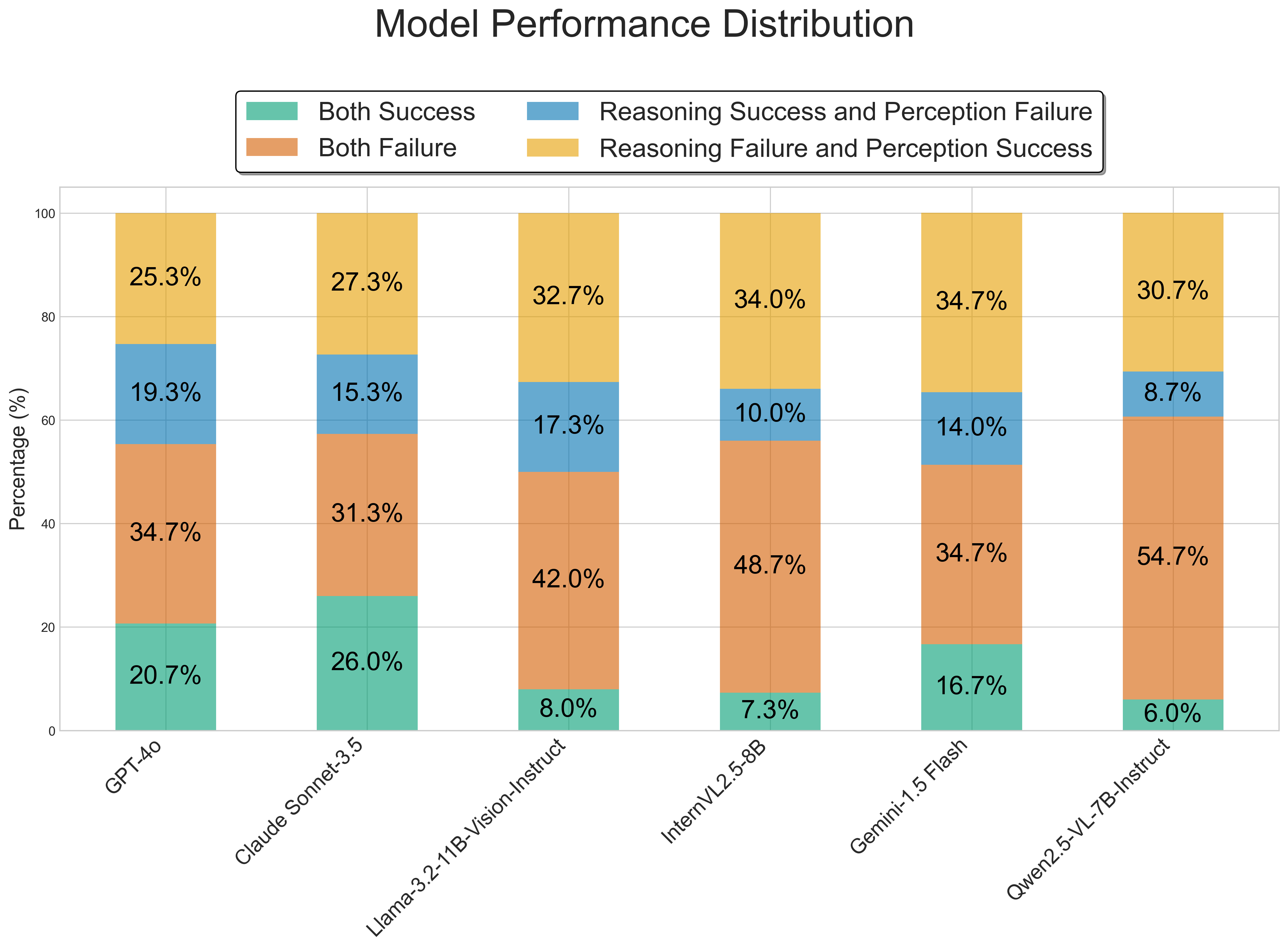}
        \caption{MLLM performance on joint perception-reasoning questions.}
        \label{fig:joint-reasoning-perception-plot}
    \end{subfigure}
    \hfill
    \begin{subfigure}{0.23\textwidth}
        \centering
        \includegraphics[width=\linewidth]{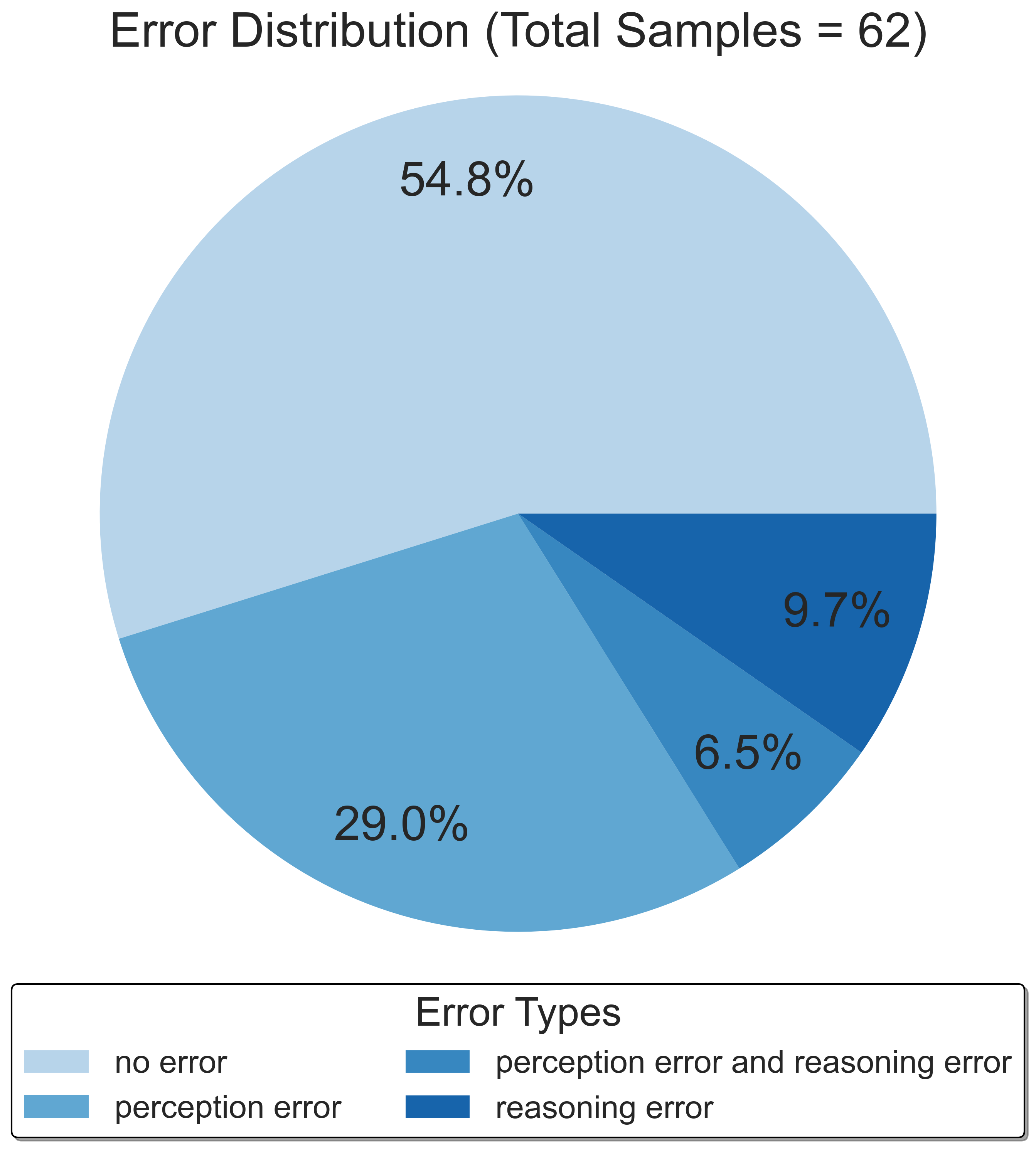}
        \caption{Error distribution for correct final answers.}
        \label{fig:reasoning-chain-error-dist-correct}
    \end{subfigure}
    \hfill
    \begin{subfigure}{0.23\textwidth}
        \centering
        \includegraphics[width=\linewidth]{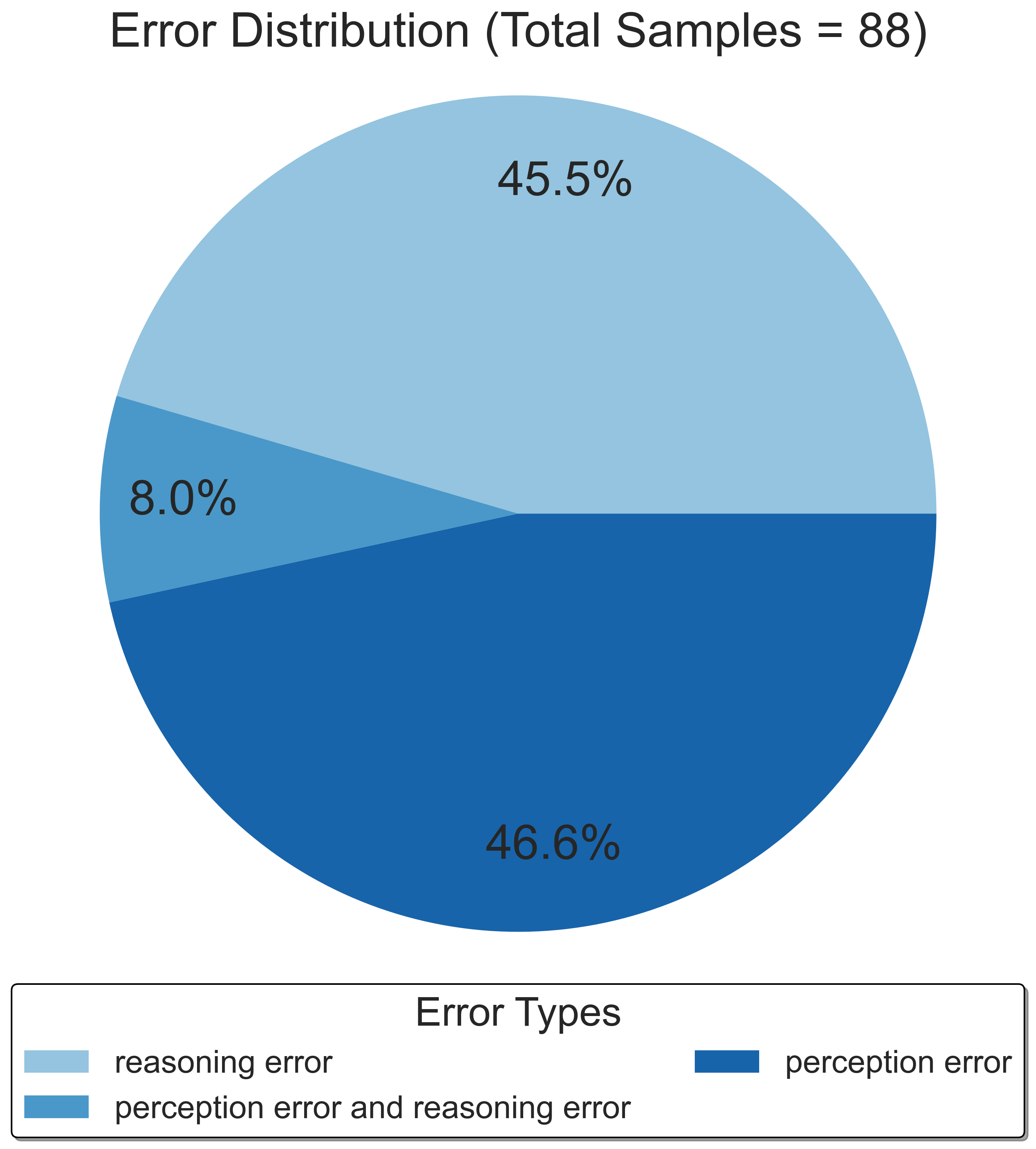}
        \caption{Error distribution for incorrect final answers.}
        \label{fig:reasoning-chain-error-dist-incorrect}
    \end{subfigure}
    \caption{\textit{Comparison of MLLM visual reasoning performance (a) and error breakdowns (b, c) for correct and incorrect final answers respectively (Claude Sonnet-3.5).}}
    \label{fig:joint-vis-perception}
\end{figure}
\textbf{Perception vs. Reasoning Performance:} \cref{fig:joint-reasoning-perception-plot} presents the distribution of MLLM performance on the paired perception and reasoning questions. While all evaluated models demonstrate varying degrees of failure across these tasks, a noteworthy and recurring pattern emerges: in several instances, models correctly answer the reasoning questions while simultaneously failing on the corresponding perception questions designed to assess their understanding of crucial visual elements. %

\textbf{Visual Perception Errors in MLLM Reasoning Response: } We manually annotated responses from the highest-performing model- \emph{Claude Sonnet-3.5}, categorizing each error as visual perception, reasoning, or arithmetic (as defined in Section~\ref{sec:prelimnary-study}). \cref{fig:joint-vis-perception} presents the error distribution for Claude Sonnet-3.5. For incorrect final answers, we found both visual perception and reasoning errors present in the response chain. Notably, even when models produced correct final answers, we identified visual perception errors in 29.0\% of these responses, with only 54.8\% of correct responses being free of all error types. These findings suggest that models may arrive at correct answers despite misperceiving visual elements, highlighting the importance of comprehensive evaluation of visual perception capabilities independent of reasoning performance.

\begin{tcolorbox}[colback=gray!10,colframe=gray!70]
\textbf{Insight: }%
Relying on final-answer accuracy to judge MLLMs’ visual reasoning can create a false sense of success, masking persistent perception errors beneath correct outcomes.
\end{tcolorbox}

\subsubsection{Do You See Me}

In this section, we present a comprehensive evaluation of popular MLLMs on \textbf{Do You See Me}. Further, we also extensively analyze possible root cause of failure.

\textbf{Overall Performance:} \cref{tab:visual_perception_all_combined} illustrate MLLM accuracy across the twelve visual subtasks in their respective 2D and 3D settings. A consistent and primary finding across both 2D and 3D tasks is that current MLLMs fall significantly short of human performance. On average, humans achieve an accuracy of 95.83\% (see \cref{sec:appendix-detailed-results} for detailed per-task scores), whereas the best-performing MLLMs operate at considerably lower accuracy levels across nearly all subtasks in both dimensional settings. Further, we observe that overall human subjects exhibit very relatively significantly low-variance across participants.

\begin{figure}[htbp]
\centering
    \begin{subfigure}{0.49\linewidth}
        \centering
        \includegraphics[width=\linewidth]{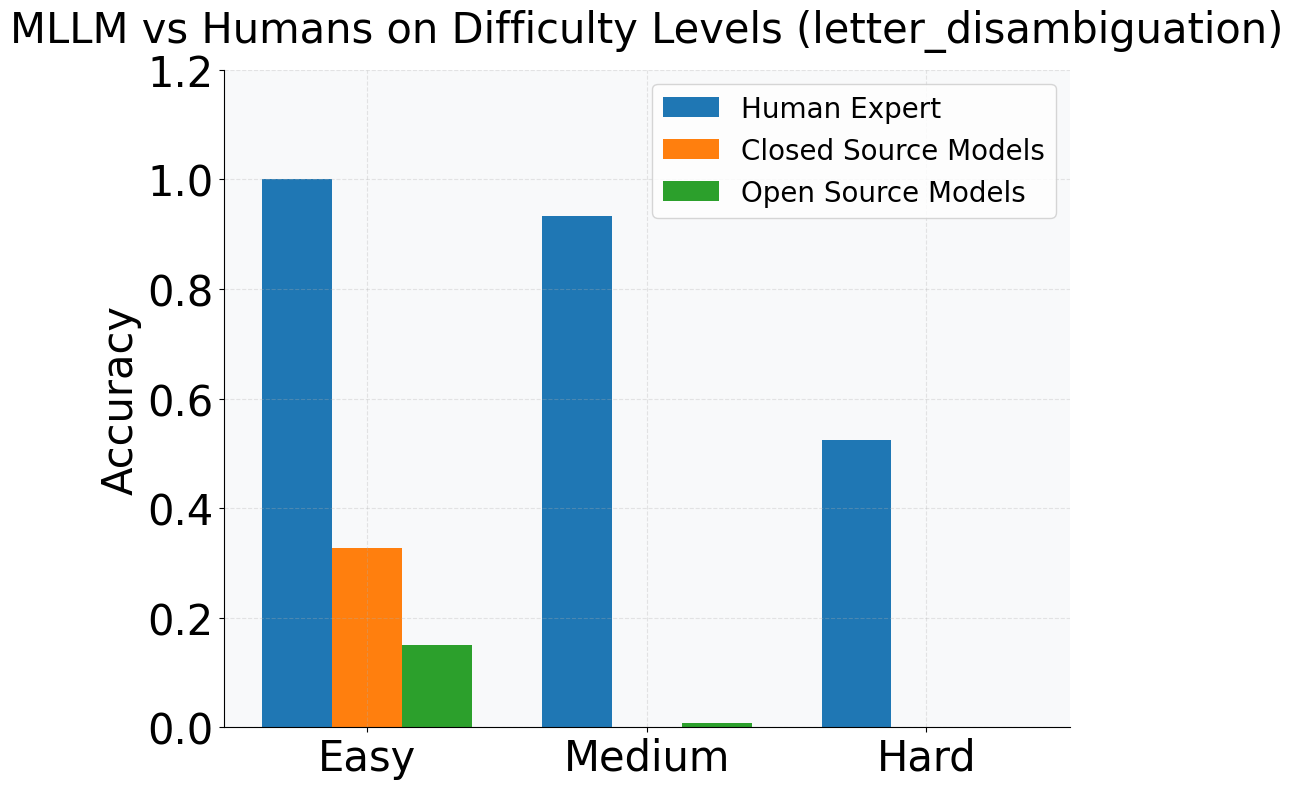}
        \caption{Letter Disambiguation}
        \label{fig:fig2}
    \end{subfigure}
    \hfill
    \begin{subfigure}{0.49\linewidth}
        \centering
        \includegraphics[width=\linewidth]{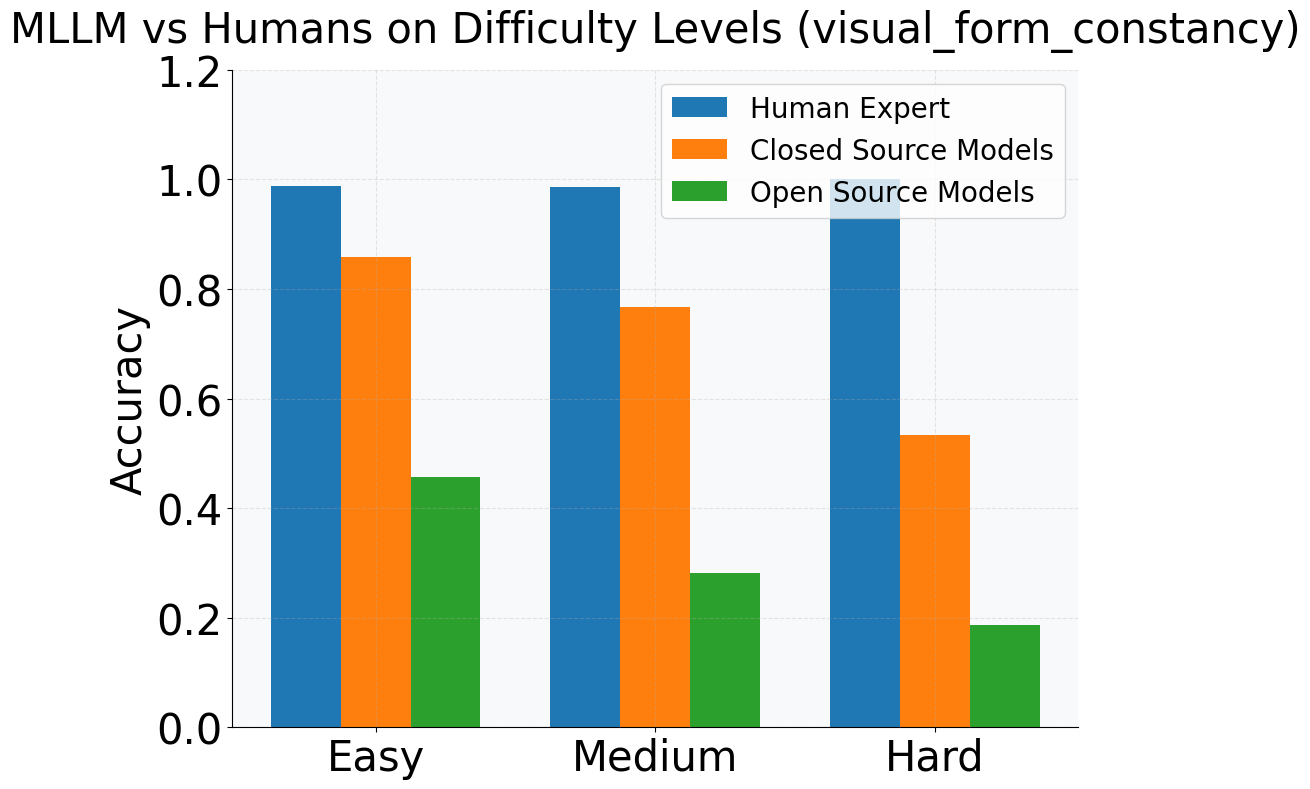}
        \caption{Visual Form Constancy}
        \label{fig:fig3}
    \end{subfigure}
\caption{\textit{Comparison of MLLM and human performance across controlled difficulty levels: (a) on Letter Disambiguation task, and (b) on Visual Form Constancy.}}
\label{fig:human-mllm-difficulty}
\end{figure}

\textbf{MLLM Performance Across Human-Perceived Difficulty Levels}: Using perceived difficulty ratings (Easy, Medium, and Hard) collected from human subjects across all seven visual perception tasks, we compared MLLM performance (grouped as open- vs. closed-source) against human performance. ~\cref{fig:human-mllm-difficulty} reveals striking patterns: in visual form constancy, the human performance gap for closed-source models widens from 12\% (Easy) to 45\% (Hard), while in letter disambiguation, both model types fail completely at Medium difficulty and beyond as humans maintain high accuracy. Similar degradation appears across all subtasks (see \cref{sec:appendix-detailed-results}).

\textbf{MLLM Performance Over Increasing Difficulty Control Parameters: } As described in \cref{sec:do-you-see-me}, we established control parameters for each of the seven subtasks to modulate visual stimuli difficulty in \textbf{Do You See Me}. We observe that all MLLMs consistetly exhibit a decline in visual perception performance as the subtasks become harder, culminating in near-zero accuracy at the most challenging settings (refer \cref{sec:appendix-detailed-results} for plots).

\begin{tcolorbox}[colback=gray!10,colframe=gray!70]
\textbf{Insight: }MLLMs show a marked visual perception deficit relative to humans on the \textbf{Do You See Me} benchmark. The gap widens sharply as task complexity increases—whether human-rated or parametrically controlled—while human accuracy remains stable.
\end{tcolorbox}

\section{Discussion}
\label{sec:discussion}

Evaluations on \textbf{Do You See Me} expose key weaknesses in current MLLMs: minimal gains from fine-tuning, sensitivity to visual complexity, and instability in visual reconstruction.

\noindent\textbf{Comparing 2D and 3D Perceptual Performance}
MLLMs' performance varies unpredictably between 2D and 3D tasks. Stronger performance in one setting does not guarantee transfer to the other (\cref{fig:syn-dataset-2D} and \cref{fig:syn-dataset-3D});—likely reflecting pretraining bias toward real-world (3D) imagery. For example, factors like occlusion and noise in 2D stimuli typically reduces MLLM accuracy in Shape Discrimination and Figure-Ground tasks, but exhibits stronger performance on similar 3D setting.

\noindent\textbf{Limited Gains from Supervised Finetuning:}
To assess if supervised finetuning (SFT) addresses core visual perception limits, we finetuned Qwen2.5-VL-7B-Instruct—with approximately 67,000 new benchmark-conformant image-text pairs (methodology in \cref{sec:appendix-mllm-finetuning-details}). While this SFT improved the model's average accuracy by approximately 11\% (from 40.91\% to 51.75\%), its performance remained significantly below human accuracy of 95.83\%. Therefore, merely scaling SFT with more benchmark-like data appears insufficient to overcome fundamental visual perception limitations in MLLMs, highlighting the need for alternative approaches.

\noindent\textbf{Impact of Task Format on Final Answer Correctness}
We analyze the possibility that high final-answer accuracy is artificially inflated by the subtask format. For example, even for humans, it is much more likely to land on the correct answer using the process of elimination when answering Multiple-Choice Questions (MCQ) versus providing open-ended answers. To test whether MLLMs also exploit task format cues to achieve higher accuracy, we performed a supplementary experiment converting an open-ended task into an MCQ format. Specifically, we adapted the open-ended Shape Discrimination (2D) subtask. The effect was significant: GPT-4o's accuracy \textit{jumped} from 23.19\% ± 0.64\% on the original task to 41.80\% ± 2.54\% on the MCQ version. This performance gain indicates that the model is not performing 'from-scratch' visual perception but is instead exploiting non-visual cues within the provided options. This finding demonstrates that high final-answer accuracy can mask a lack of true multimodal understanding by allowing the model to exploit task format cues.

\noindent\textbf{Probing Visual Stimuli Reconstruction:}
To probe how MLLMs internally represent visual inputs, we prompted models to first \textit{reconstruct 2D stimuli as SVG code} before answering related questions. This served as an explicit test of their low-level visual perception and structural understanding. However, reconstruction-based prompting generally \textit{failed to improve}—and often \textit{degraded}—task accuracy compared to direct visual questioning (quantitative results in Appendix~\ref{sec:appendix-visual-stimuli-reconstruction}). Qualitative inspection revealed frequent reconstruction errors in shape, orientation, and spatial alignment, underscoring a key perceptual bottleneck. 
Interestingly, the quality of SVG reconstructions \textit{correlates with downstream task performance}: models that generated more faithful stimuli (e.g., \emph{Claude} on color–shape disambiguation) also performed better on corresponding tasks. This suggests that visual reconstruction can serve as a useful \textbf{probing exercise}—revealing not only whether MLLMs can “see” correctly, but also how perception quality constrains reasoning success.

\noindent\textbf{The Role of Verbalization in Multimodal Reasoning.}
To probe how language-based reasoning interacts with visual perception, we conducted an ablation comparing \textit{direct-answer} and \textit{Chain-of-Thought (CoT)} prompting.
In the CoT setup, models were instructed to reason step-by-step before giving a final answer. 

We find that a task’s \textbf{verbalizability}—\textit{the degree to which its visual content can be accurately and completely expressed in language}—is a key determinant of CoT effectiveness. 
High-verbalizability tasks, such as \textit{Visual Spatial} or \textit{Letter Disambiguation}, benefit from CoT because describing spatial arrangements or symbolic details provides a structured reasoning scaffold.
In contrast, low-verbalizability tasks, such as \textit{Form Constancy} or \textit{Visual Closure}, depend on holistic visual understanding that cannot be faithfully captured in words. 
Here, CoT prompting often \textit{degrades} performance by imposing a \textbf{lossy verbalization} of visual information, causing models to reason over their own flawed verbal reconstructions rather than the raw perceptual signal.

This trend is mirrored in model comparisons: \textit{O3}, optimized for structured internal reasoning, significantly outperforms GPT-4o on high-verbalizability tasks, but not on low-verbalizability ones.
Together, these findings suggest that verbalizability fundamentally governs when language-based reasoning helps or hinders multimodal understanding.

\begin{tcolorbox}[colback=gray!10,colframe=gray!70]
\textbf{Insight: } MLLMs can reason, but they still struggle to see. Their perception gaps persist beyond finetuning and worsen when vision is forced into words— underscoring the need for better perceptual grounding. %

\end{tcolorbox}

\section{Conclusion}

In this work, we first motivated the need for a strong visual perception dataset by documenting flawed visual perception in seemingly correct reasoning answers by current MLLMs; our preliminary analysis found that 29\% of correct reasoning answers from a leading MLLM still contained perception errors. We proposed Do You See Me as a diagnostic tool, a programmatically generated and scalable benchmark with over 2,600 questions designed to evaluate core visual skills. Our evaluations showed a stark performance deficit: on our human-psychology-inspired benchmark, human performance was approximately 95.83\%, while most MLLMs performed below 50\%. We also found that while humans maintain a high degree of accuracy on harder samples, MLLM performance drops drastically as task complexity increases.

Further diagnostic analyses revealed these deficits to be foundational. Supervised finetuning offered only modest gains of around 11\% , and we found that models often bypass genuine perception by exploiting task "shortcuts," with performance on one task jumping from 23.19\% to 41.80\% in an MCQ format. Moreover, the utility of Chain-of-Thought (CoT) prompting is contingent on a task's "verbalizability," often degrading performance on holistic visual tasks by forcing a lossy translation of the image into text. These findings underscore that robust visual perception remains a critical bottleneck, suggesting that current models can reason over what they are told, but still struggle to truly see.

\section{Limitations}
\label{sec:appendix-limitations}
Our work has a few limitations that we acknowledge and plan to address in future research. First, the size of our joint perception-reasoning dataset is relatively small. However, we have made every effort to include all possible samples where it was feasible to generate non-ambiguous and correlated visual perception questions. To address this limitation, we plan to employ LLM + Image-Diffusion techniques in the future to generate a more diverse and controlled format of perception+reasoning questions, thus expanding our dataset. Next, in the current setup, we have restricted our visual perception prompts to the English language only, including the letter disambiguation task. This decision was made in the interest of managing the overall cost of benchmarking closed-source MLLMs. However, we recognize the importance of language diversity and plan to expand our coverage to other non-English languages in future iterations of our work. Overall, we believe that our work provides a valuable contribution to the understanding of MLLM capabilities in visual perception tasks and lays the foundation for future research in this area.

\bibliography{main}
\newcommand{\mytcolorbox}[2]{
    \begin{tcolorbox}[colback=gray!10, colframe=gray!80, 
    title=Example #1]
        #2
    \end{tcolorbox}
}

\newcommand{\analysisimage}[2]{
    \begin{figure}[H]
        \centering
        \includegraphics[width=0.3\textwidth]{#1}
        \caption{#2}
        \label{fig:example-#1}
        \vspace{-5mm}
    \end{figure}
}

\newpage
\appendix
\onecolumn

\newpage
\section{Table of Contents}
\begin{enumerate}
    \item \hyperref[sec:appendix-experimental-details]{Experimental Details}
    \begin{itemize}
        \item Closed-Source Model Configuration %
        \item Open-Source Model Configuration   %
        \item Evaluation Protocol             %
    \end{itemize}
    \item \hyperref[sec:appendix-prelim-dataset-stats]{Prelimnary Dataset Statistics}
    \item \hyperref[sec:appendix-do-you-see-me]{Do You See Me - Additional Details}
    \item \hyperref[sec:appendix-do-you-see-me-control-parameters-det]{Do You See Me - Control Parameter Details}
    \item \hyperref[sec:appendix-detailed-results]{Detailed Results}
    \item \hyperref[sec:appendix-human-benchmarking]{Human Performance Benchmarking}
    \item \hyperref[sec:appendix]{Joint Perception-Reasoning Dataset - Qualitative Analysis}
    \begin{itemize}
        \item Correct Reasoning Incorrect Perception %
        \item Qualitative Analysis of Reasoning Chain Errors %
    \end{itemize}
    \item \hyperref[sec:appendix-mllm-prompts]{MLLM Prompts}
    \item \hyperref[sec:appendix-parameter-importance-study]{Parameter Importance}
    \item \hyperref[sec:appendix-mllm-finetuning-details]{Finetuning Details}
    \item \hyperref[sec:appendix-visual-stimuli-reconstruction]{Visual Stimuli Reconstruction}
    \item \hyperref[sec:appendix-sparse-attention]{Sparse Attention to Query Relevant Regions}
    \item \hyperref[sec:appendix_limits_vp]{Limits of Visual Perception}
    \begin{itemize}
        \item \hyperref[subsec:form_constancy_appendix]{Visual Form Constancy: Sensitivity to Rotation at Varying Scales}
        \item \hyperref[subsec:shape_counting_appendix]{Visual Discrimination: Shape Counting Accuracy at Varying Scales}
    \end{itemize}
\end{enumerate}

\newpage
\section{Experimental Details}
\label{sec:appendix-experimental-details}
We evaluated a range of MLLMs, categorized as follows:
\begin{itemize}
    \item \textbf{Closed-Source Models:}
    \begin{itemize}
        \item GEMINI-1.5 FLASH 
        \item GEMINI-2.0 FLASH 
        \item GEMINI-2.5 FLASH 
        \item GPT-4o (\texttt{gpt-4o\_2024-08-06}) 
        \item CLAUDE-SONNET-3.5 (\texttt{claude-3-5-sonnet-2024102}) 
        \item O3 (\texttt{o3\_2025-04-16}) 
    \end{itemize}
    \item \textbf{Open-Source Models:}
    \begin{itemize}
        \item LLAMA-3.2-11B-VISION 
        \item GEMMA-3-12B-INSTRUCT 
        \item PHI-4-MULTIMODAL-INSTRUCT-5.7B
        \item QWEN2.5-VL-7B-INSTRUCT 
        \item INTERNVL2.5-8B 
    \end{itemize}
\end{itemize}

\subsection*{Closed-Source Model Configuration}
The closed-source models were accessed via their respective proprietary APIs (Google AI API for gemini-1.5-flash, OpenAI API for gpt-4o\_2024-08-06, and Anthropic API for claude-3-5-sonnet-20241022). For all closed-source model evaluations, consistent generation parameters were used to ensure fair comparison:
\begin{itemize}
    \item Temperature: 1.0
    \item Top P (nucleus sampling): 0.95
    \item Maximum new tokens: 200
\end{itemize}

\subsection*{Open-Source Model Configuration}
The open-source MLLMs were run locally on a single NVIDIA A100 80GB GPU. To ensure fair comparison and reproducibility, the same hyperparameter settings used for the closed-source models were also applied to the open-source models for evaluation purposes:
\begin{itemize}
    \item Temperature: 1.0
    \item Top P (nucleus sampling): 0.95
    \item Maximum new tokens: 200
\end{itemize}

\subsection*{Evaluation Protocol}
All models were provided with identical visual content and uniform textual prompts for each task. Recent Large Language Models (LLMs) and Multimodal Large Language Models (MLLMs) are increasingly instructed to produce extended textual outputs rather than concise responses, making earlier rule-based or template-matching methods~\cite{Lu2022LearnTE} difficult to apply. Inspired by recent benchmarks for MLLMs~\cite{Lu2023MathVistaEM, Zhang2024MathVerseDY}, we employ an expert LLM to evaluate answers. Our framework proceeds in three stages. In the first stage, a MLLM generates a detailed response according to a predefined template (see ~\cref{sec:appendix-mllm-prompts}), which includes the task description, the question, and possible choices. Next, an answer extractor (~\cref{sec:appendix-mllm-prompts}), based on GPT-4o \cite{openai2024gpt4technicalreport}, parses these extended outputs to yield a concise answer. Prior work has shown that such an expert LLM can extract the correct answer with near 100\% accuracy \cite{Lu2023MathVistaEM}. Finally, the extracted text is standardized (e.g., reduced to multiple-choice labels or numeric values), and performance metrics are computed. Since the \textbf{Do You See Me} dataset contains both multiple-choice (textual) and free-response (numeric) questions, accuracy is used as a measure of performance.

\newpage
\section{Joint-Perception and Reasoning Dataset Statistics}
\label{sec:appendix-prelim-dataset-stats}
\begin{table}[h!]
\centering
\caption{Distribution of examples across MathVista and LogicVista. 
“IQ/Logic Qs.” refers to pattern-based or spatial reasoning questions.}
\begin{tabular}{c|c|c|c}
\hline
Dataset & Original Size & IQ/Logic Qs & Selected \\
\hline
$MathVista_{mini}$ (IQtest) & 1000 & 37 & 15 \\
$LogicVista$ (Diagrams) & 448 & 223 & 135 \\
\hline
\end{tabular}
\label{tab:vr-benchmark-dist}
\end{table}

\newpage
\section{Do You See Me - Additional Details}
\label{sec:appendix-do-you-see-me}

\noindent The \textbf{Do You See Me} benchmark is structured into seven distinct subtasks, each meticulously designed to evaluate specific facets of visual perception in Multimodal Large Language Models (MLLMs), as detailed in Section 4.1 of the main paper. For subtasks presented in a 3D setting, a consistent set of five basic geometric shapes is utilized: \textit{sphere, cube, cone, cylinder,} and \textit{torus}. A critical parameter for modulating difficulty in 3D tasks where object occlusion is relevant—specifically Shape Discrimination and Joint Shape-Color Discrimination—is the visibility factor, $\beta_{occ} \in [0, 1]$. This factor dictates the target percentage of an object's surface area that must be visible from the camera, enabling systematic control over task complexity, ranging from fully visible objects ($\beta_{occ}=1$) to various degrees of occlusion (e.g., $\beta_{occ}=0.7$ signifies 70\% visibility). For other 3D subtasks, namely Visual Form Constancy (3D), Letter Discrimination (3D), and Visual Spatial (3D), occlusion is not a variable factor, and $\beta_{occ}$ is effectively maintained at 1.

\noindent\textbf{Shape Discrimination (2D and 3D):} This subtask evaluates an MLLM's proficiency in identifying and counting specific shapes within a composite visual scene.

\noindent\textit{2D Setting:} Seven fundamental geometric shapes are employed: \textit{rectangle, triangle, circle, pentagon, hexagon, octagon,} and \textit{star}, each rendered with solid black borders and transparent interiors. The complexity is systematically varied using three control parameters: the number of unique shape types present ($S$), the maximum number of instances permitted for each shape type ($S_{I}$), and a separation factor ($d_{sep}$). To ensure distinct object boundaries for certain conditions, non-overlapping placements ($d_{sep} \ge 0$) are achieved using the Separating Axis Theorem (SAT), which maintains a minimum distance $d_{sep}$ between any two shapes. Conversely, negative values for the separation factor ($d_{sep} < 0$) allow for controlled degrees of overlap, where shapes can interpenetrate by up to $|d_{sep}|$ units. Each generated 2D image is accompanied by a counting question (e.g., ``How many circles are in the image?'') for which a programmatic ground truth answer is available.

\noindent\textit{3D Setting:} This version uses the standard set of five 3D shapes previously mentioned. Task difficulty is modulated by the number of unique 3D shape types ($S$), the maximum instances per shape type ($S_{I}$), and the visibility parameter ($\beta_{occ}$) which controls occlusion. Questions are structurally similar to those in the 2D setting (e.g., ``How many spheres are in the scene?'').

\noindent\textbf{Joint Shape-Color Discrimination (2D and 3D):} This subtask assesses the MLLM's ability to handle compositional queries that require the simultaneous identification of object shape and color through counting tasks.

\noindent\textit{2D Setting:} Six distinct 2D shapes (\textit{star, triangle, pentagon, hexagon, octagon, cross}) are used in conjunction with eight standard colors (\textit{red, green, blue, orange, purple, black, gray, yellow}). To prevent ambiguity in shape-color binding, all shapes are rendered without overlap. The difficulty level is primarily controlled by two parameters: the number of unique shape types ($S$) and the number of unique colors ($C$) present in the image. An example query is: ``Count all red triangles''.

\noindent\textit{3D Setting:} The standard five 3D shapes are utilized, rendered with the same eight colors available in the 2D version. Difficulty is adjusted through three main parameters: the number of unique 3D shape types ($S_{3D}$), the maximum number of instances allowed for any unique shape-color pair ($I_{sc}$), and the visibility parameter ($\beta_{occ}$) controlling occlusion. Questions require joint discrimination, such as: ``How many red spheres are present in the scene?''.

\noindent\textbf{Letter Discrimination (2D and 3D):} This subtask is designed to test an MLLM's capability to recognize textual characters presented under various conditions.

\noindent\textit{2D Setting:} Characters are rendered as patterns within a $5 \times 7$ grid of LED-style blocks. The difficulty of letter identification is manipulated by three parameters: the block spacing factor ($\beta$), which adjusts the distance between constituent blocks of a letter; the color contrast ($\Delta C$) between the letter blocks and the image background; and the total number of letters ($N$) present in the stimulus.

\noindent\textit{3D Setting:} Letters are visually constructed by strategically arranging a collection of simpler 3D primitive shapes—specifically \textit{spheres, cubes,} or \textit{cylinders}—such that they collectively form the appearance of a target letter when viewed from the camera's perspective. The modulation of difficulty in this setting is achieved by varying the type of 3D primitive used for construction, the size of these primitives, the spacing between them, and the total number of distinct letters ($N$) displayed. A typical question for this task is: ``What letter(s) can you identify in this scene?''.

\noindent\textbf{Visual Form Constancy (2D and 3D):} This subtask challenges the MLLM to recognize a target pattern even after it has undergone geometric transformations or substitutions, requiring the model to identify the correct match from several options.

\noindent\textit{2D Setting:} A target pattern is first constructed using simple 2D primitives such as a \textit{circle, square, line,} or \textit{triangle}. Alongside the target, three distractor variants are generated by applying transformations controlled by a shape substitution factor ($ssf$), a scaling factor ($\alpha$), and a rotation factor ($\theta_r$). The MLLM's task is to select the one option from four (target + three variants) that perfectly matches the original target's arrangement.

\noindent\textit{3D Setting:} This version evaluates the MLLM's ability to discern and compare the orientations of 3D objects. Two groups of objects are presented: a "left group" (target) and a "right group" (comparison). Both groups are initially identical and composed of the standard 3D shapes, with the right group spatially offset from the left. In each instance, a rotation (by angle $\theta_r$) may or may not be randomly applied to the entire right group. If the right group is rotated, its constituent shapes' orientations will not match those of the left group (leading to a "no" ground truth answer); otherwise, they will match ("yes" answer). The MLLM must respond to the question: ``Do all the shapes in the left group and right group have the same orientations?''.

\noindent\textbf{Visual Spatial (2D and 3D):} This task evaluates the MLLM's understanding of object positions and their spatial relationships within structured layouts.

\noindent\textit{2D Setting:} The stimuli consist of one or more $H \times W$ grids. Each cell within these grids contains one of three basic shapes—\textit{circle, square,} or \textit{triangle}—which can be rendered either as solid black or merely outlined. Key parameters controlling the scene complexity include the grid dimensions ($D_{grid}$) and the number of grids ($G$) presented. Queries require the MLLM to locate and count shapes based on their spatial position relative to a reference coordinate provided in the question (e.g., ``How many solid circles are above the triangle in row 3, column 2?'').

\noindent\textit{3D Setting:} In this version, a grid-like structure is populated using the standard five 3D shapes. Analogous to the 2D setting, the MLLM is required to count 3D shapes based on their positions relative to a specified reference coordinate within the grid. An example query illustrates this: ``The image consists of a grid like layout with multiple 3D shapes, starting from the cylinder at position (row 2, column 1), how many tori are there to the right of it in the same row?''.

\noindent\textbf{Visual Figure-Ground (2D):} This subtask builds upon the visual form constancy framework by incorporating distracting background elements, thereby challenging the MLLM to distinguish a target pattern from its surroundings. The complexity is primarily controlled by two parameters: the number of shapes ($N$) composing the target pattern and any distractor patterns, and the background density factor ($bdf$), which dictates the quantity of visual noise introduced into the scene. The MLLM's objective is to identify the target pattern among several candidates, despite the presence of this visual noise. As noted in the main paper (Section 4.1), a distinct 3D version for Visual Figure-Ground is not included because the photorealistic 3D rendered scenes used in other 3D subtasks inherently assess this perceptual ability.

\noindent\textbf{Visual Closure (2D):} This subtask assesses an MLLM's ability to mentally complete a partially obscured or incomplete shape and match it to its corresponding complete form. Seven basic 2D shapes are used as targets: \textit{capsule, star, hexagon, circle, pentagon, rectangle,} and \textit{triangle}. For each trial, one shape is selected as the complete target. An incomplete version of this target is created by removing some of its edges. Additionally, three "noisy" distractor options are generated by taking the incomplete target and applying distortions to its vertex positions. The MLLM is presented with the complete target and four options (the correctly incomplete shape and the three distractors) and must identify which of the incomplete options would correctly form the target if its missing parts were filled in. This subtask is exclusively 2D; a 3D equivalent is not provided due to the inherent difficulties and ambiguities in judging 3D shape closure from a single, static camera viewpoint, as mentioned in Section 4.1 of the main paper.

\newpage

\section{Do You See Me - Control Parameter Details}
\label{sec:appendix-do-you-see-me-control-parameters-det}

\begin{table}[h!]
\centering
\caption{Control parameters and question types for subtasks\textsuperscript{\dag}.}
\resizebox{\linewidth}{!}{
\begin{tabular}{|c|c|c|c|c|} 
\hline
\textbf{\shortstack{Division}} & \textbf{\shortstack{Subdivision}} & \textbf{\shortstack{Setting(s)}} & \textbf{\shortstack{2D Control\\Parameters}} & \textbf{\shortstack{Dataset\\Statistics (2D)}} \\
\hline
\multirow{15}{*}{Visual Discrimination} & \multirow{4}{*}{Shape Discrimination} & \multirow{4}{*}{2D \& 3D} & 
Number of Shapes: $S \in [3, 7]$ & \multirow{4}{*}{\begin{tabular}[c]{@{}c@{}}Unique Images: 241\\ Questions: 241\\ (Integer)\end{tabular}} \\
& & & Instances per Shape: $S_I \in [3, 6, 10]$ & \\
& & & Overlap Factor: $\alpha \in [-40, -30, -20, 10]$ & \\
& & & & \\ %
\cline{2-5}
& \multirow{3}{*}{Joint Shape-Color} & \multirow{3}{*}{2D \& 3D} & 
Number of Shapes: $S \in [2, 4, 6]$ & \multirow{3}{*}{\begin{tabular}[c]{@{}c@{}}Unique Images: 90\\ Questions: 408\\ (Integer)\end{tabular}} \\
& & & Number of Unique Colors: $C \in [2, 4, 6]$ & \\
& & & & \\ %
\cline{2-5}
& \multirow{4}{*}{Letter Discrimination} & \multirow{4}{*}{2D \& 3D} &
Number of Letters: $N \in [1, 5, 9]$ & \multirow{4}{*}{\begin{tabular}[c]{@{}c@{}}Unique Images: 135\\ Questions: 135\\ (Text)\end{tabular}} \\
& & & Foreground-Background Contrast: $\Delta C \in [1, 2, 3]$ & \\
& & & Block Size: [0.04, 0.08, 0.1] & \\
& & & & \\ %
\cline{2-5}
& \multirow{4}{*}{Form Constancy} & \multirow{4}{*}{2D \& 3D} &
Shape Substitution Factor : $ssf \in [0, 1]$ & \multirow{4}{*}{\begin{tabular}[c]{@{}c@{}}Unique Images: 270\\ Questions: 270\\ (MCQ)\end{tabular}} \\
& & & Scaling Factor: $\alpha \in [0.8, 1.1, 1.4]$ & \\
& & & Rotation Factor: $\theta_r\degree \in [5, 25, 50]$ & \\
& & & Aspect Ratio: $\beta \in [0.8, 1.1, 1.4]$ & \\
\hline
\multirow{3}{*}{Visual Spatial} & \multirow{3}{*}{Spatial Grids} & \multirow{3}{*}{2D \& 3D} &
Grid Dimension: $D \subseteq [3, 6, 9] \times [3, 6, 9]$ & \multirow{3}{*}{\begin{tabular}[c]{@{}c@{}}Unique Images: 270\\ Questions: 806\\ (Integer)\end{tabular}} \\
& & & Number of Grids: $G \in [1, 3, 5]$ & \\
& & & & \\ 
\hline
\multirow{4}{*}{Visual Figure-Ground} & \multirow{4}{*}{N.A.} & \multirow{4}{*}{2D} &
Number of Shapes: $N \in [2, 6, 10]$ & \multirow{4}{*}{\begin{tabular}[c]{@{}c@{}}Unique Images: 90\\ Questions: 90\\ (MCQ)\end{tabular}} \\
& & & Background Density Factor: $bdf \in [0.1, 0.3, 0.5]$ & \\
& & & & \\ 
& & & & \\
\hline
\multirow{4}{*}{Visual Closure} & \multirow{4}{*}{N.A.} & \multirow{4}{*}{2D} &
Number of Full Edges to Remove: $ k \in [1, 3]$ & \multirow{4}{*}{\begin{tabular}[c]{@{}c@{}}Unique Images: 166\\ Questions: 166\\ (MCQ)\end{tabular}} \\
& & & Number of Partial Edges to Remove: $ l \in [1, 3]$ & \\
& & & Number of Edges to Distort: $ m \in [1, 3]$ & \\
& & & Distortion Factor: $ \delta \in [0.1, 0.12, 0.14]$ & \\
\hline
\end{tabular}}
\label{tab:visual-perception-constraints}
\textsuperscript{\dag} \tiny For tasks with 3D versions, 3D-specific counts and parameters are in \cref{sec:appendix-do-you-see-me-control-parameters-det}.
\end{table}

\begin{table}[h!]
\centering
\caption{Control parameters and question types for 3D subtasks in Do You See Me.}
\resizebox{\linewidth}{!}{
\begin{tabular}{|c|c|c|c|c|} 
\hline
\textbf{\shortstack{Division}} & \textbf{\shortstack{Subdivision}} & \textbf{\shortstack{Setting(s)}} & \textbf{\shortstack{3D Control\\Parameters}} & \textbf{\shortstack{Dataset\\Statistics (2D)}} \\
\hline
\multirow{15}{*}{Visual Discrimination} & \multirow{4}{*}{Shape Discrimination} & \multirow{4}{*}{2D \& 3D} & 
Number of Shapes: $S \in [1, 2, 3, 4, 5]$ & \multirow{4}{*}{\begin{tabular}[c]{@{}c@{}}Unique Images: 120\\ Questions: 120\\ (Integer)\end{tabular}} \\
& & & Instances per Shape: $S_I \in [1, 2, 3]$ & \\
& & & Occlusion factor: $\beta_{occ} \in [0.7, 0.8, 0.9, 0.99]$ & \\
& & & & \\ %
\cline{2-5}
& \multirow{3}{*}{Joint Shape-Color} & \multirow{3}{*}{2D \& 3D} & 
Number of Shapes: $S \in [1, 2, 3, 4, 5]$ & \multirow{3}{*}{\begin{tabular}[c]{@{}c@{}}Unique Images: 120\\ Questions: 120\\ (Integer)\end{tabular}} \\
& & & Instances per Shape: $S_I \in [1, 2, 3]$  & \\
& & & Occlusion factor: $\beta_{occ} \in [0.7, 0.8, 0.9, 0.99]$ & \\ %
\cline{2-5}
& \multirow{4}{*}{Letter Discrimination} & \multirow{4}{*}{2D \& 3D} &
Number of Letters: $N \in [1, 2]$ & \multirow{4}{*}{\begin{tabular}[c]{@{}c@{}}Unique Images: 96\\ Questions: 96\\ (Text)\end{tabular}} \\
& & & Shape Size: $\mathcal{S} \in [0.05, 0.08, 0.11, 0.14]$ & \\
& & & Inter-Shape Spacing:$\mathcal{S}_{spacing} \in [0.4, 0.5]$ & \\
& & & & \\ %
\cline{2-5}
& \multirow{4}{*}{Form Constancy} & \multirow{4}{*}{2D \& 3D} &
Rotation : $\theta_r \in [5, 7.5, 10, 12.5, 15]$ & \multirow{4}{*}{\begin{tabular}[c]{@{}c@{}}Unique Images: 80\\ Questions: 80\\ (MCQ)\end{tabular}} \\
& & & Number of Shapes $S \in [1, 2, 3, 4]$ & \\
& & & & \\
& & & & \\
\hline
\multirow{3}{*}{Visual Spatial} & \multirow{3}{*}{Spatial Grids} & \multirow{3}{*}{2D \& 3D} &
Grid Dimension: $D \subseteq [2, 3, 4, 5] \times [2, 3, 4, 5]$ & \multirow{3}{*}{\begin{tabular}[c]{@{}c@{}}Unique Images: 80\\ Questions: 80\\ (Integer)\end{tabular}} \\
& & & Number of Grids: $G \in [1]$ & \\
& & & & \\ 
\hline
\end{tabular}}
\label{tab:visual-perception-constraints-3D}
\vspace{-8mm}

\end{table}

\newpage
\section{Human Performance Benchmarking}
\label{sec:appendix-human-benchmarking}
We recruited 15 participants (11 men, 4 women) for the human performance benchmarking study. The primary selection criterion was that participants had no history of vision-related disorders. All participants were between 25 and 35 years of age. Each of the twelve subtask evaluations took approximately 10 minutes per participant to complete, resulting in a total active testing time of about 120 minutes per participant. To mitigate mental fatigue, these evaluations were conducted over two days. As a token of appreciation for their participation, subjects received a food voucher equivalent to the local minimum wage for two hours. To assess inter-rater reliability, we calculated Fleiss's Kappa. The agreement on task correctness was 0.796, indicating substantial agreement among participants. Additionally, the low standard deviations for human performance, when contrasted with the large gap to mean MLLM performance, demonstrate a clear and significant difference.

\begin{figure}[htbp]
\centering
\begin{subfigure}{0.43\textwidth}
    \centering
    \includegraphics[width=\linewidth]{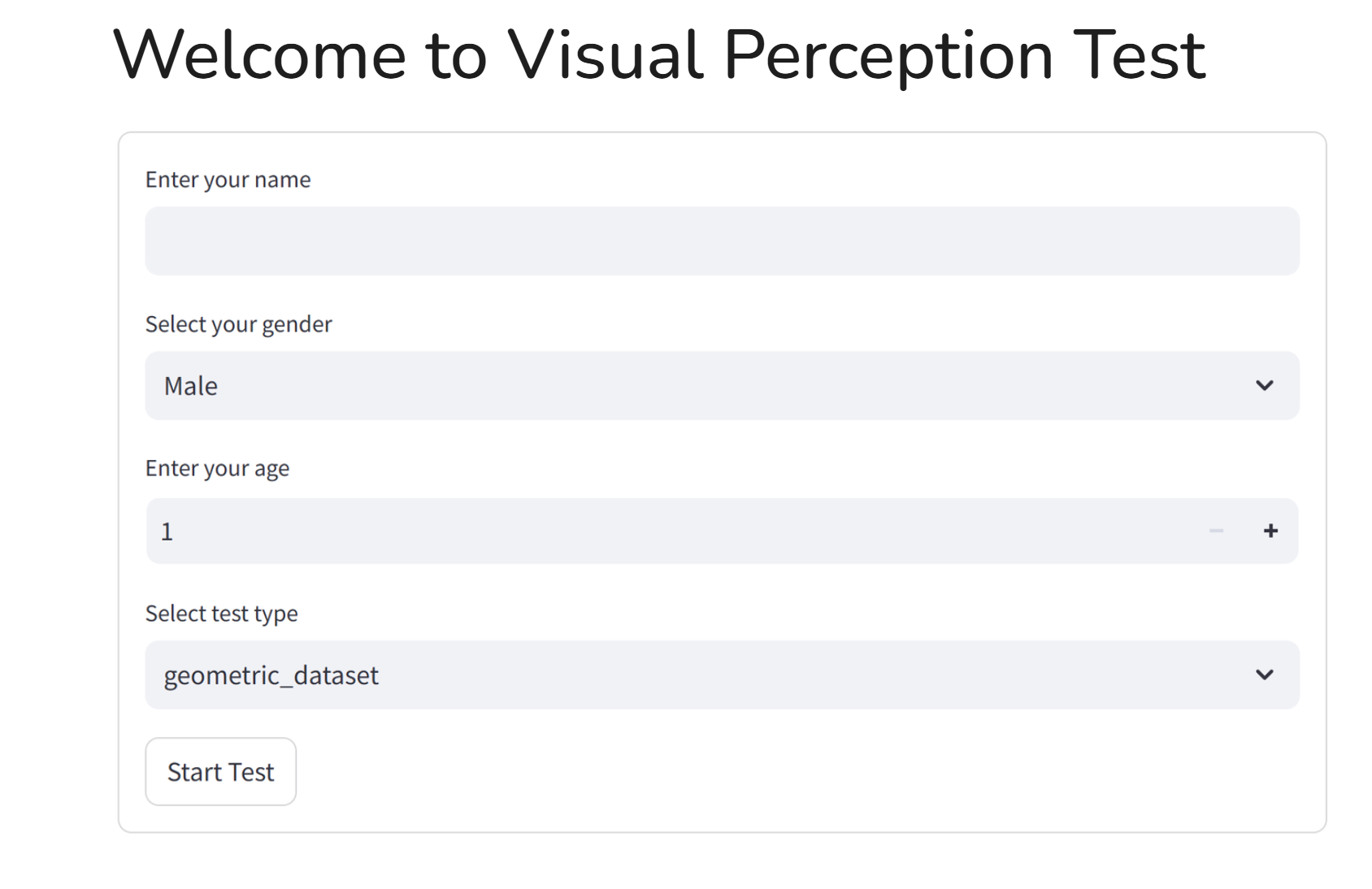}
    \caption{\small Information Collected in the Test}
    \label{}
\end{subfigure}
\hspace{10pt}
\begin{subfigure}{0.43\textwidth}
    \centering
    \includegraphics[width=\linewidth]{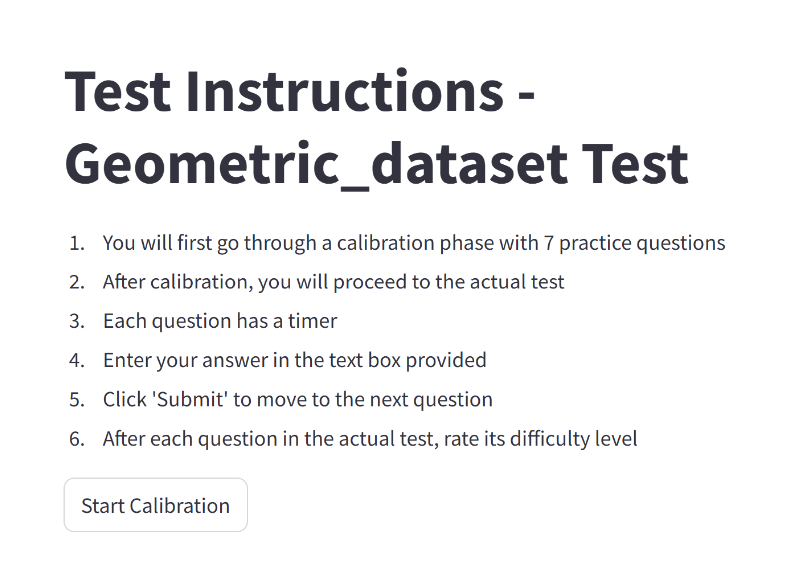}
    \caption{\small Calibration Instructions}
    \label{}
\end{subfigure}
\caption{Human Performance Benchmarking}
\label{}
\end{figure}

\begin{figure}[h!]
    \centering
    \includegraphics[width=0.5\linewidth]{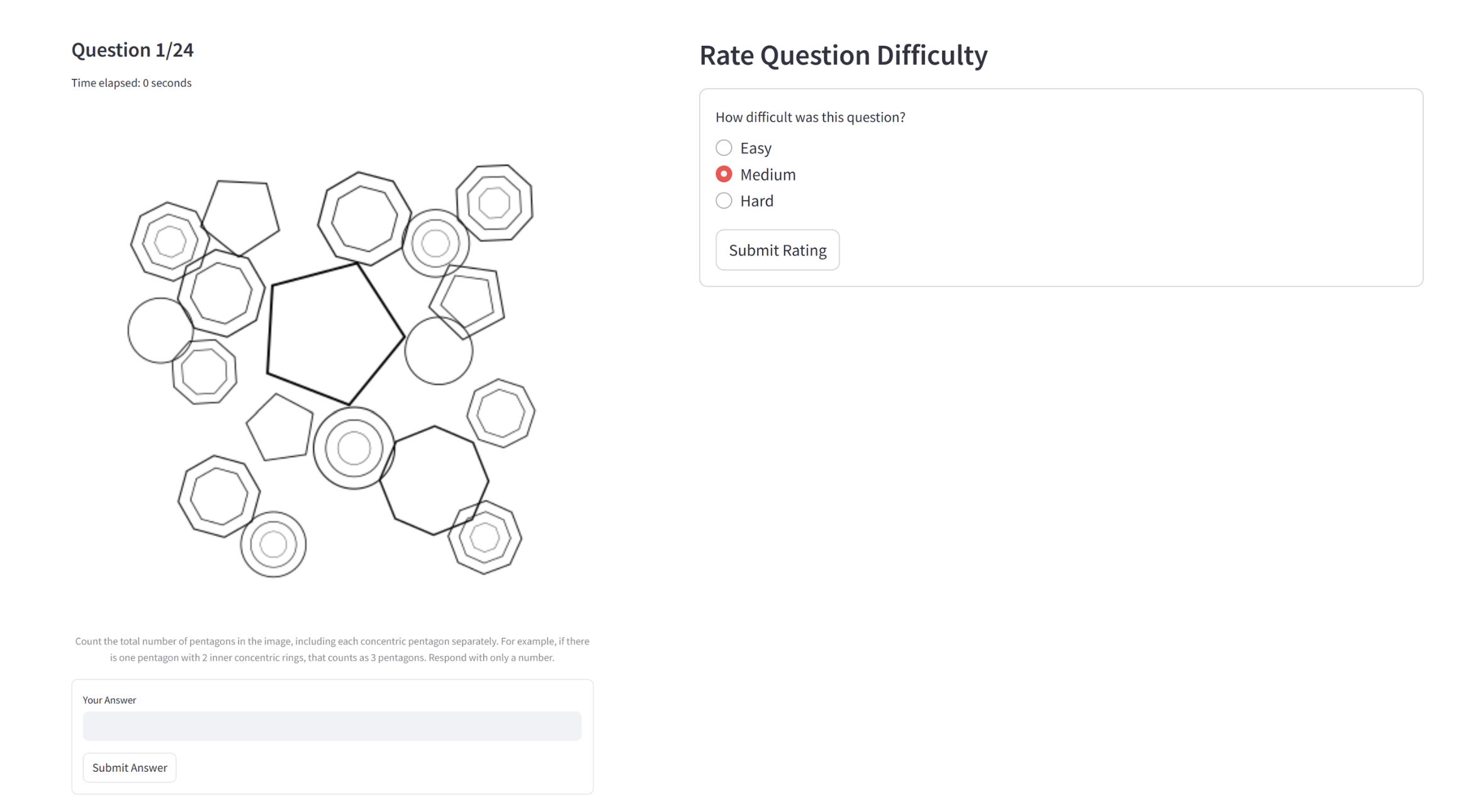}
    \caption{Actual Test and Human Difficulty Rating Collection}
    \label{fig:enter-label}
\end{figure}

\section{Joint Perception-Reasoning Dataset}
\label{sec:appendix}

\subsection{Correct Reasoning Incorrect Perception}
\mytcolorbox{1}{
    \analysisimage{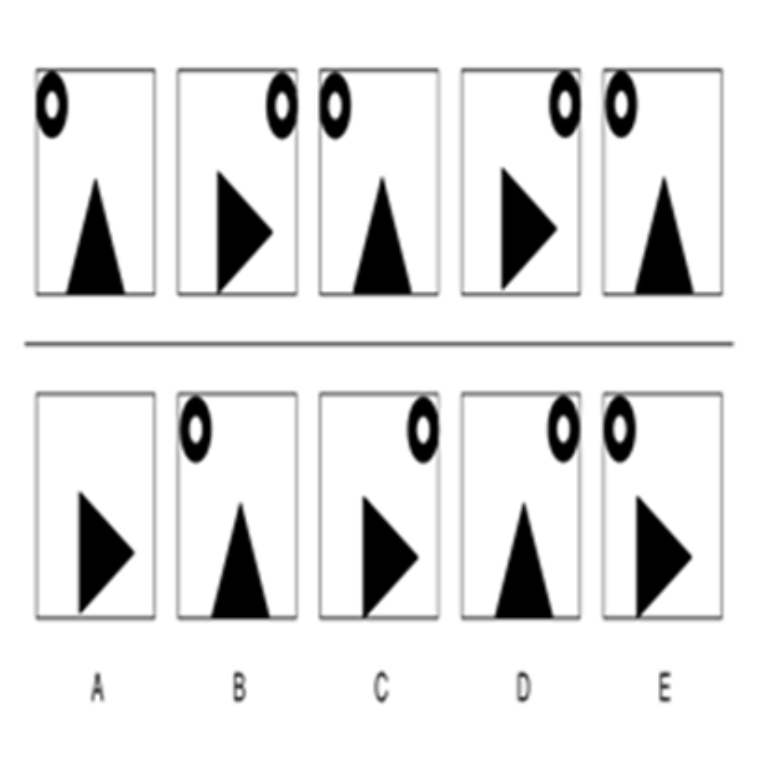}{}
    
    \textbf{Reasoning Question: } Which of the boxes comes next in the sequence? Select answers from A-E. Explain the reasoning behind the answer.
    
    \textbf{Reasoning Answer (GT): } C
    
    \textbf{Reasoning Explanation: } Triangle alternates from pointing up, to pointing right with each turn. Doughnut shape alternates from left corner to right corner with each turn.

    \textbf{Claude Sonnet-3.5 Answer: }  Let's solve this step by step:

Looking at the top sequence, I notice a pattern of shapes and dots:

Single triangle, single dot above
Single triangle, single dot above
Single triangle, two dots above
Single triangle, single dot above
Single triangle, single dot above
Also notice that the triangles alternate between pointing left and right

The pattern appears to be:

The triangles alternate direction
The dots follow a pattern of: 1, 1, 2, 1, 1, 2, ...
Following this pattern, the next box should have:

A triangle pointing right
Two dots above (because we're at the point in the sequence where two dots should appear)
Looking at the options A through E: Option C shows a triangle pointing right with two dots above it.

Therefore, the answer is C.

The reasoning is based on both the alternating direction of the triangles and the repeating pattern of dots (1, 1, 2).
    
    \textbf{VLM Final Answer: } YES

    \textbf{Visual Perception Question: } Using visual inspection, count the number of black dots on the top-left side of the box where the triangle present in the box is resting on its side and not the vertex.

    \textbf{Visual Perception Answer (GT): } 4

    \textbf{Claude Sonnet-3.5 Answer: } Let me help you count the black dots as requested.

    Looking at the boxes where the triangle is resting on its side (not on its vertex), we need to look at the top-left portion of those boxes.
    
    The triangles resting on their sides (not on vertex) appear in positions B, D, and E in the top row and positions A and E in the bottom row.
    
    Among these:
    
    Top row: \textcolor{red}{B has 2 dots, D has 2 dots, E has 1 dot
    Bottom row: A has 0 dots, E has 2 dots} (\textbf{incorrect perception})
    Therefore, there are a total of 7 black dots when counting only the top-left portions of boxes where the triangle is resting on its side.
    
    \textbf{Analysis:} The visual perception question shows how model fails answering the question on number of dots on top-left side. The same failure occurs in answering the visual reasoning question. 
}

\newpage
\mytcolorbox{2}{
    \analysisimage{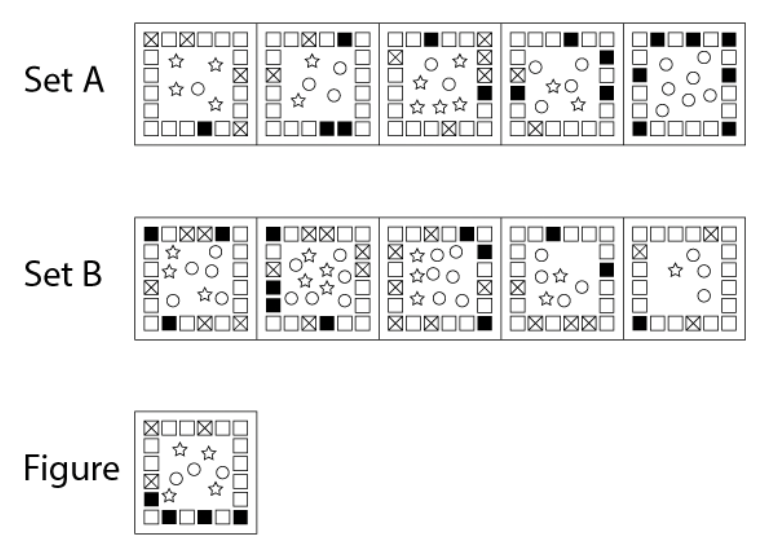}{}
    
    \textbf{Reasoning Question: } Which set does the Figure belong to? Select from A, B, and C. (A) Set A (B) Set B (C) Neither set A nor set B. Explain the reasoning behind the answer.
    
    \textbf{Reasoning Answer (GT): } B
    
    \textbf{Reasoning Explanation: } Set A: The total number of crossed boxes is equal to the total number of stars. The total number of shaded boxes is equal to the total number of circles. Set B: The total number of crossed boxes is equal to the total number of circles. The total number of shaded boxes is equal to the total number of starts. Figure: There are four shaded boxes and four stars. There are three crossed boxes and three circles.

    \textbf{Claude Sonnet-3.5 Answer: }  The Figure belongs to (B) Set B.

Reasoning:

Looking at the pattern in both sets, we need to focus on the arrangement of black squares and circles in the cards.

Key observations about the Figure:

It has circles arranged in a specific pattern
It has three black squares at the bottom
The overall layout is similar to other cards in both sets
When comparing with Set A and Set B:
Set A's cards generally have black squares distributed throughout different positions
Set B's cards consistently show black squares grouped at the bottom of each card, similar to the Figure
The circle patterns in Set B also appear more similar to the Figure's arrangement
The decisive factor is the positioning of the black squares:
The Figure shows three black squares aligned at the bottom
This arrangement is consistent with the pattern seen in Set B, where black squares are consistently placed at the bottom of the cards
This pattern is not characteristic of Set A
Therefore, based on the consistent pattern of black square placement at the bottom and similar circle arrangements, the Figure belongs to Set B.
    
    \textbf{MLLM Final Answer: } Correct

    \textbf{Visual Perception Question: } In Set B of the diagram, looking at the second square from the left, how many stars (?) are inside that square?

    \textbf{Visual Perception Answer (GT): } 4

    \textbf{Claude Sonnet-3.5 Answer: } Looking at Set B, in the second square from the left, I can count 2 stars within that square.

    \textbf{MLLM Final Answer: } Incorrect

    \textbf{Analysis:} The that number of stars within the box are linked to correctly the reasoning question and Claude Sonnet-3.5 fails to answer the count correctly. Further, from the reasoning chain, it is clear that it makes a reasoning error that black squares are grouped/ungrouped in Set A and B, and answers the reasoning question on this basis. This shows a reasoning error in its response, however, the final answer given is correct. Additionally, there is visual hallucination about distribution of black sqaures as well.

}

\newpage

\subsection{Qualitative Analysis of Reasoning Chain Errors}

\mytcolorbox{1}{
    \analysisimage{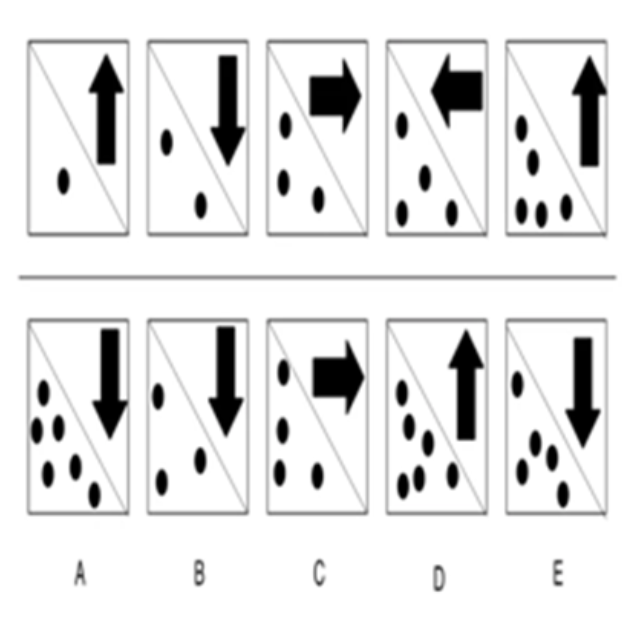}{}
    
    \textbf{Reasoning Question: } Which of the boxes comes next in the sequence? Select from A-E

    \textbf{Reasoning Answer (GT): } A

    \textbf{Reasoning Explanation: }

    \textbf{GPT4o Answer: } To determine which box comes next in the sequence, observe the direction of the arrows and the distribution of dots in each box. 1. The direction of arrows alternates from pointing up to down in every subsequent box. The sequence starts with an arrow pointing up. 2. The number of dots increases incrementally by one in each subsequent box. The first box has one dot, the second has two, the third has three, and so on. Following this pattern:
    
    - The first box in the sequence has one dot and an arrow pointing up.
    
    - The second box has two dots and an arrow pointing down.
    
    - The third box has three dots and an arrow pointing right.
    
    - \textcolor{red}{The fourth box has four dots and an arrow pointing down.} \textbf{(perception error)}
    
    - The fifth box has five dots and an arrow pointing up.
    
    Given these patterns, the next box should have an arrow pointing down and six dots. Among the choices A to E:
    
    - Choice A fits with an arrow pointing down and six dots.
    
    Therefore, the correct answer is box A.

    \textbf{VLM Final Answer: } Correct.

    \textbf{Key Observation: } Correct Final Answer But presence of Incorrect Visual Perception in Reasoning Chain. 

    \textbf{Analysis:}  GPT4o makes a visual perception error on the fourth box, the pattern was up, down, right, left, up. Whereas, GPT-4o determines that the pattern is up, down, right, down, up. \emph{This is an example of visual perception error in reasoning chain even while the model gives a correct final answer.}
}

\newpage

\mytcolorbox{2}{
    \analysisimage{images/appendix/gpt4o_2.png}{}
    
    \textbf{Reasoning Question: } Which of the boxes comes next in the sequence? Select from A-E

    \textbf{Reasoning Answer (GT): } C

    \textbf{Reasoning Explanation: }

    \textbf{GPT4o Answer: }  To determine which box comes next in the sequence, we need to observe the patterns of the shapes and their rotations in the given boxes. 
1. The position of the circle: 

   - \textcolor{red}{The circle moves through four quadrants: top-right, top-left, bottom-left, and bottom-right.} (\textbf{visual perception error: the circle never goes to bottom left or right}) 

   - Based on this pattern, after the \textcolor{red}{bottom-right} position in the fifth box, the next position should be top-right. 

    2. The rotation of the triangle: 

   - The triangle rotates clockwise: pointing up, right, down, and then left. 

   - \textcolor{red}{In the fifth box, the triangle is pointing left. Therefore, in the next box, it should point up again. }

 Given these patterns, we are looking for a box with the circle in the top-right quadrant and the triangle pointing up. 

 Option **C** shows a circle in the top-right quadrant and a triangle pointing up. This matches the patterns observed, so option C is the correct answer for the next box in the sequence. 

    \textbf{VLM Final Answer: } Correct.

    \textbf{Key Observation: } Correct final answer but \emph{presence of incorrect visual perception} in Reasoning Chain. 

    \textbf{Analysis:}  The circle goes between top-right/left. \emph{Example of visual perception error}
}

\newpage

\mytcolorbox{3}{
    \analysisimage{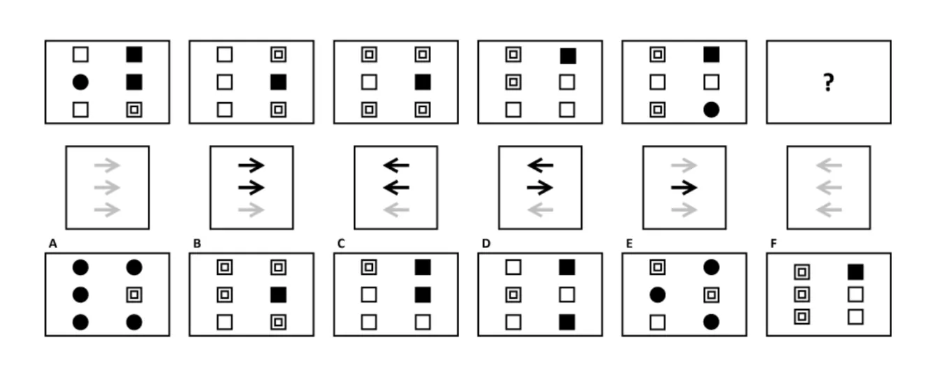}{}
    
    \textbf{Reasoning Question: }  The bottom boxes create a rule that has to be applied in the box directly above them. Select which of options A to F corresponds to the rule below the box with the question mark.. Explain the reasoning behind the answer.
    
    \textbf{Reasoning Answer (GT): } E
    
    \textbf{Reasoning Explanation: } The simple rule in this question is that the number of grey arrows in the box below must equal the number of black shaded shapes in the box above. There are three grey arrows below the missing box, so the correct answer must contain three black shapes.

    \textbf{GPT4o Answer: }  To determine which option (A to F) corresponds to the rule that should be applied above the question mark, we need to analyze the pattern in each column and how the bottom boxes affect the top boxes.

    Column Analysis:
    
    Column 1: The bottom box has arrows pointing right. The top box changes the orientation of the shapes accordingly.
    
    Column 2: The bottom box has arrows pointing left. The top box changes by mirroring or flipping the orientation.
    
    Column 3: The bottom box has inward-facing arrows, and the shapes in the top box are transformed or interchanged.
    
    Column 4: The bottom box has outward-facing arrows. It seems to involve swapping the positions of certain elements in the top box.
    
    Column 5 (Question Mark): The bottom box has left-facing arrows, implying a mirroring or flipping operation similar to column 2.
    
    Rule Application:
    
    Given the bottom box under the question mark has left-facing arrows, we should select an option that mirrors or flips the elements horizontally compared to the option from Column 4.
    
    Options Evaluation:
    
    Option E is a horizontally flipped version of the top box in Column 4, making it the correct answer.
    Therefore, the correct option corresponding to the rule below the box with the question mark is Option E.
    
    \textbf{VLM Final Answer: } Correct

    \textbf{Key Observation: } Correct answer, incorrect reasoning.

    \textbf{Analysis:} The simple rule here is the number of grey arrows in the box below must equal the number of black shaded shapes in the box above. \emph{However, the model reasons about changing orientations based on the direction of arrows.} However, even with wrong reasoning GPT-4o reaches the correct answer.
}

\newpage

\mytcolorbox{4}{
    \analysisimage{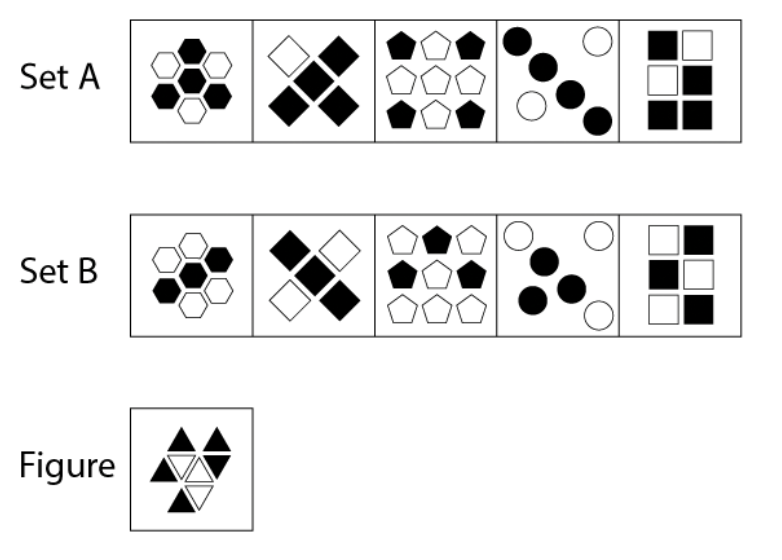}{}
    
    \textbf{Reasoning Question: } Which set does the Figure belong to? Select from A, B, and C. (A) Set A (B) Set B (C) Neither set A nor set B. Explain the reasoning behind the answer.

    \textbf{Reasoning Answer (GT): } C
    
    \textbf{Reasoning Explanation: } Set A: Each box contains four shaded shapes. Set B: Each box contains three shaded shapes. Figure: The Figure contains five shaded shapes.

    \textbf{GPT4o Answer: }  The figure belongs to (C) Neither set A nor set B.
    Reasoning:
    
    Set A:
    
    In Set A, the figures alternate between filled and empty shapes, or there is a symmetry of arrangement.
    Set B:
    
    In Set B, there is a symmetrical arrangement as well, but the colors (filled and empty) alternate in a different manner than Set A.
    Figure Analysis:
    
    The given figure is composed of triangles, both filled and empty, arranged in a different pattern from any of the figures in Set A or Set B.
    It does not follow the alternation pattern or symmetry rules observed in either set.
    Therefore, the figure does not belong to Set A or Set B.
    
    \textbf{VLM Final Answer: } Correct

    \textbf{Key Observation: } Correct answer, incorrect reasoning.

    \textbf{Analysis:} GPT4o reasons that there is a different alternate pattern to fill shapes in Set A and B, and the new shape has both these patterns, hence cannot be alloted to any one set. However, the simple rule here is that 3 shapes are shaded in set B, and 4 shapes are shaded in shape A. Thus, while the model reasons incorrectly, it arrives at the right option.

}
\newpage

\section{MLLM Prompts}
\label{sec:appendix-mllm-prompts}

\begin{tcolorbox}[colback=gray!10, colframe=gray!80,
title=Expert LLM Evaluator]
    You are given a question, a long-form answer generated by a vision-language model, and the ground-truth answer. The model's answer may contain reasoning or explanation, but always ends with a specific count or choice.

    Your task is to extract **only the final integer answer** from the model's response and output it in the format:
    <answer>X</answer>
    
    If no clear answer is found, respond with:
    <answer>ERROR</answer>

    The extracted answer should match the type of the ground truth answer — whether it is a count or an option number. Note, do not directly copy over the ground-truth answer in your response. The answer should only be extracted from the long-form model response.

    **Examples**

    Question: In grid 2, starting from the white circle at position (row 1, column 8), how many squares are there left of it in the same row?  
    Ground Truth Answer: 0  
    VLM Answer: In grid 2, starting from the white circle at position (row 1, column 8), there is 1 square to the left of it in the same row.  
    <answer>1</answer>

    Question: The figure consists of a Target image... Option <your answer (choose between 1, 2, 3, or 4)>.  
    Ground Truth Answer: 4  
    VLM Answer: After analyzing the images, the third option appears most similar to the target image. Option 3.  
    <answer>3</answer>

    Question: In grid 1, starting from the white square at position (row 2, column 1), how many black objects are there up of it in the same column?  
    Ground Truth Answer: 0  
    VLM Answer: There are no black objects above it.  
    <answer>0</answer>

    Question: How many circles are present in the image?  
    Ground Truth Answer: 2  
    VLM Answer: I see that there are 2 circles and 4 triangles. Therefore, there are 2 circles in the image.
    <answer>2</answer>

    Now process the following:
   
    Question: \{question\}
    
    Ground Truth Answer: \{gt\_answer\}  
    
    VLM Answer: \{vlm\_answer\}  
\end{tcolorbox}

\begin{tcolorbox}[colback=gray!10, colframe=gray!80,
title=Example Prompts For Subtasks in Do You See Me]
    \textbf{Visual Figure Ground:} The figure consists of a Target image, which is embedded in some background noise. Out of the four given options, your task is to pick the option which has the same figure as the target image. Respond as follows: Option your answer (choose between 1, 2, 3, or 4).
\\

    \textbf{Letter Disambiguation:} The image shows one or more letters formed by a grid of small squares. What letter(s) can you identify in this image? Please respond with only the letter(s) you see.
\\

   \textbf{ Visual Form Constancy:} The figure consists of a Target image. Out of the four given options, your task is to pick the option which has the same figure as the target image. Respond as follows: Option your answer (choose between 1, 2, 3, or 4).
\\

    \textbf{Visual Closure:} The figure consists of a target image which is complete, Out of the four given options (which are partially complete), your task is to pick the option which when completed matches the target image. Respond as follows: Option your answer (choose between 1, 2, 3, or 4).
\\

    \textbf{Visual Spatial:} In grid 5, starting from the white square at position (row 1, column 5), how many circles are there down of it in the same column?
\\

    \textbf{Color Disambiguation:} Count the number of cross's that are purple.
\\

    \textbf{Shape Discrimination:} Count the total number of stars in the image, including each concentric star separately. For example, if there is one star with 2 inner concentric rings, that counts as 3 stars. Respond with only a number.
\end{tcolorbox}

\newpage

\section{Parameter Importance}
\label{sec:appendix-parameter-importance-study}
To gain a deeper understanding of the factors driving performance variations among Multimodal Large Language Models (MLLMs) on the \textbf{2D tasks} within the ``Do You See Me'' benchmark, we conducted a detailed parameter importance study. To ascertain statistical significance, we employed Kruskal-Wallis tests, a non-parametric form of one-way ANOVA, considering a parameter's impact significant if the $p$-value fell below $0.05$.

This analysis revealed that most of the 8 MLLMs had at least one significant factor affecting their performance on any given 2D task, suggesting that models exhibiting no sensitivity to parameter changes might be resorting to random guessing. Several consistent patterns emerged. For instance, the \texttt{number\_of\_letters} was a critical determinant in \texttt{letter\_disambiguation} for \textbf{6 out of the 8 MLLMs}. A similar proportion of these models (\textbf{5 out of 8}) found \texttt{aspect\_ratio} to be a significant factor in \texttt{visual\_form\_constancy}. In \texttt{joint\_shape\_color\_discrimination} tasks, the \texttt{number\_of\_shapes} (significant for \textbf{4 out of 8 models}) and the \texttt{number\_of\_unique\_colors} (significant for \textbf{4 out of 8 models}) were frequently influential. An interesting distinction arose when comparing the open-source and closed-source MLLMs within this 8-model subset. On average, the closed-source models tended to have a higher number of significant variables affecting their performance on these 2D tasks. This suggests that while these models might achieve higher overall accuracy, their performance is also discernibly modulated by a broader range of specific input complexities within 2D contexts.

Furthermore, the study highlighted the impact of supervised fine-tuning (SFT) by comparing the QWEN2.5-VL-7B-INSTRUCT model with its fine-tuned version, QWEN2.5-VL-7B-INSTRUCT-SFT, on these 2D tasks. The SFT model demonstrated a marked increase in the number of significant variables to which it was sensitive, from four in the base model to nine in the fine-tuned version. This increased sensitivity was particularly evident in 2D tasks related to spatial reasoning, shape discrimination, and letter discrimination. For example, after fine-tuning, \texttt{instances\_per\_shape} and \texttt{number\_of\_shapes} became significant for \texttt{shape\_discrimination}, and \texttt{block\_spacing} became an additional factor for \texttt{letter\_disambiguation}. This observation points that higher number of significant parameters indicates lower randomness, and thus better accuracy on the Do You See Me benchmark.

\newpage
\section{Finetuning Details}
\label{sec:appendix-mllm-finetuning-details}
To assess the impact of supervised finetuning (SFT) on the visual perception capabilities of MLLMs, specifically concerning the tasks presented in the \textit{DoYouSeeMe} benchmark, we finetuned the Qwen2.5-VL-7B-Instruct model. The finetuning process was conducted using the Llama Factory framework on the \textit{DoYouSeeMe-Train} dataset, which comprises approximately 67,000 benchmark-conformant image-text pairs.

We employed LoRA (Low-Rank Adaptation) for efficient finetuning, targeting the vision encoder and the LLM with a LoRA rank of $8$. The training was conducted for $5$ epochs with a per-device training batch size of $8$ and $8$ gradient accumulation steps, resulting in an effective batch size of $64$. A learning rate of \textit{$1.0e^{-4}$} was used with a cosine learning rate scheduler and a warmup ratio of $0.1$. The training utilized bf16 precision. Evaluation was performed every $500$ steps on a validation set comprising 10\% of the training data.

The impact of SFT on the Qwen2.5-VL-7B-Instruct model's performance across various visual perception tasks is presented in Table~\ref{tab:sft_performance}.

\begin{table}[h!]
\centering
\caption{Performance of Qwen2.5-VL-7B-Instruct Before and After SFT [cite: 601]}
\label{tab:sft_performance}
\resizebox{\textwidth}{!}{%
\begin{tabular}{@{}lcccccccc@{}}
\toprule
Model & Average Acc. (\%) & Visual Figure Ground & Visual Spatial & Color Disamb. & Shape Disamb. & Letter Disamb. & Visual Form Const. & Visual Closure \\
\midrule
Qwen2.5-VL-7B-Instruct & 40.91 & 27.78 & 40.69 & 81.86 & 19.58 & 2.96 & 50.37 & 63.10 \\
Qwen2.5-VL-7B-Instruct-SFT & 51.75 & 41.11 & 46.15 & 60.78 & 49.58 & 13.33 & 92.96 & 75.89 \\
\midrule
Human & 94.33 & 100.0 & 92.59 & 100.0 & 100.0 & 77.77 & 98.14 & 91.66 \\
\bottomrule
\end{tabular}%
}
\end{table}

As shown in Table~\ref{tab:sft_performance}, supervised finetuning led to an improvement in the average accuracy of the Qwen2.5-VL-7B-Instruct model from 40.91\% to 51.75\%. Notable gains were observed in tasks such as Shape Disambiguation (from 19.58\% to 49.58\%) and Visual Form Constancy (from 50.37\% to 92.96\%). However, despite these improvements, the finetuned model's performance still remained significantly below human accuracy levels (average 94.33\%) across all tasks. For instance, in tasks like Letter Disambiguation, the finetuned model achieved 13.33\% compared to human performance of 77.77\%. Furthermore, in Color Disambiguation, the performance slightly decreased post-SFT (from 81.86\% to 60.78\%). This outcome suggests that while SFT on benchmark-conformant data can yield some gains, it does not drastically overcome the fundamental visual perception limitations observed in MLLMs, highlighting the need for alternative approaches or more fundamental architectural changes to enhance MLLM visual perception.

\newpage

\section{Visual Stimuli Reconstruction}
\label{sec:appendix-visual-stimuli-reconstruction}
To further probe the internal visual understanding of Multimodal Large Language Models (MLLMs) and to explore alternative modalities for task resolution, we investigated an intermediate step of visual stimuli reconstruction. In this approach, for each 2D visual task within the \textit{DoYouSeeMe} benchmark, the MLLM was first prompted to convert the given visual stimulus into its constituent Scalable Vector Graphics (SVG) representation. In a subsequent, distinct step, the MLLM was then asked to answer the associated perception question, relying solely on the SVG code it had previously generated. The primary motivations for this exploration were twofold: first, to transform the vision-text task into a text-only one by leveraging the LLM component's text-based reasoning on the structured SVG format; and second, to perform a qualitative perceptual analysis, as the generated SVG itself serves as a tangible artifact of the MLLM's interpretation of the visual scene, offering direct insights into ``what the MLLM sees''. \textit{Note: Due to the high token output and cost of SVG generation, we sampled two images per parameter sweep for this reconstruction analysis.} 

The evaluation of MLLM performance via SVG reconstruction involved a two-step process for each 2D image-question pair. \textbf{Step 1: SVG Generation} involved providing the MLLM with the visual stimulus and a prompt such as: \texttt{``Given the image, generate the complete SVG code that accurately reconstructs its visual content.''}. \textbf{Step 2: Question Answering based on Generated SVG} followed, where the MLLM, after generating the SVG, was presented with the original perception question and explicitly instructed to answer based \textit{only} on the SVG code it had generated, using a prompt like: \texttt{``Using only the following SVG code: <SVG-Code>, answer the question: <Visual-Perception-Question>''}. This two-step methodology ensures that the question-answering performance in the SVG-mediated condition is directly dependent on the quality and accuracy of the MLLM's own SVG reconstruction.

As detailed in the main paper's Discussion section, this strategy of intermediate SVG reconstruction generally \textbf{failed to improve, and often notably degraded, task performance} compared to direct visual questioning. This quantitative outcome suggests that generating accurate and detailed SVG representations is a significant challenge for current MLLMs, with errors in SVG generation likely cascading to the question-answering phase. Table~\ref{tab:mllm_performance_svg} presents a performance comparison for GPT-4o and Claude-3.5. ``-Image'' denotes performance when answering directly from the visual stimulus, while ``-SVG'' denotes performance based on the MLLM's own generated SVG. All scores are accuracy percentages (\%).

\begin{table}[h!]
\centering
\caption{MLLM Performance: Image vs. SVG-mediated. All scores are accuracy percentages (\%).}
\label{tab:mllm_performance_svg}
\begin{tabular}{@{}lccccccc@{}}
\toprule
Model & Shape & Color & Letter & Visual & Visual & Figure- & Form \\
      & Disamb. & Disamb. & Disamb. & Spatial & Closure & Ground & Constancy \\
\midrule
GPT4o – Image & 12.50 & 77.77 & 25.92 & 25.92 & 58.33 & 11.11 & 83.33 \\
GPT4o – SVG & 29.16 & 44.44 & 11.11 & 18.51 & 41.66 & 0.00 & 33.33 \\
Claude-3.5 – Image & 41.66 & 77.77 & 7.40 & 25.92 & 54.16 & 55.55 & 94.40 \\
Claude-3.5 – SVG & 33.33 & 77.77 & 11.11 & 11.11 & 50.00 & 11.11 & 40.70 \\
\bottomrule
\end{tabular}
\end{table}

Observations from these benchmark results indicate a general performance degradation when using generated SVGs. For instance, GPT4o's Form Constancy drops from 83.33\% (Image) to 33.33\% (SVG), and Claude-3.5's Figure-Ground performance falls from 55.55\% (Image) to 11.11\% (SVG). The extent of degradation varies across tasks; Color Disambiguation for Claude-3.5 shows identical performance (77.77\%) for both Image and SVG, while for GPT4o, it drops significantly. Notably, Figure-Ground for GPT4o drops to 0\% with SVG. An interesting outlier is GPT4o's Shape Disambiguation, which increased from 12.50\% (Image) to 29.16\% (SVG). These results underscore the difficulty MLLMs face in accurately translating visual information into SVG.

\begin{figure}[h!]
    \centering
    \begin{subfigure}[b]{0.3\textwidth}
        \includegraphics[width=\textwidth]{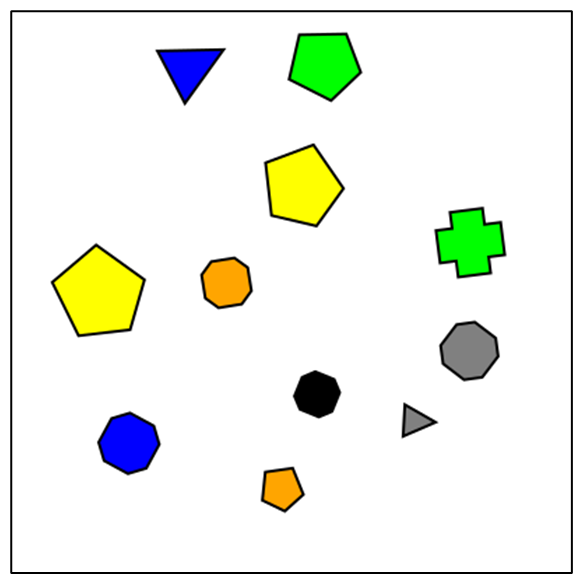}
        \caption{Original Stimulus}
        \label{fig:original_stimulus_ex1}
    \end{subfigure}
    \hfill 
    \begin{subfigure}[b]{0.3\textwidth}
        \includegraphics[width=\textwidth]{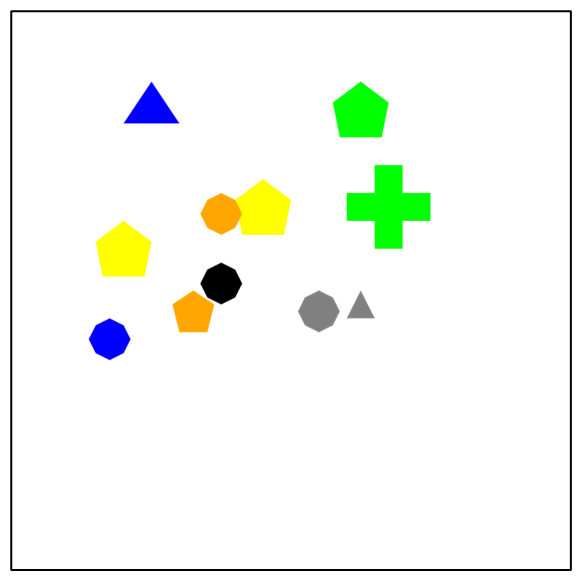}
        \caption{Claude-3.5 Generated}
        \label{fig:claude_generated_ex1}
    \end{subfigure}
    \hfill 
    \begin{subfigure}[b]{0.3\textwidth}
        \includegraphics[width=\textwidth]{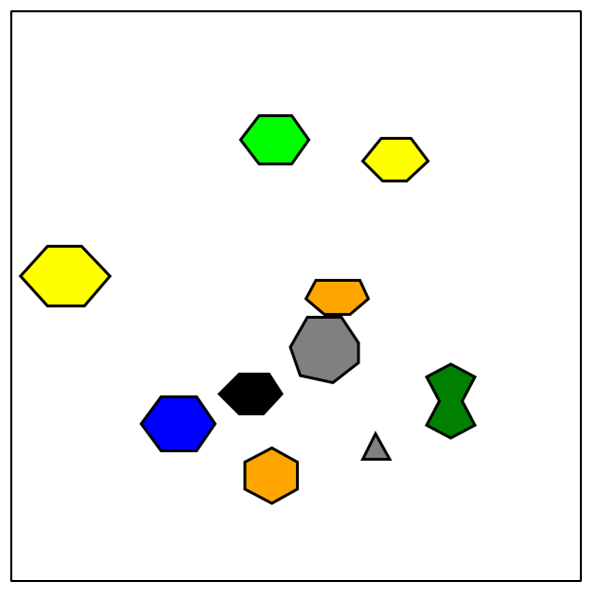}
        \caption{GPT-4o Generated}
        \label{fig:gpt_generated_ex1}
    \end{subfigure}
    \caption{Comparison of an original visual stimulus with MLLM-generated SVG representations (rendered as images). Such comparisons highlight perceptual errors made by the models.}
    \label{fig:stimuli_comparison_ex1}
\end{figure}

A detailed qualitative analysis of the SVGs generated by the MLLMs revealed several common categories of perceptual errors, providing insights into the visual attributes MLLMs struggle with:
\begin{enumerate}
    \item \textbf{Shape Inaccuracies:} Including misidentification (e.g., circle as oval) and geometric distortion (e.g., equilateral triangle as scalene).
    \item \textbf{Orientation and Positional Errors:} Incorrect rotation and inaccurate relative or absolute positioning of shapes.
    \item \textbf{Color and Fill/Stroke Errors:} Misidentified colors or incorrect representation of fill and stroke properties.
    \item \textbf{Count and Completeness Errors:} Omission of existing elements, inclusion of spurious elements, or incorrect counts of repeated shapes.
    \item \textbf{Grouping and Hierarchical Structure:} General failure to capture complex grouping or hierarchical relationships (concentric shapes).
\end{enumerate}

\begin{figure}[h!]
    \centering
    \begin{subfigure}[b]{0.3\textwidth}
        \includegraphics[width=\textwidth]{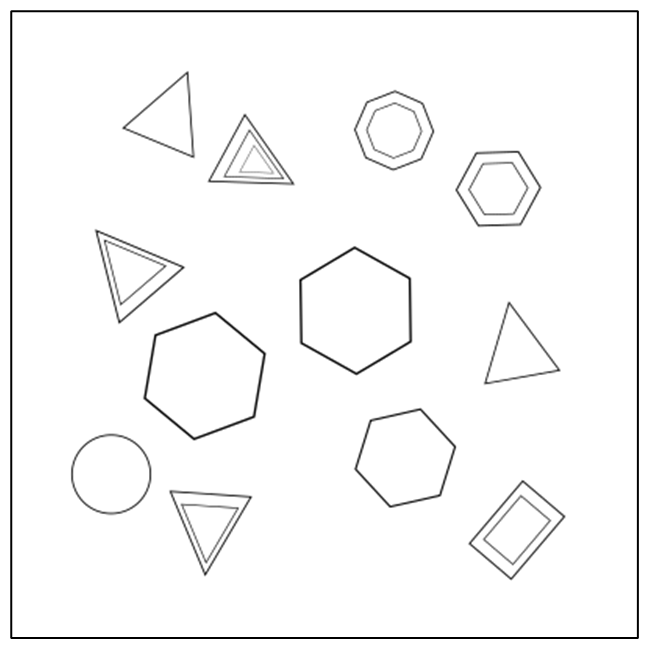}
        \caption{Original Stimulus}
        \label{fig:original_stimulus_ex2}
    \end{subfigure}
    \hfill 
    \begin{subfigure}[b]{0.3\textwidth}
        \includegraphics[width=\textwidth]{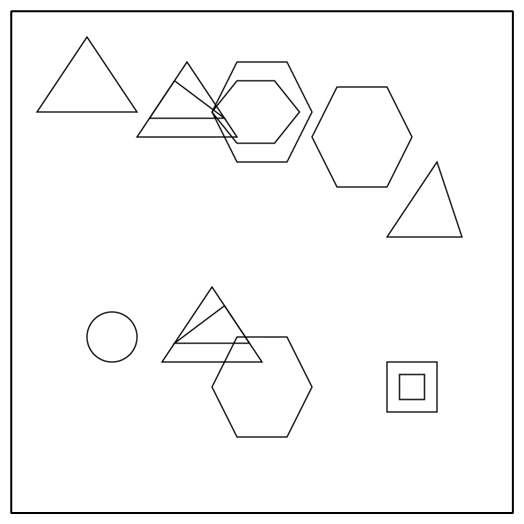}
        \caption{Claude-3.5 Generated}
        \label{fig:claude_generated_ex2}
    \end{subfigure}
    \hfill 
    \begin{subfigure}[b]{0.3\textwidth}
        \includegraphics[width=\textwidth]{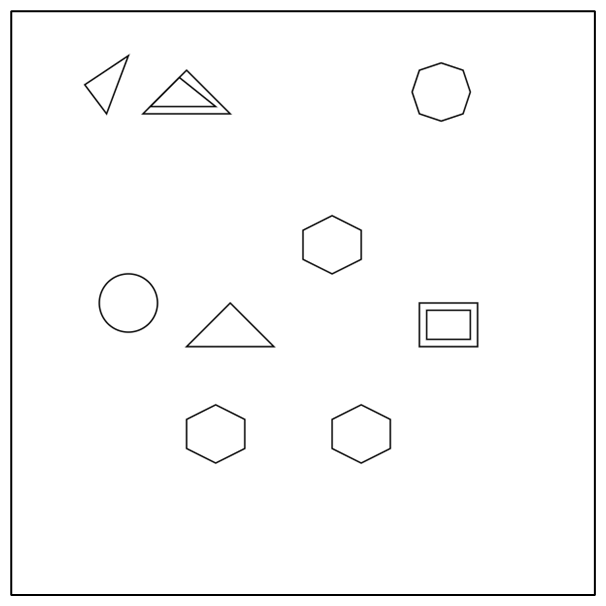}
        \caption{GPT-4o Generated}
        \label{fig:gpt_generated_ex2}
    \end{subfigure}
    \caption{Further comparison of an original visual stimulus with MLLM-generated SVG representations, illustrating common perceptual inaccuracies.}
    \label{fig:stimuli_comparison_ex2}
\end{figure}

These qualitative errors in SVG generation directly reflect underlying perceptual limitations. For example, the inability to correctly render the number of sides of a pentagon indicates a failure in fine-grained shape discrimination.

In conclusion, using SVG reconstruction as an intermediate step proved challenging and often led to a decline in performance. However, this challenge served as a valuable diagnostic tool. The qualitative inaccuracies in MLLM-generated SVGs offer a more direct window into their perceptual processing (or misprocessing) than final task answers alone. Examining these structured visual outputs allows for more effective pinpointing of specific perceptual flaws, such as difficulties with shape constancy, spatial relationships, or attribute binding. This understanding is crucial for guiding future research towards developing MLLMs with more robust and accurate visual perception capabilities.

\newpage
\section{Do Models exploit shortcuts to answer viusal perception questions?}
\label{subsec:viz-percept-cot}
\cref{tab:cot_comparison} indicates that dimensions where verbalization of the image is difficult such as visual closure, visual form constancy, there is a clear drop in performance when model is prompted with CoT style prompt. Whereas, tasks such as visual spatial, and letter disambiguation show a drastic performance improvement when prompted with CoT. Note: We generated the CoT based responses by adding a simple "think step-by-step" prompt to the original question.

\begin{table}[h]
\centering
\begin{tabular}{lcc}
\hline
\textbf{Dimension} & \textbf{CoT Prompt} & \textbf{Regular Prompt} \\
\hline
Letter Disambiguation & 47.4 & 31.85 \\
Form Constancy & 55.19 & 74.07 \\
Visual Closure & 48.21 & 57.74 \\
Visual Spatial & 35.86 & 28.91 \\
3D Visual Spatial & 38.75 & 31.25 \\
3D Color Disambiguation & 79.17 & 95.83 \\
3D Letter Disambiguation & 32.29 & 22.92 \\
3D Shape Disambiguation & 80.0 & 81.67 \\
3D Form Constancy & 30.0 & 48.75 \\
\hline
\end{tabular}
\caption{Performance comparison between CoT Prompt and Regular Prompt across different visual dimensions}
\label{tab:cot_comparison}
\end{table}

\newpage
\section{Limits of Visual Perception}
\label{sec:appendix_limits_vp}

\subsection{Visual Form Constancy: Sensitivity to Rotation at Varying Scales}
\label{subsec:form_constancy_appendix}

\begin{figure}[htbp]
    \centering
    \begin{subfigure}{0.45\textwidth}
        \centering
        \includegraphics[width=\textwidth]{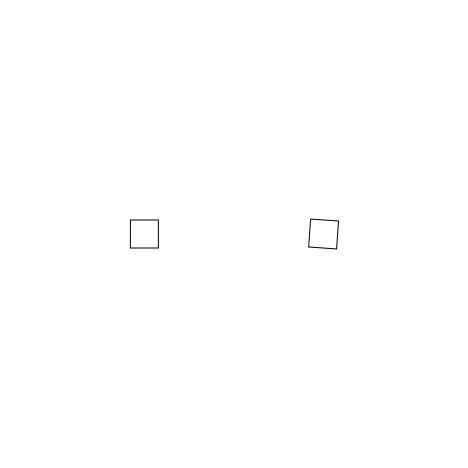}
        \caption{$Square \;Size (D=28)$; $rotation=2$}
        \label{fig:appendix-form-constancy-low-patch}
    \end{subfigure}
    \hfill
    \begin{subfigure}{0.45\textwidth}
        \centering
        \includegraphics[width=\textwidth]{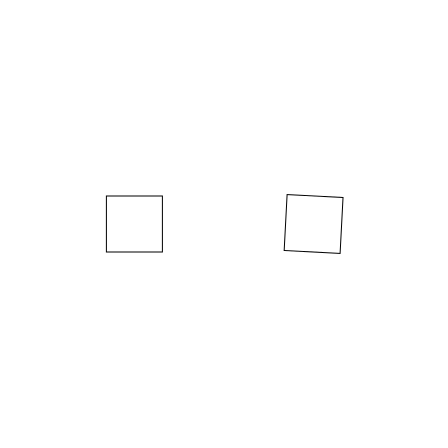}
        \caption{$Square \;Size (D=56)$; $rotation=2$}
        \label{fig:right_image}
    \end{subfigure}
    \caption{Form Constancy}
    \label{fig:appendix-form-constancy}
\end{figure}

This experiment investigates the MLLM's ability to maintain visual form constancy by detecting subtle rotations of an object, with the primary goal of determining how this perceptual capability is influenced by the absolute size of the visual stimuli, especially in relation to $P$, the patch size of the vision-encoder (assumed to be 14 pixels for this study).
Stimuli consisted of images containing two squares placed side-by-side. The left square served as a static, unrotated reference, while the right square was the target, potentially rotated by a specific angle. An example of such a stimulus is depicted in Figure~\ref{fig:appendix-form-constancy-low-patch}. The MLLM was prompted with the question: "Is the right square rotated with respect to the left one? Answer Yes or No." Key parameters varied included the square size ($D$), ranging from approximately $0.5P$ to $8P$, and the rotation angle ($\theta_r$), from $0^\circ$ to $4^\circ$. Accuracy was evaluated based on the model's correct identification of rotation presence or absence.

\begin{figure}[H]
    \centering
    \includegraphics[width=0.7\textwidth]{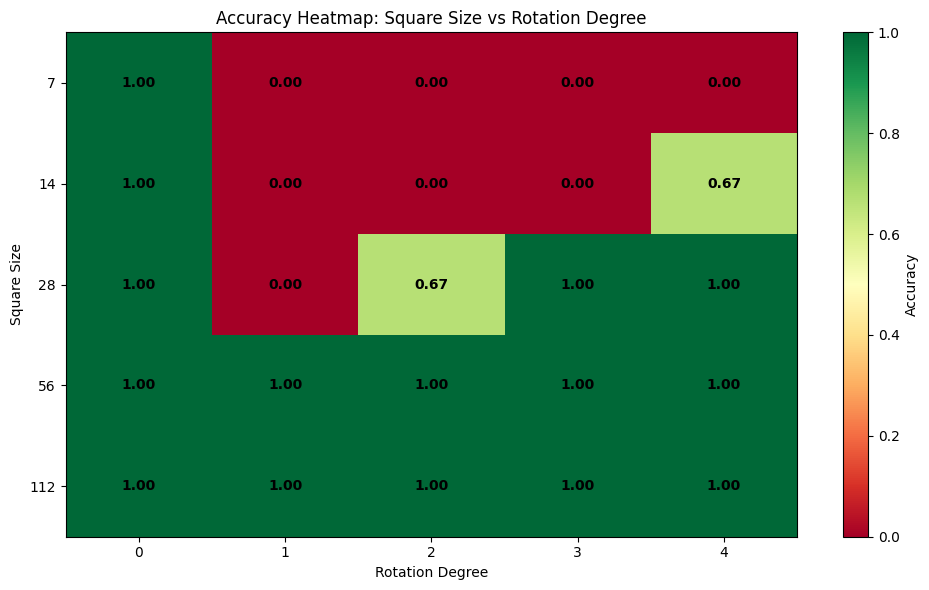}
    \caption{Claude-Sonnet-3.5 performance on rotation detection for various square sizes (scaled relative to patch size $P$) across different rotation angles ($0^\circ-4^\circ$).}
    \label{fig:form_constancy_results_app}
\end{figure}

The performance of Claude-Sonnet-3.5 in detecting rotations is illustrated in Figure~\ref{fig:appendix-form-constancy}. With no rotation ($0^\circ$), the model achieved $100\%$ accuracy for both $28\text{px}$ ($2P$) and $84\text{px}$ ($6P$) squares specifically highlighted in initial tests, and generally across other tested sizes as shown in the broader heatmap analysis. However, as seen in Figure~\ref{fig:form_constancy_results_app}, a subtle $1^\circ$ rotation caused accuracy to drop to $0\%$ for sizes up to $2P$, highlighting a significant challenge in perceiving minimal rotational changes, even when the object size substantially exceeded the patch size $P$.

A clear trend emerges regarding the interplay between object size and the ability to detect rotations.
When no rotation was applied ($0^\circ$), the model consistently achieved perfect accuracy ($1.00$) across all tested square sizes, correctly identifying the absence of transformation. The challenge arises with the introduction of even minimal rotation. For a $1^\circ$ rotation:
\begin{itemize}
    \item With sub-patch size squares ($D=7\text{px}$, $\approx 0.5P$), accuracy was $0.00$.
    \item With patch-sized squares ($D=14\text{px}$, $\approx 1P$), accuracy remained at $0.00$.
    \item For squares twice the patch size ($D=28\text{px}$, $\approx 2P$), accuracy was still $0.00$.
    \item However, for larger squares ($D=56\text{px}$, $\approx 4P$, and $D=112\text{px}$, $\approx 8P$), the model achieved perfect accuracy ($1.00$), successfully detecting this subtle $1^\circ$ rotation.
\end{itemize}
This pattern indicates a critical size threshold for perceiving minimal rotations. Objects at or below $2P$ were insufficient for the model to discern a $1^\circ$ change, but larger objects ($ \ge 4P$) provided enough visual information.

As the rotation angle increased, performance improved for smaller square sizes:
\begin{itemize}
    \item For $D=7\text{px}$ (sub-patch), the model failed to detect rotations up to $4^\circ$.
    \item For $D=14\text{px}$ (patch-size), accuracy remained at $0.00$ for rotations up to $3^\circ$, with a partial recovery to $0.67$ at $4^\circ$. This suggests that even when an object is nominally the size of a patch, its internal features might not be sufficiently resolved to detect small angular changes until the rotation becomes more pronounced.
    \item For $D=28\text{px}$ ($2P$), accuracy was $0.00$ at $1^\circ$, recovered to $0.67$ at $2^\circ$, and reached $1.00$ for $3^\circ$ and $4^\circ$ rotations.
    \item For $D=56\text{px}$ ($4P$) and $D=112\text{px}$ ($8P$), the model maintained $1.00$ accuracy across all tested rotation angles from $1^\circ$ to $4^\circ$.
\end{itemize}

These results refine our understanding of MLLM perceptual limits. Firstly, there is a clear perceptual threshold for detecting subtle rotations, which is heavily dependent on the object's size relative to the encoder's patch resolution. Sub-patch ($0.5P$) and patch-sized ($1P$) objects are particularly challenging for detecting small angular displacements. Secondly, even for objects larger than a single patch (e.g., $2P$), very subtle rotations ($1^\circ$) can remain imperceptible. It appears that a significantly larger effective object size (around $4P$ or more in this experiment) is required for robust detection of minimal (e.g., $1^\circ$) rotations. This suggests that the model requires a certain aggregation of features over multiple patches, or a higher resolution representation of the object's boundaries, to reliably discern such fine-grained geometric transformations. The findings underscore that robust form constancy is not merely about resolving an object but also about accurately interpreting its subtle geometric attributes, a capability that scales with the object's representation quality, which in turn is linked to its size relative to the visual encoding mechanism.

\subsection{Visual Discrimination: Shape Counting Accuracy at Varying Scales}
\label{subsec:shape_counting_appendix}

\begin{figure}[htbp]
    \centering
    \begin{subfigure}{0.45\textwidth}
        \centering
        \includegraphics[width=\textwidth]{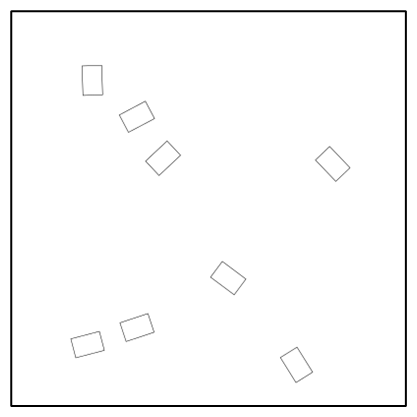}
        \caption{$Rectangle \;Size (D=28)$; $Instances=8$}
        \label{fig:rect-scaling-vp-low}
    \end{subfigure}
    \hfill
    \begin{subfigure}{0.45\textwidth}
        \centering
        \includegraphics[width=\textwidth]{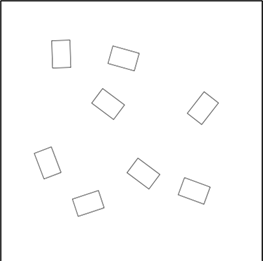}
        \caption{$Rectangle \;Size (D=56)$; $Instances=8$}
        \label{fig:right_image}
    \end{subfigure}
    \caption{Shape Discrimination}
    \label{fig:rect-scaling-vp}
\end{figure}

\begin{figure}[H]
    \centering
    \includegraphics[width=0.8\textwidth]{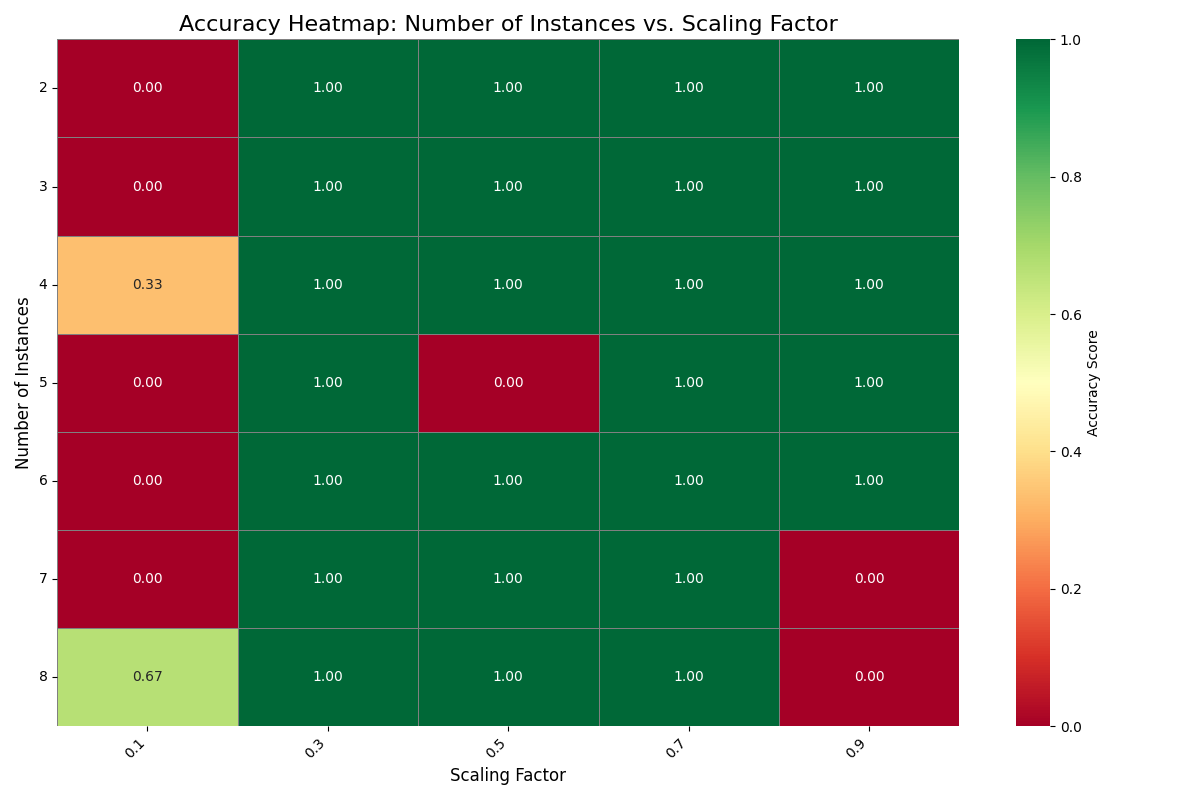}
    \caption{Accuracy heatmap for Claude-Sonnet-3.5 on rectangle (square) counting. Performance varies with the number of instances and scaling factor (size). Darker green indicates higher accuracy.}
    \label{fig:shape_counting_results_app}
\end{figure}

This experiment evaluates the MLLM's visual discrimination capabilities, focusing on its accuracy in counting instances of simple rectangles under varying conditions of object size and number of instances (density). The study investigates how these factors influence enumeration, particularly considering the object size relative to $P$.
Stimuli comprised images with a varying number of non-overlapping rectangles (specifically squares in this instance) of uniform size within each image. An example stimulus is shown in Figure~\ref{fig:rect-scaling-vp}. The MLLM was prompted: "How many rectangles are in the image? Answer with a number." Parameters varied were the number of rectangles (2 to 8) and their size. The "Scaling Factor" in the results (Figure~\ref{fig:shape_counting_results_app}) corresponds to rectangle sizes ($D$) from approximately $7\text{px}$ ($P \times 0.5$) to $112\text{px}$ ($P \times 8$). Accuracy was measured by comparing the model's count to the ground truth.\textit{ Note that the recatngle size here refers to the longer edge, the shorter edge is fixed at 0.8*rectangle size.}

Figure~\ref{fig:shape_counting_results_app} presents a heatmap of Claude-Sonnet-3.5's counting accuracy. A critical factor was object size: for the smallest rectangles (Scaling Factor 0.1, $D \approx P \times 0.5$), accuracy was consistently low ($0.00$ for 2-7 instances, with what appears to be an anomalous $0.67$ for 8 instances, possibly a statistical blip). This suggests significant difficulty in resolving or individuating sub-patch-sized objects, likely due to information conflation within single patches.
Conversely, for larger rectangles (Scaling Factor $\ge 0.3$, $D \ge P \times 2.1$), the model generally achieved perfect or near-perfect accuracy ($1.00$) across most instance counts. This indicates that if individual objects are clearly resolvable, the basic counting task is manageable. However, anomalous performance drops occurred even for larger rectangles at specific instance counts (e.g., 5 instances at SF 0.5; 7 and 8 instances at SF 0.9, all dropping to $0.00$ accuracy). These failures, not attributable to object resolvability by size alone, might stem from attentional lapses, visual crowding effects where objects impair perception of each other, or specific challenging spatial configurations arising from random placement.

In conclusion, MLLM visual discrimination is highly dependent on object size relative to patch resolution. While objects significantly larger than $P$ are generally counted accurately, sub-patch sized objects pose a substantial challenge. Furthermore, unexpected failures with larger objects suggest that factors beyond simple size and count, such as attention mechanisms, resilience to clutter, and interpretation of spatial arrangements, also critically influence perceptual accuracy. These experiments collectively highlight that even fundamental visual tasks can expose nuanced limitations in MLLM perception, emphasizing the need for continued research into enhancing the robustness and fidelity of their visual understanding.

\newpage
\section{Detailed Results}
\label{sec:appendix-detailed-results}
\begin{figure}[htbp]
\vspace{-5mm}

    \centering
    \begin{minipage}{0.4\textwidth}
        \centering
        \includegraphics[width=\linewidth]{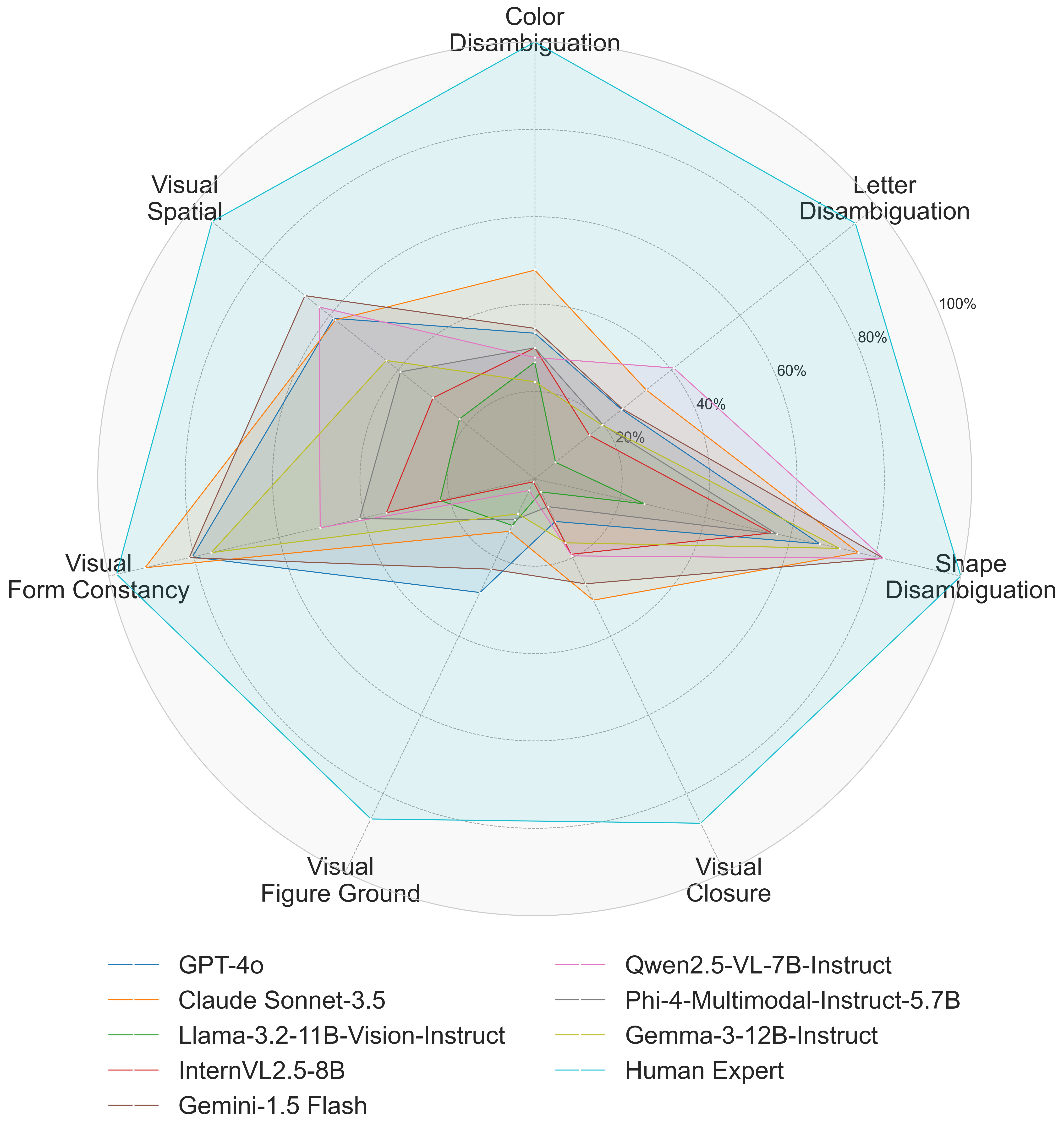}
        \caption{\textit{MLLM performance on the seven subtasks in \textbf{Do You See Me (2D)} benchmark dataset.}}
        \label{fig:syn-dataset-2D}
    \end{minipage}
    \hfill
    \begin{minipage}{0.4\textwidth}
        \centering
        \includegraphics[width=\linewidth]{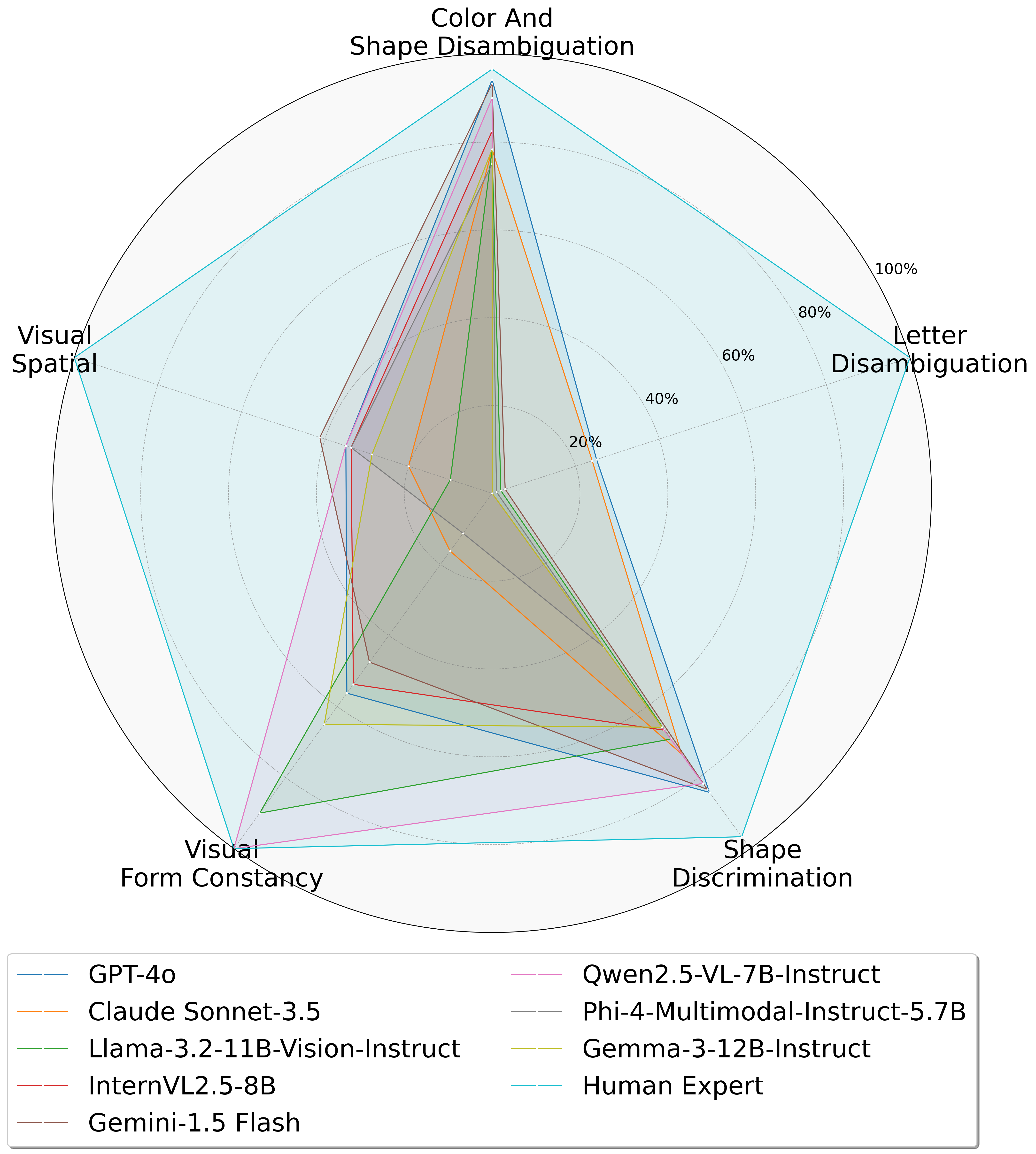}
        \caption{\textit{MLLM performance on the seven subtasks in \textbf{Do You See Me (3D)} benchmark dataset.}}
        \label{fig:syn-dataset-3D}
    \end{minipage}
\vspace{-5mm}
\end{figure}

\begin{table}[h!]
\centering
\caption{Comparison of model performance. Claude Sonnet-3.5 leads in both \emph{reasoning} and \emph{visual perception} questions.}
\resizebox{0.7\linewidth}{!}{
\begin{tabular}{lcc}
\toprule
\textbf{Model} & \textbf{Reasoning Acc.~(\%)} & \textbf{Perception Acc.~(\%)} \\
\midrule
Claude Sonnet-3.5 & \textbf{40.95} & \textbf{45.21} \\
GPT-4o & 32.97 & 42.55 \\
Gemini 1.5 Flash & 32.97 & 44.68 \\
\hline
Qwen2.5-VL-7B-Instruct & 35.63 & 35.10 \\
Intern2.5-VL-8B & 27.66 & 37.23 \\
Phi-4-Multimodal-Instruct-5.7B & 28.72 & 29.78 \\
Llama3.2-11B-Vision-Instruct & 26.06 & 31.91 \\
\bottomrule
\end{tabular}}
\vspace{-2mm}
\vspace{-4mm}
\label{tab:model-accuracy-2D}
\end{table}

\newpage

\begin{figure}[htb]
    \centering
    \begin{tabular}{cc}
        \includegraphics[width=0.4\textwidth]{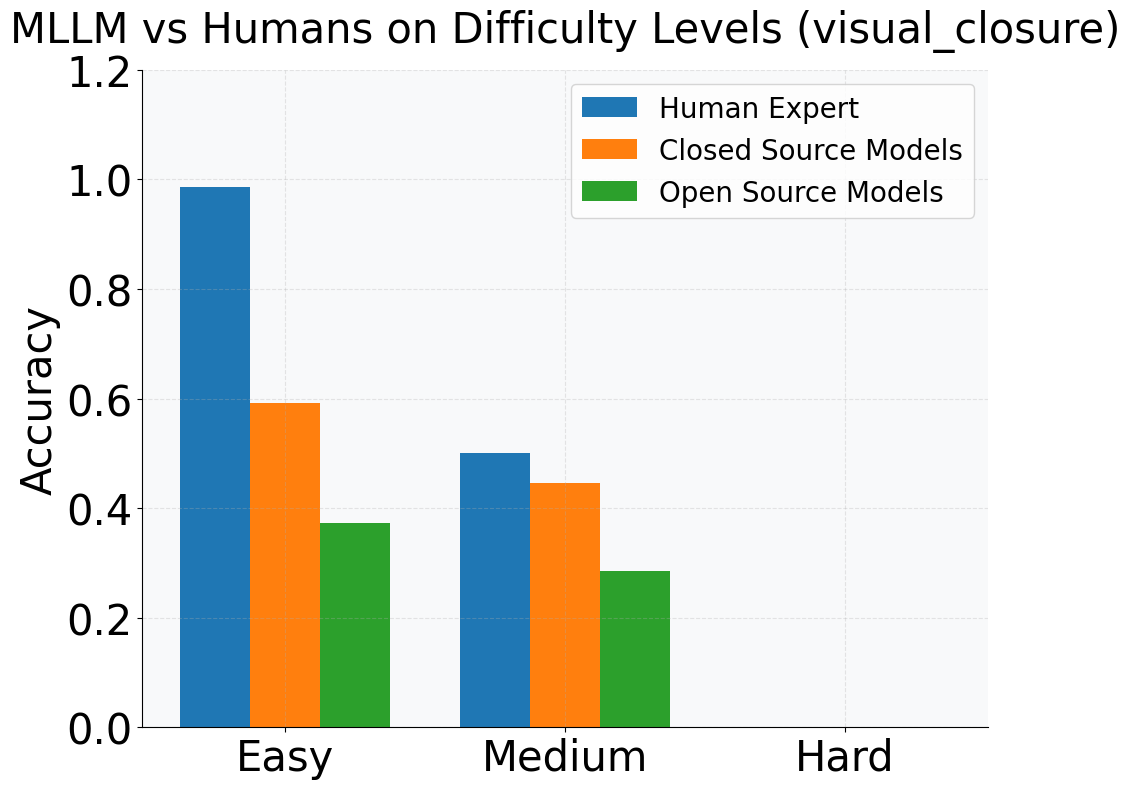} &
        \includegraphics[width=0.4\textwidth]{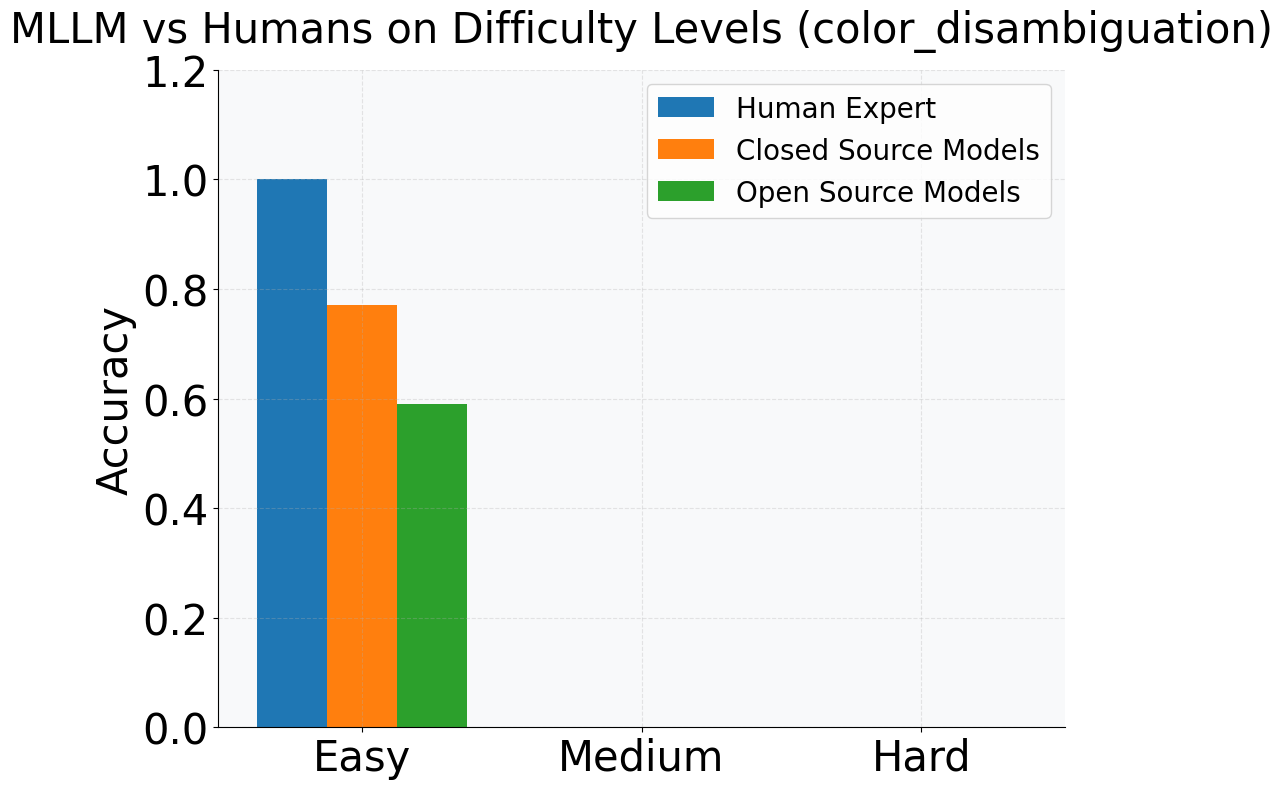} \\
        \includegraphics[width=0.4\textwidth]{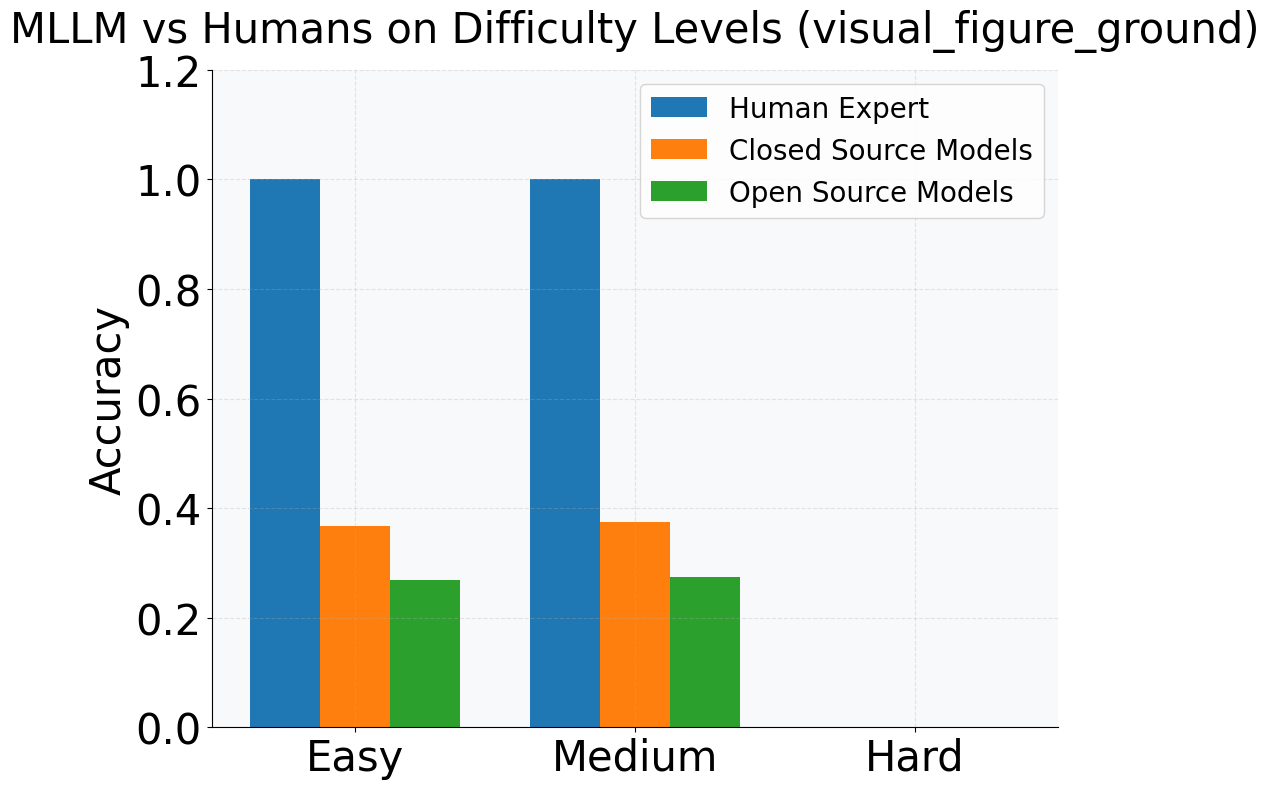} &
        \includegraphics[width=0.4\textwidth]{images/sweep_results/model_performance_sweep_vs_human_subjective_pref_form_constancy.png} \\
        \includegraphics[width=0.4\textwidth]{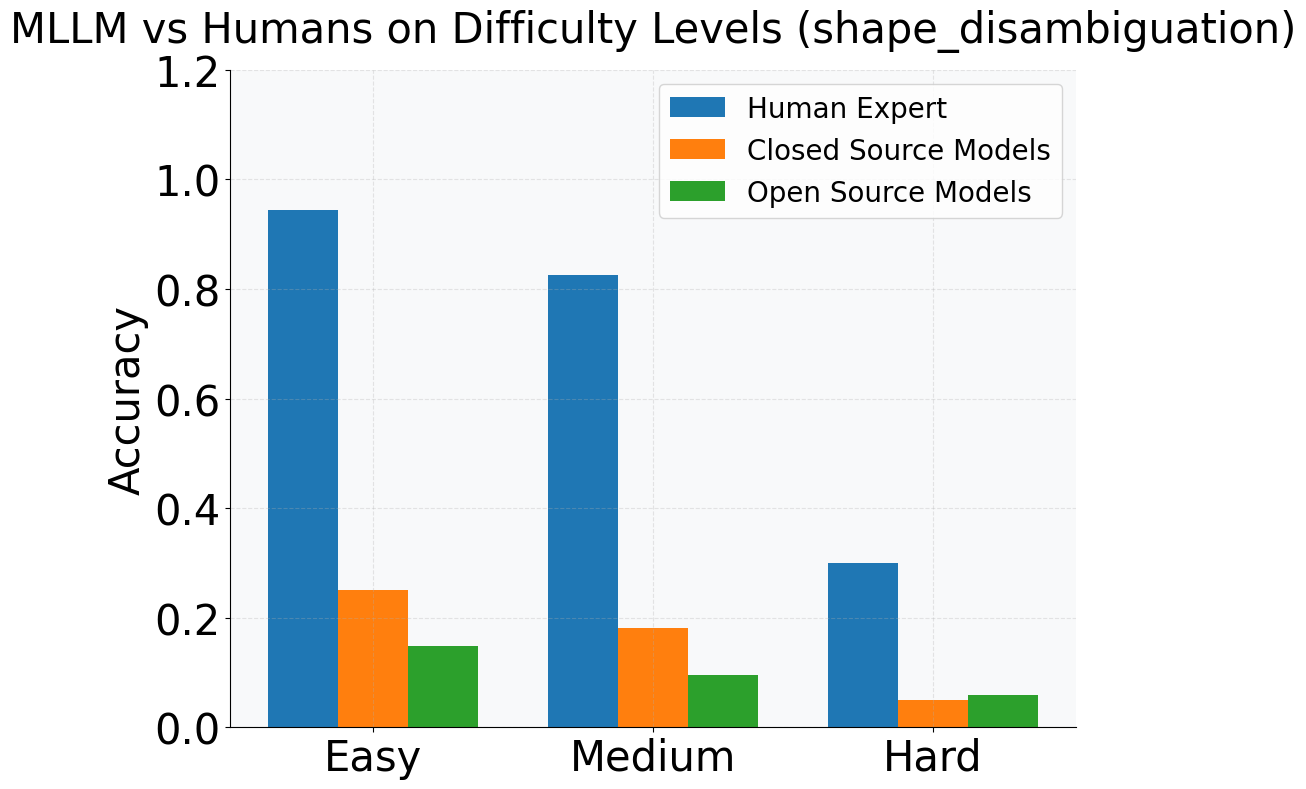} &
        \includegraphics[width=0.4\textwidth]{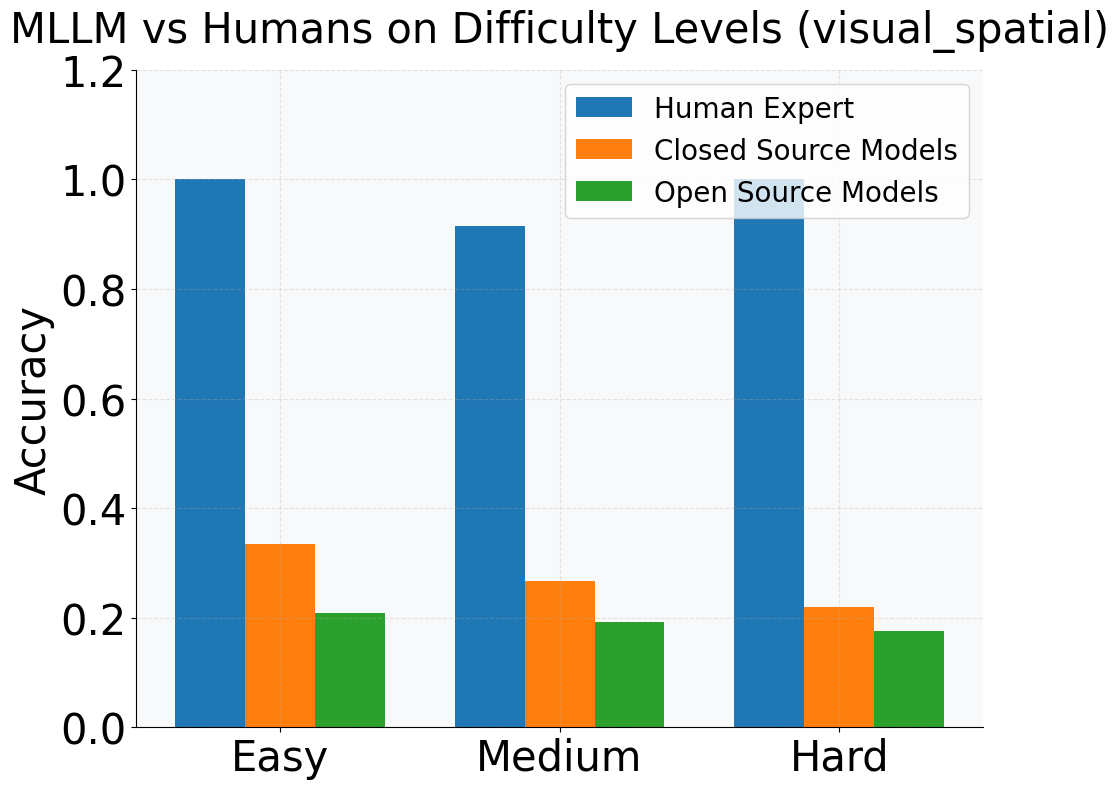} \\
        \includegraphics[width=0.4\textwidth]{images/sweep_results/model_performance_sweep_vs_human_subjective_pref_letter.png} &
        \\
        
    \end{tabular}
    \caption{MLLM performance on Human Rated Difficulty Levels. Note: \emph{Empty human bar for a difficulty level indicates that no samples were attributed the corresponding difficulty level.}}
    \label{fig:yourfigure}
\end{figure}

\newpage

\begin{figure}[htb]
    \centering
    \begin{tabular}{cc}
        \includegraphics[width=0.4\textwidth]{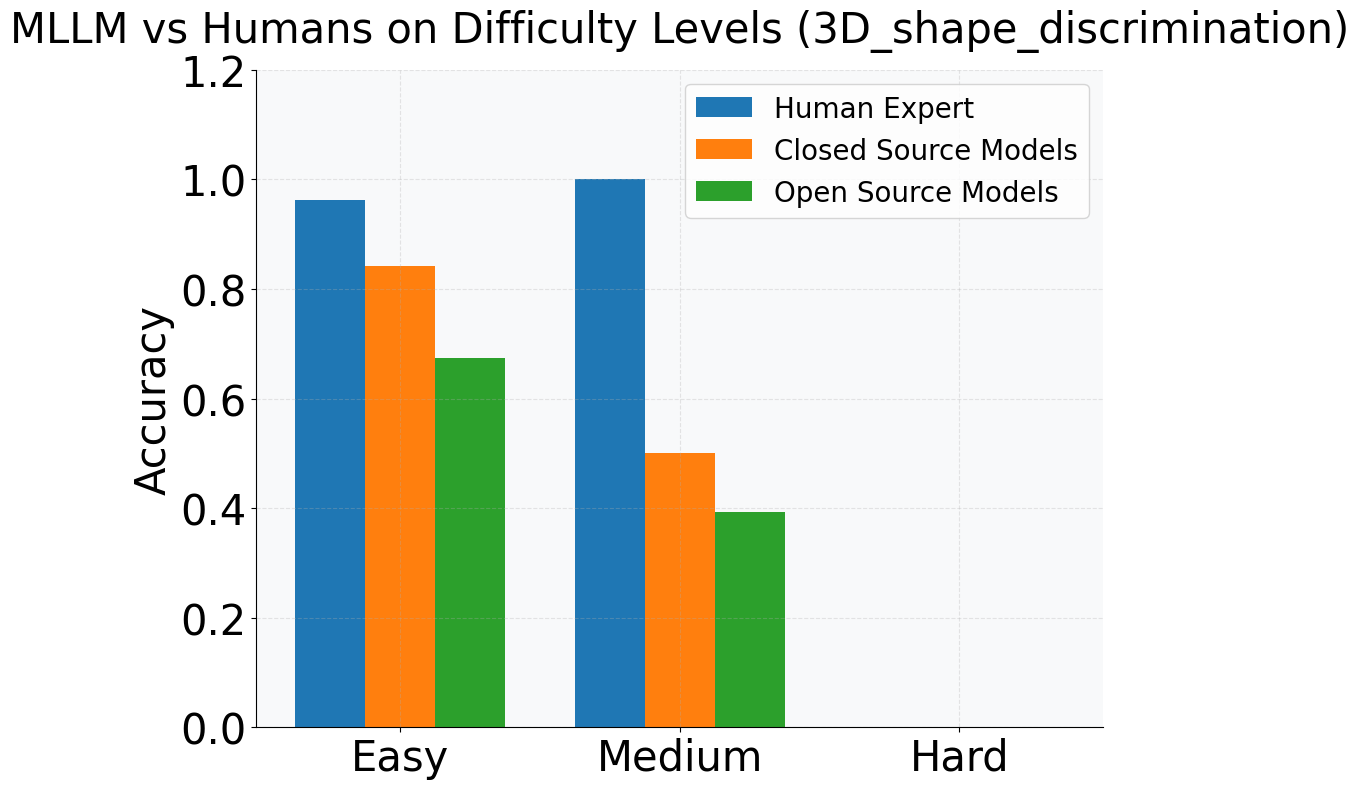} &
        \includegraphics[width=0.4\textwidth]{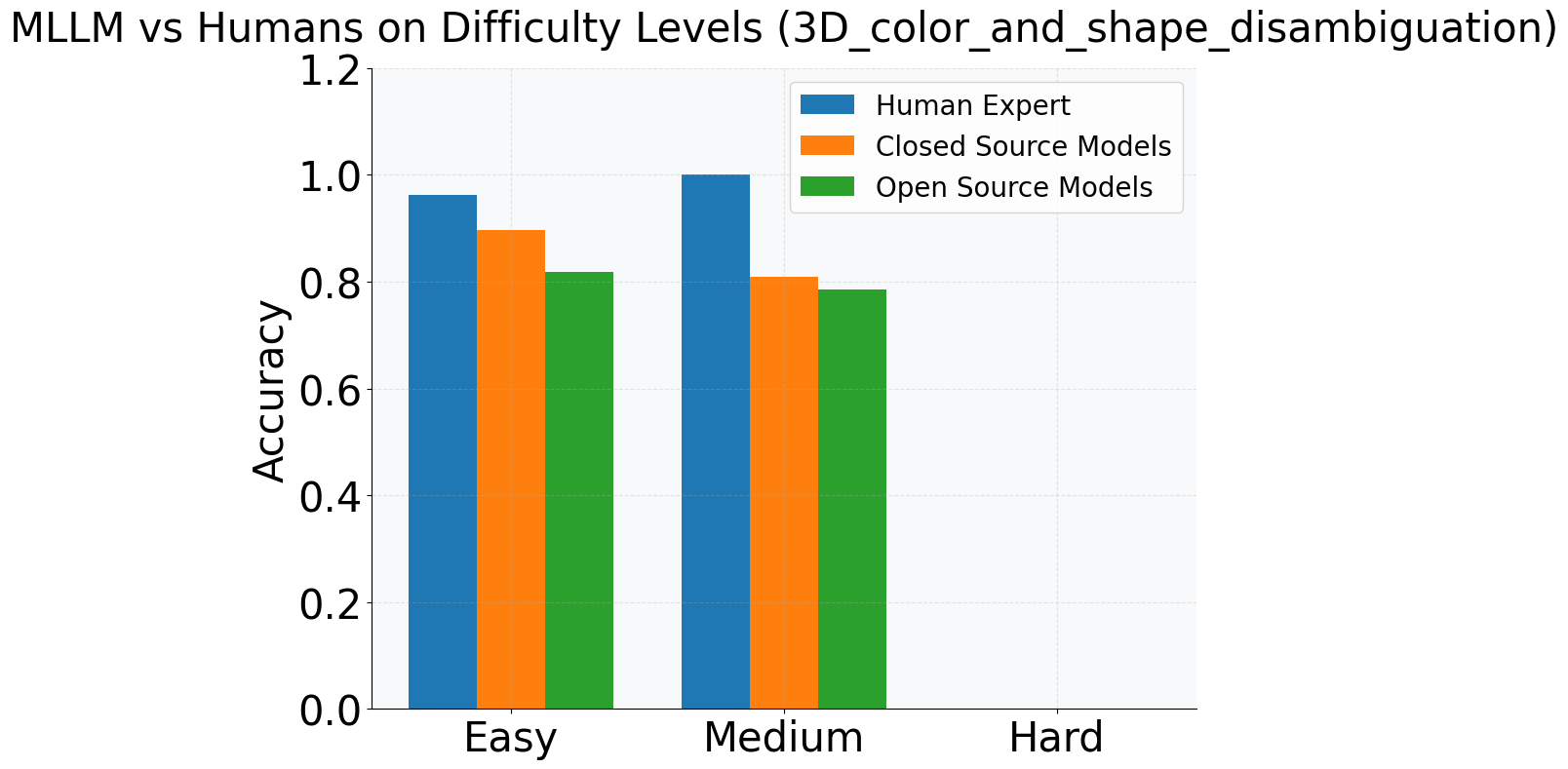} \\
        \includegraphics[width=0.4\textwidth]{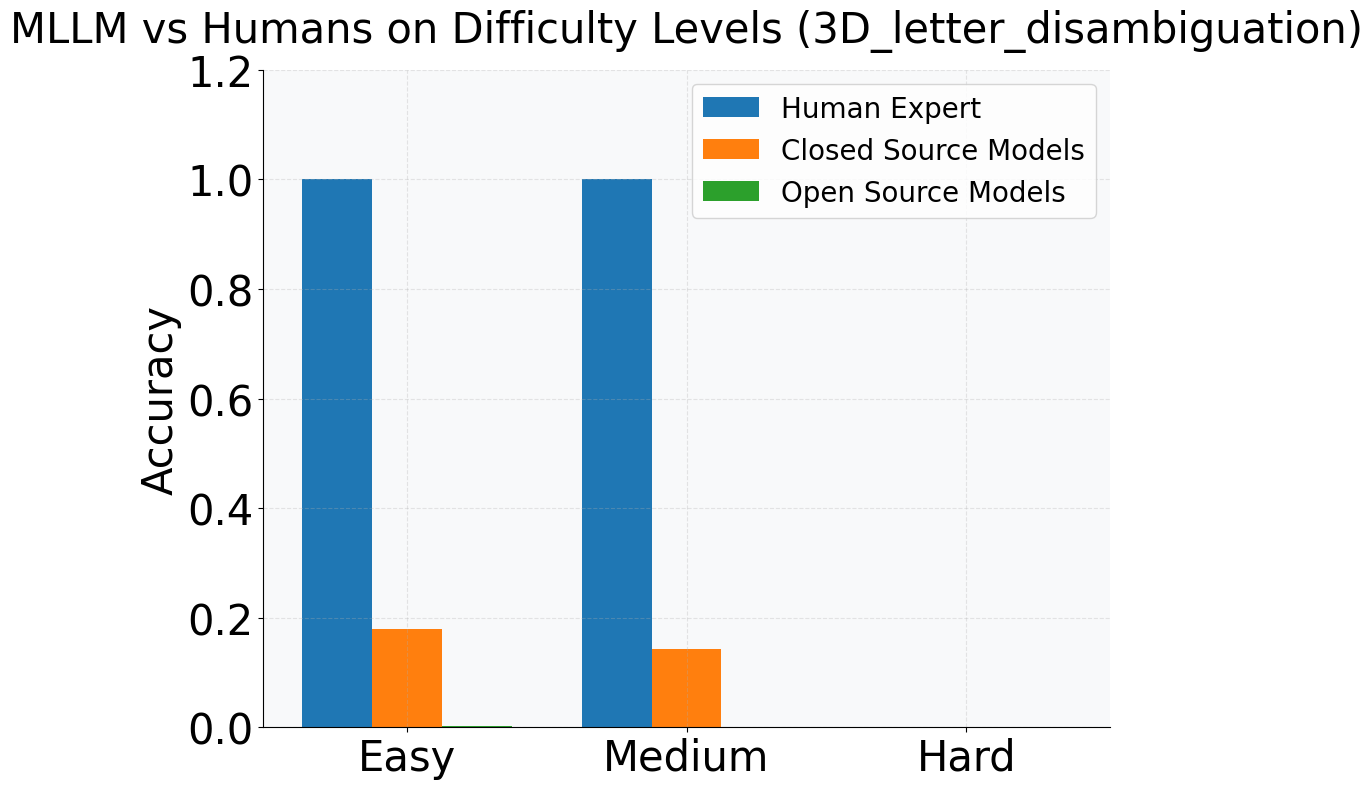} &
        \includegraphics[width=0.4\textwidth]{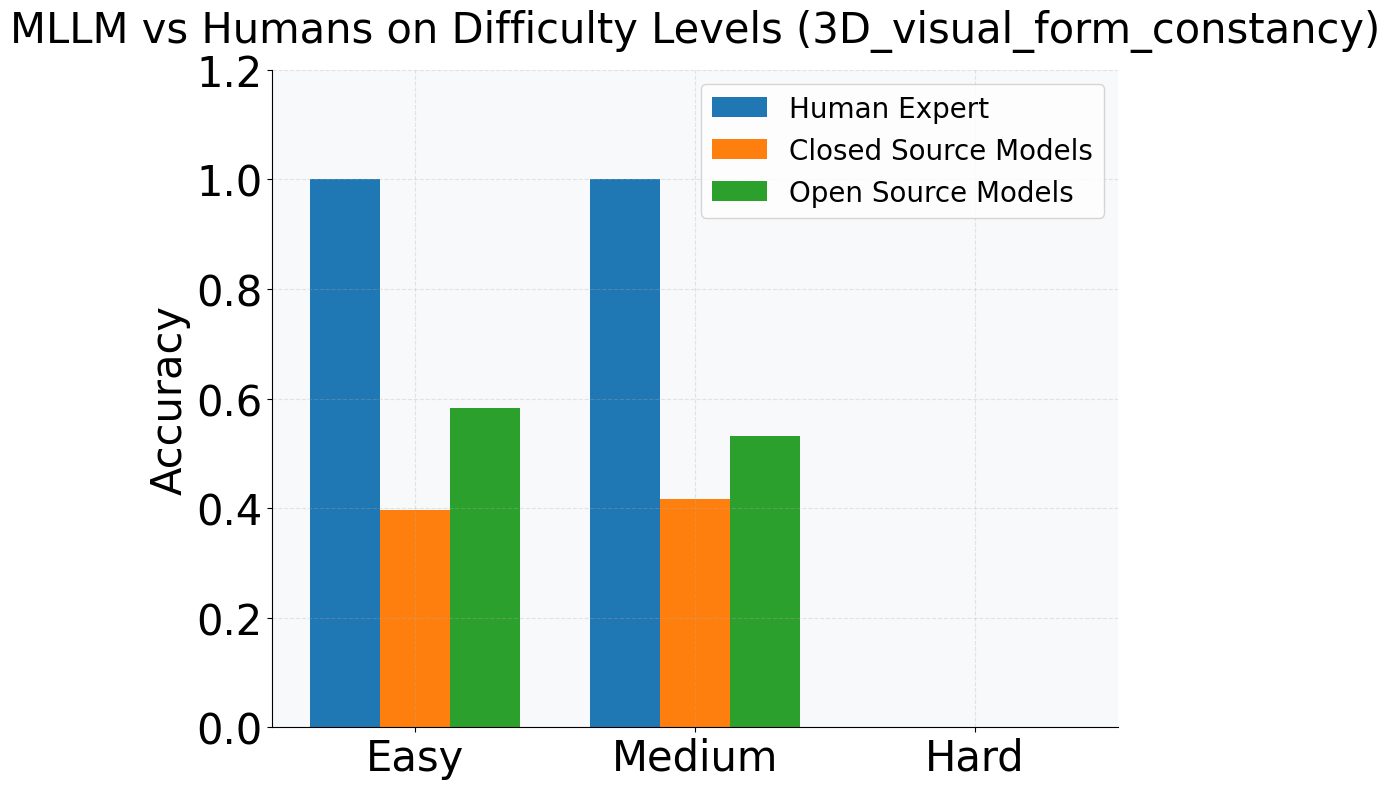} \\
    \end{tabular}
    \caption{MLLM performance on Human Rated Difficulty Levels. Note: \emph{Empty human bar for a difficulty level indicates that no samples were attributed the corresponding difficulty level}(3D tasks).}
    \label{fig:yourfigure}
\end{figure}

\newpage

\begin{figure}[h!]
    \centering
    \begin{tabular}{c}
        \includegraphics[width=0.7\textwidth]{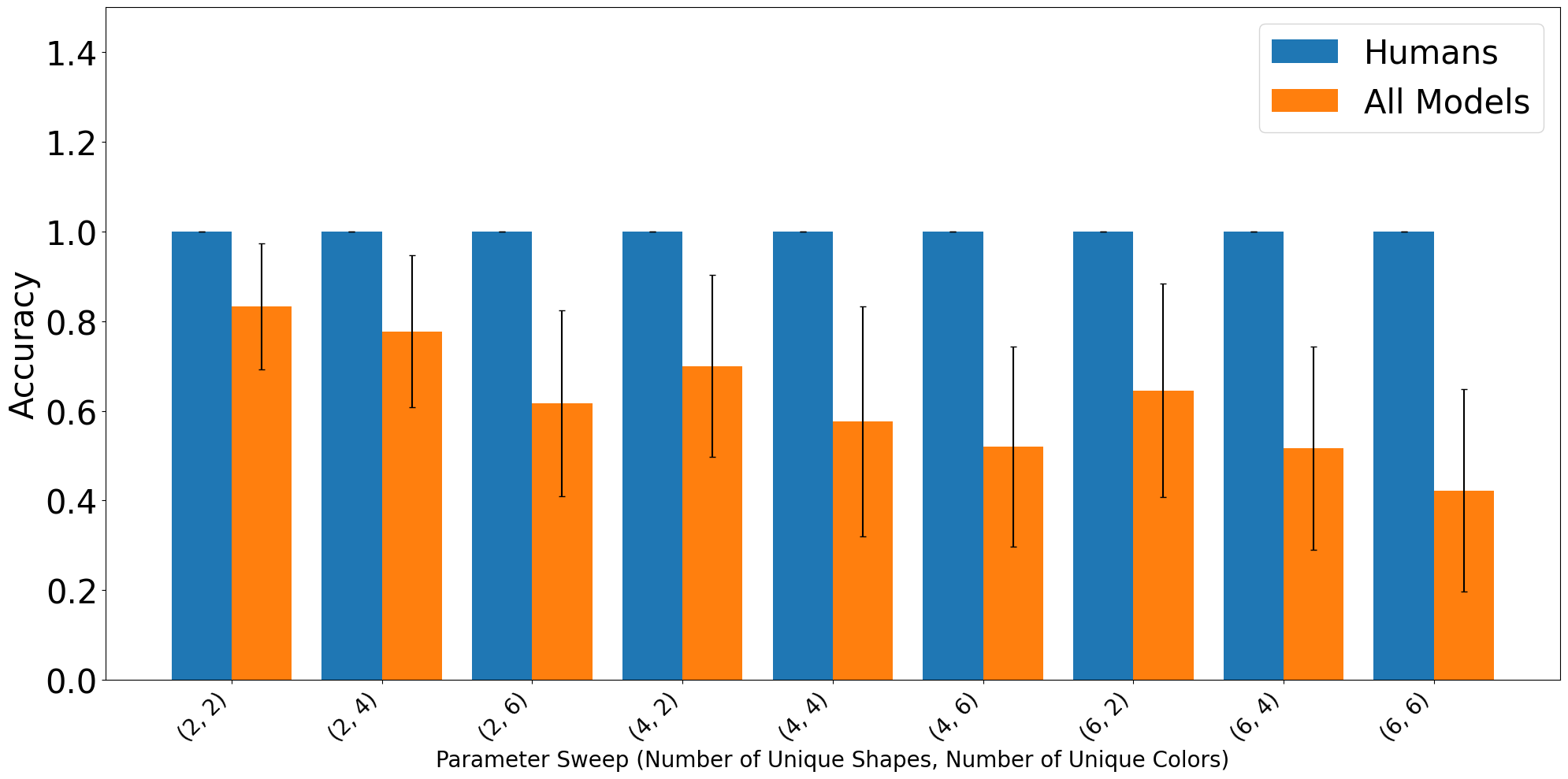} \\
        \includegraphics[width=0.7\textwidth]{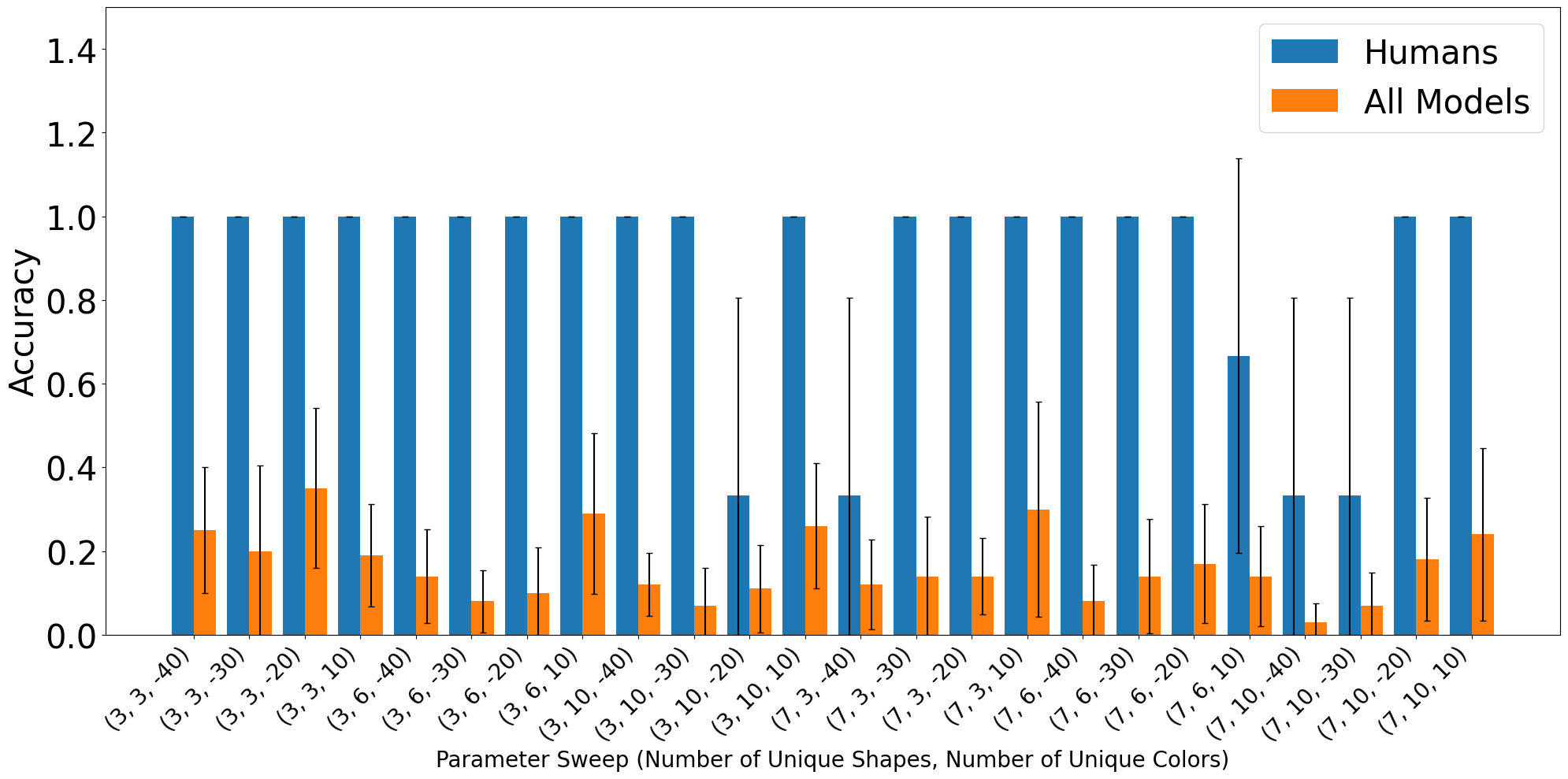} \\
        \includegraphics[width=0.7\textwidth]{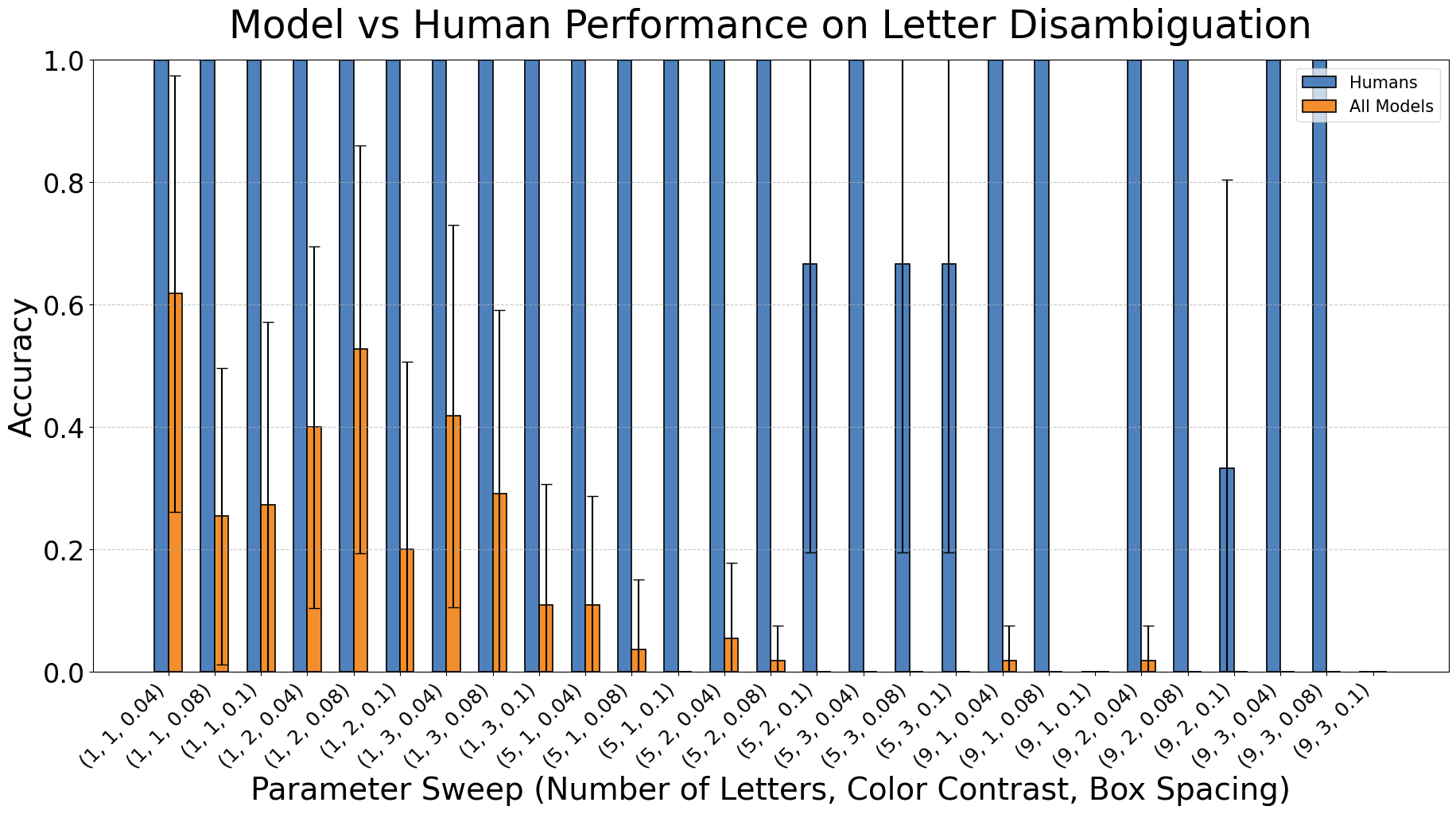} \\
        \includegraphics[width=0.7\textwidth]{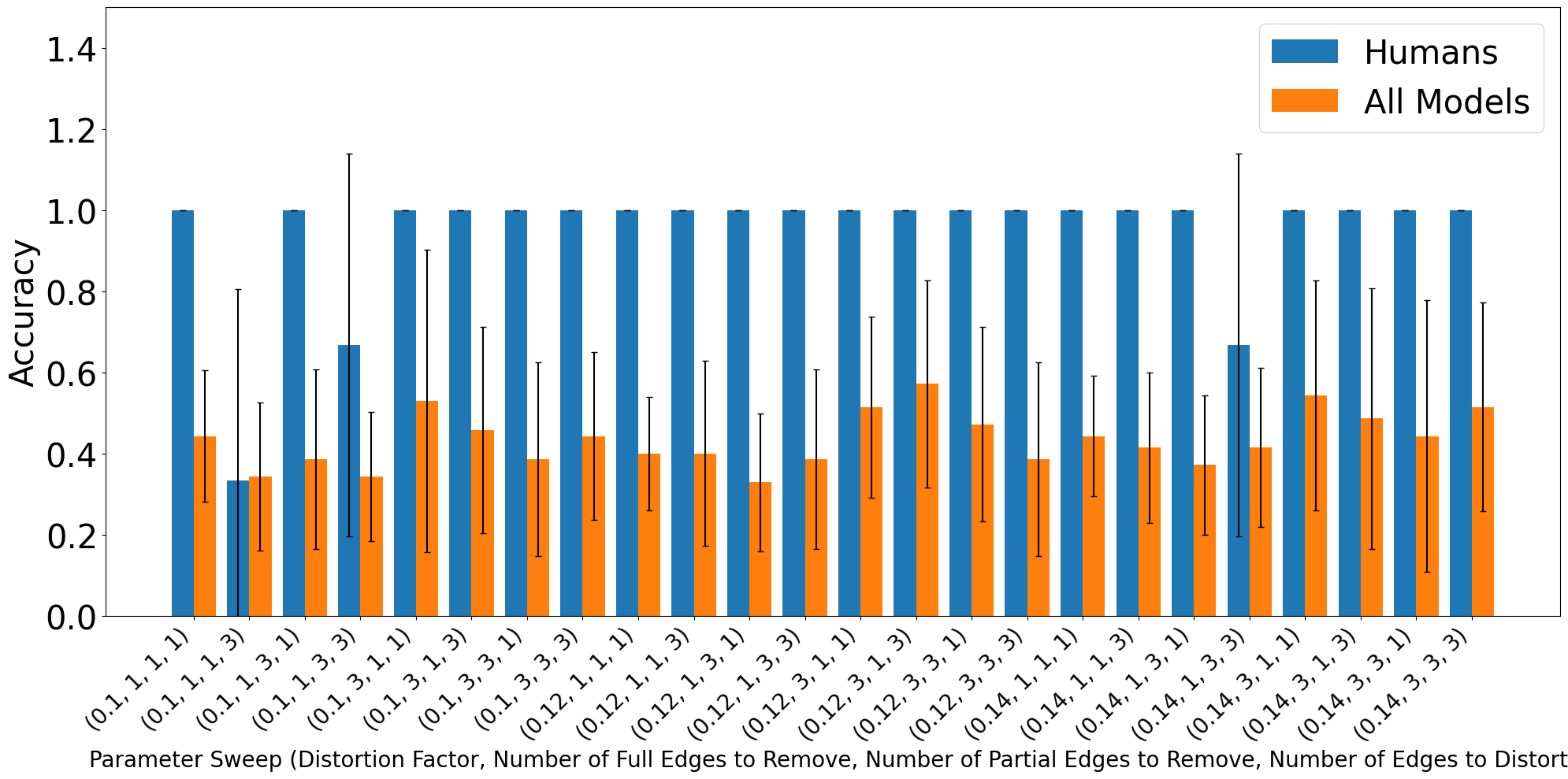} \\
        \\
    \end{tabular}
    \caption{Average MLLM performance over a sweep of combinations of control parameters. }
    \label{fig:yourfigure}
\end{figure}

\begin{figure}[h!]
    \centering
    \begin{tabular}{c}
        \includegraphics[width=0.7\textwidth]{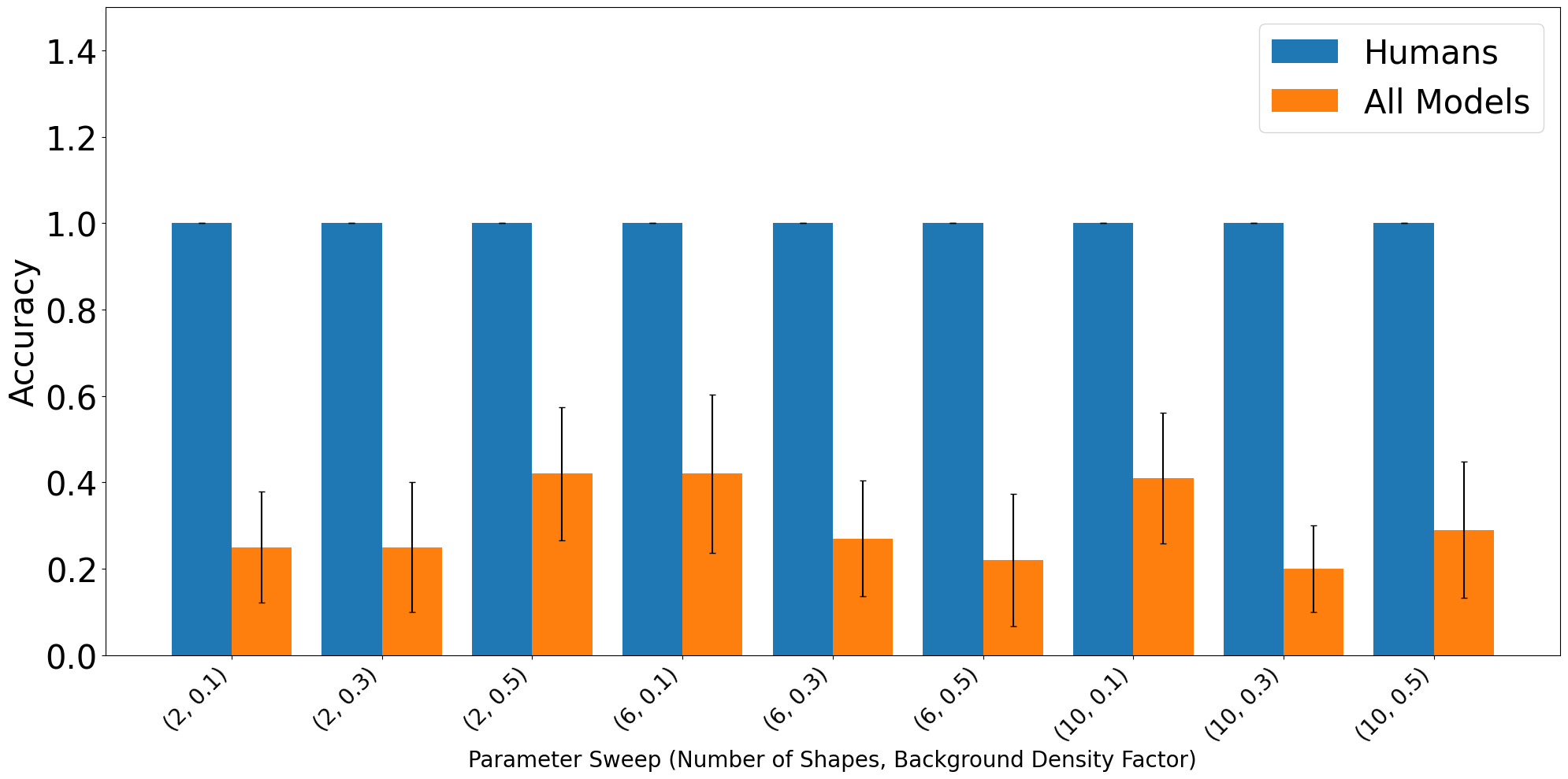} \\
        \includegraphics[width=0.7\textwidth]{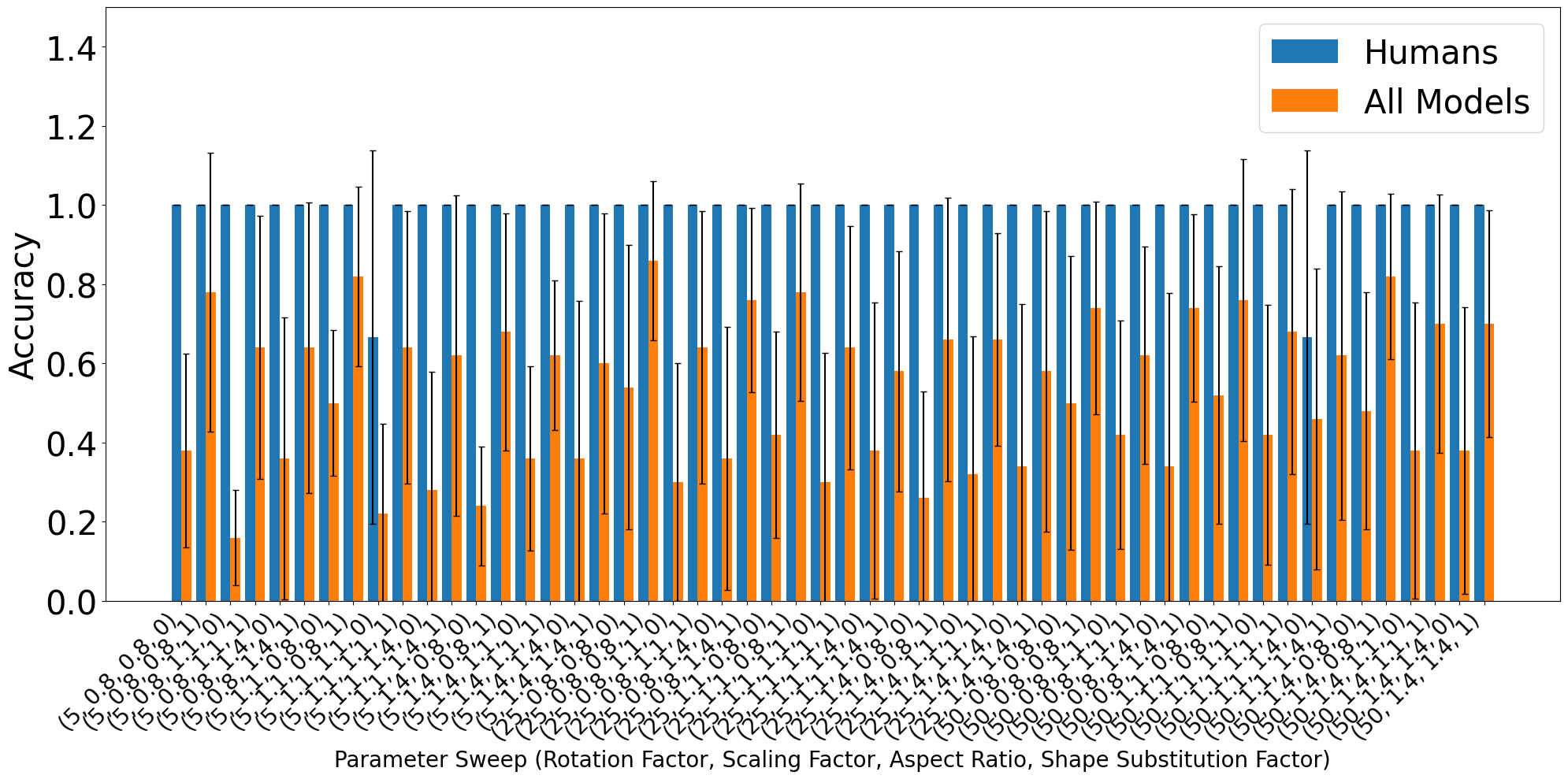} \\
        \includegraphics[width=0.7\textwidth]{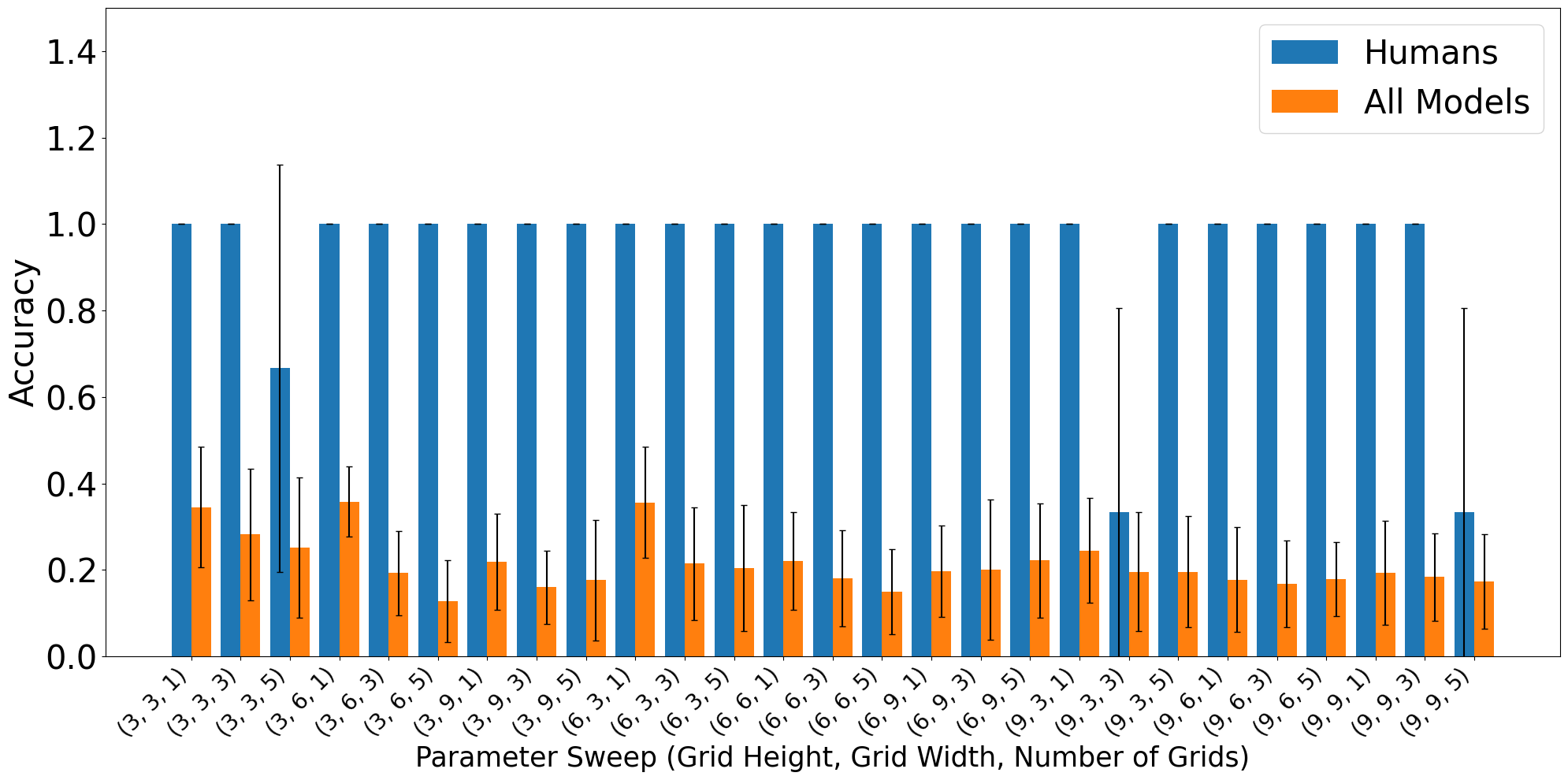} \\
        \\
    \end{tabular}
    \caption{Average MLLM performance over a sweep of combinations of control parameters. }
    \label{fig:yourfigure}
\end{figure}

\begin{figure}[h!]
    \centering
    \begin{tabular}{c}
        \includegraphics[width=0.7\textwidth]{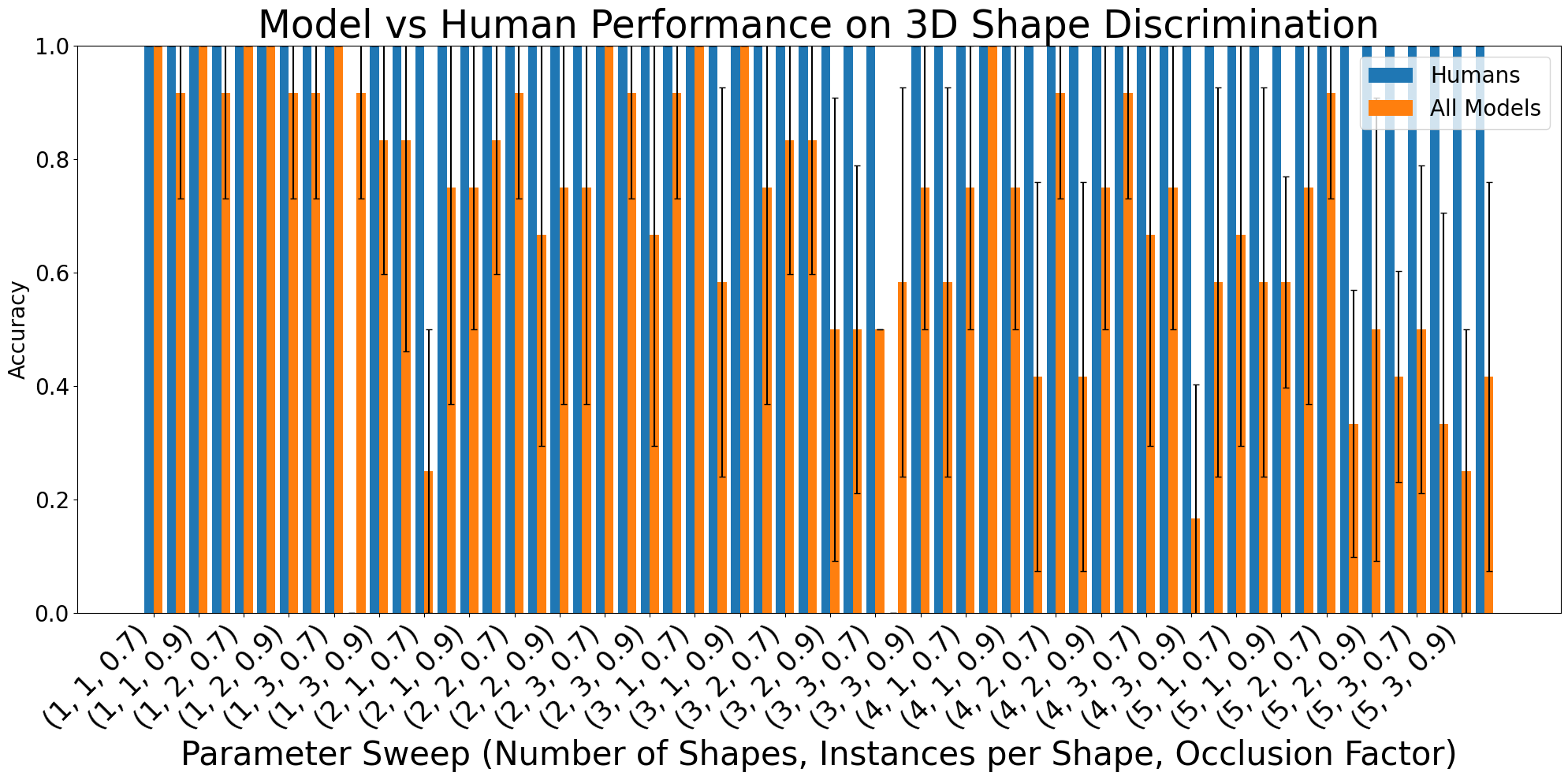} \\
        \includegraphics[width=0.7\textwidth]{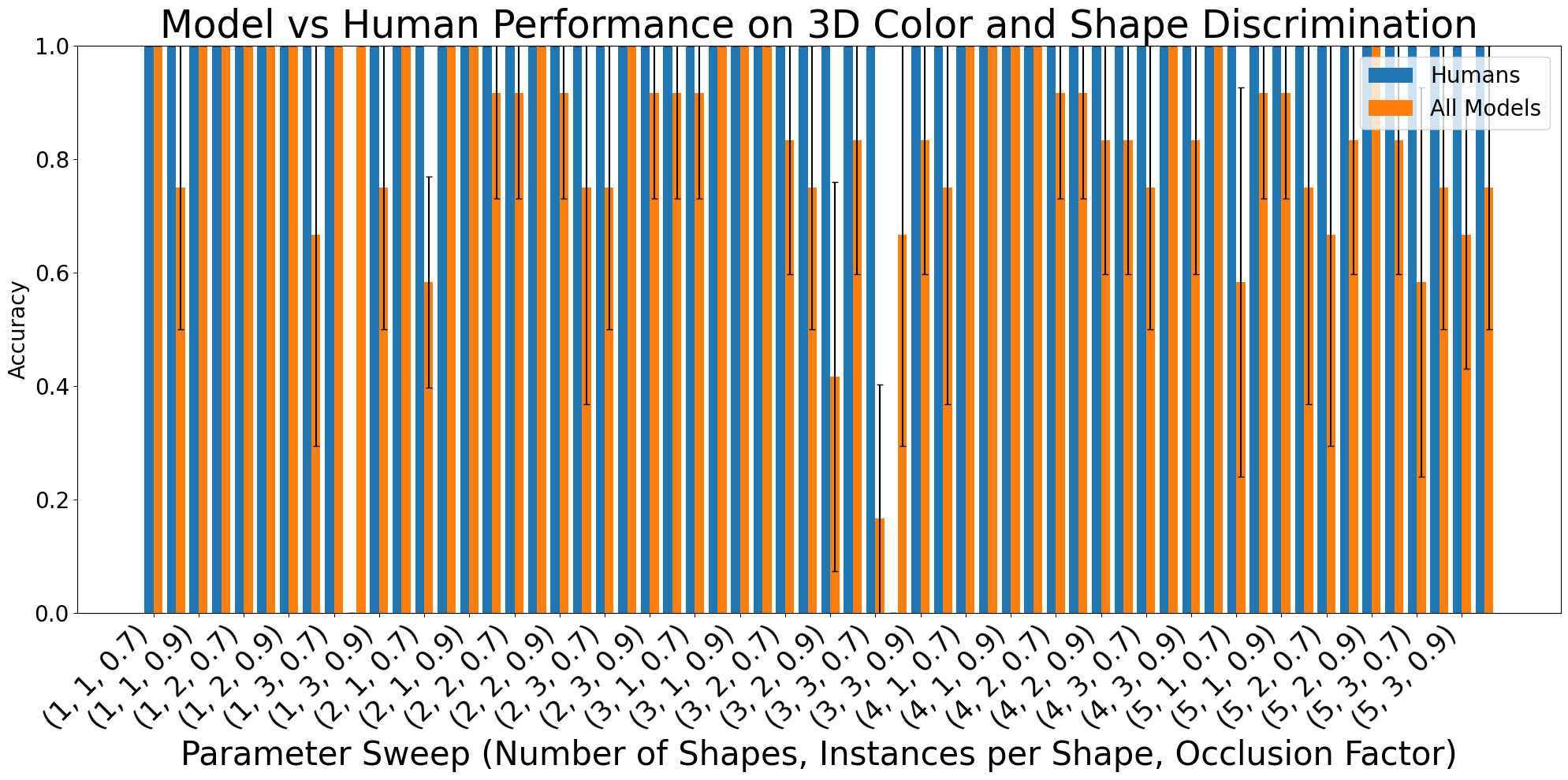} \\
        \includegraphics[width=0.7\textwidth]{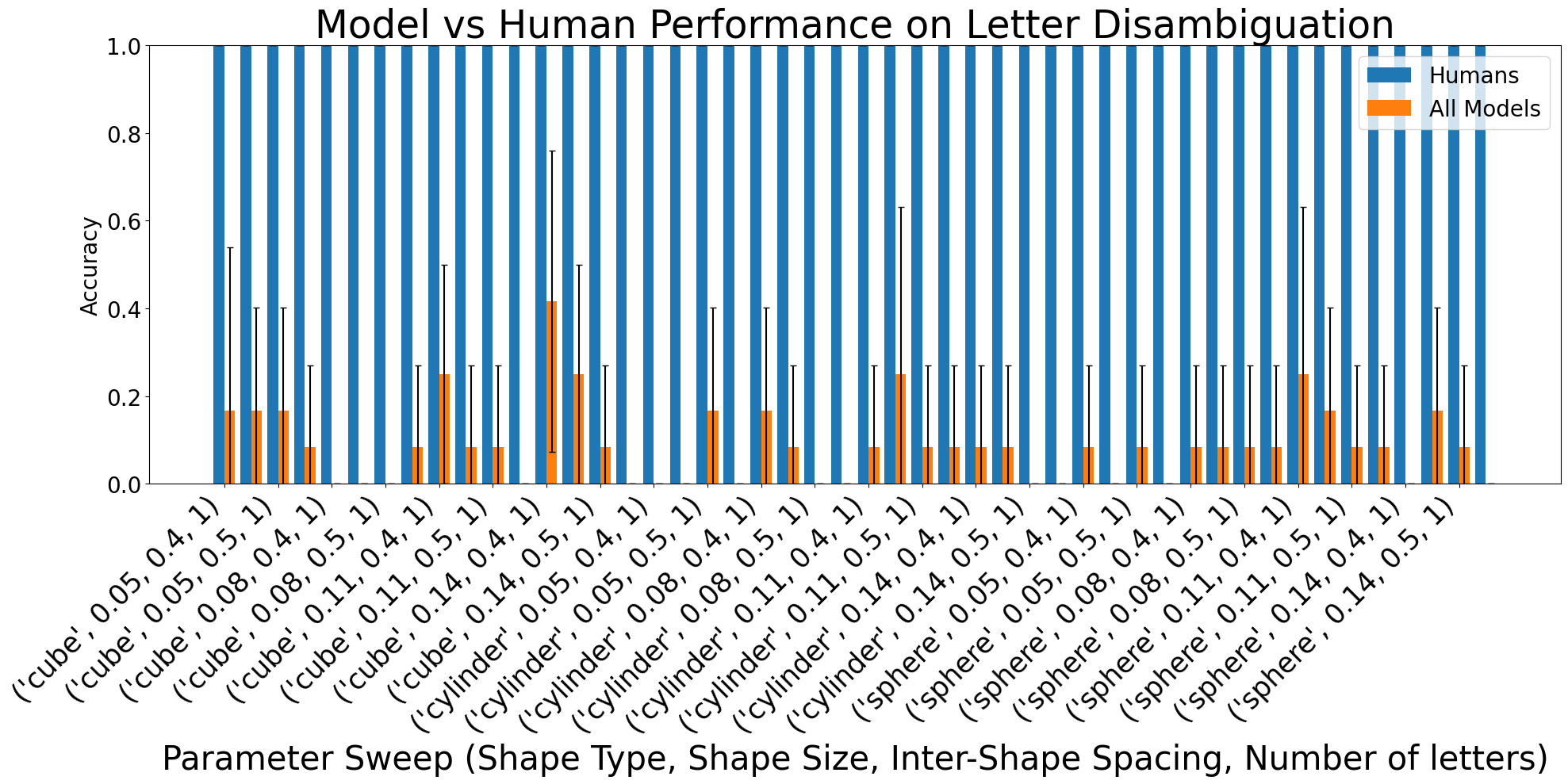} \\
        \includegraphics[width=0.7\textwidth]{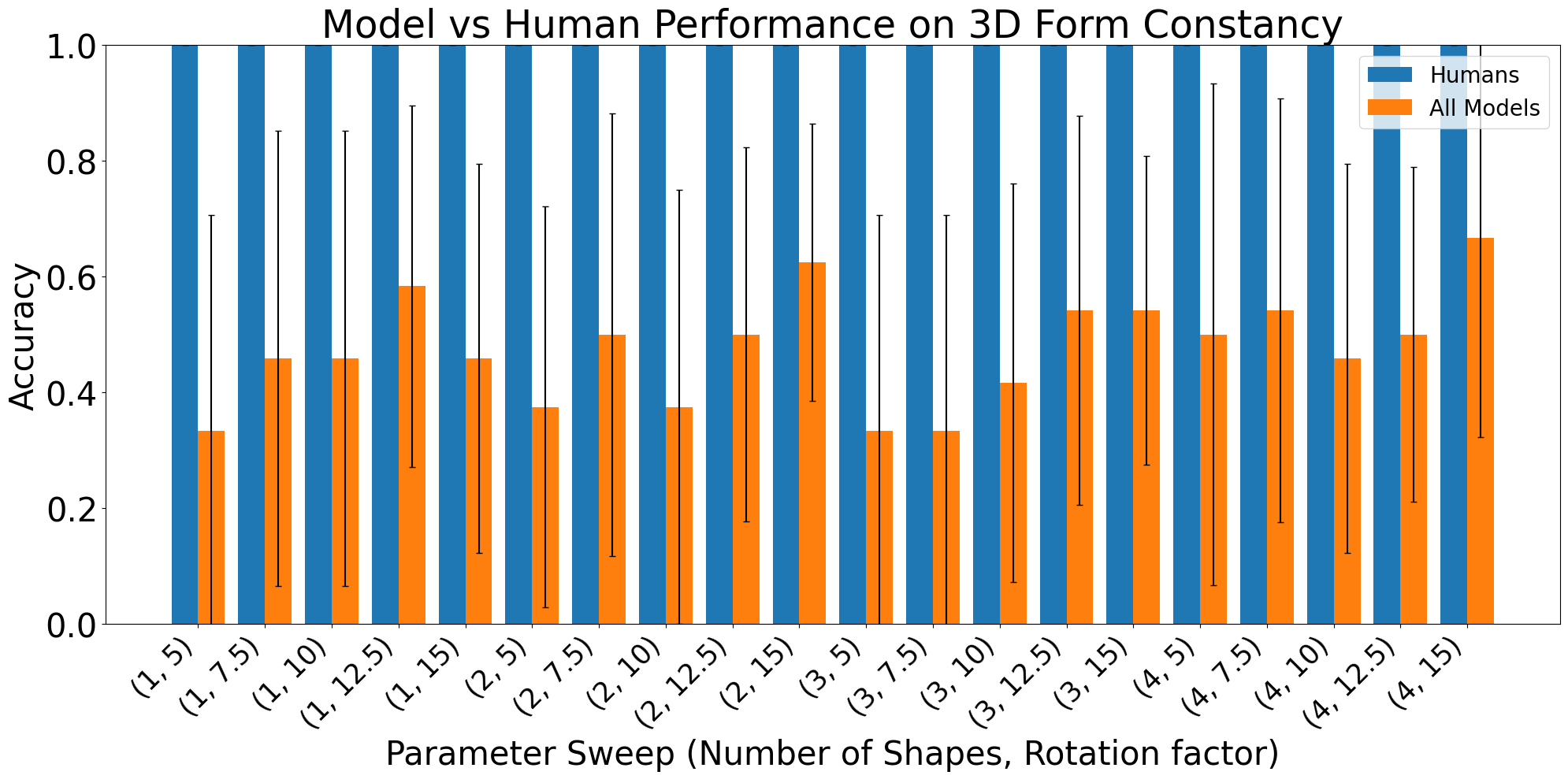} \\
    \end{tabular}
    \caption{Average MLLM performance over a sweep of combinations of control parameters (3D tasks). }
    \label{fig:yourfigure}
\end{figure}

\end{document}